\pdfoutput=1

\documentclass[11pt]{article}

\usepackage[]{ACL2023}

\usepackage{times}
\usepackage{latexsym}
\usepackage{times}
\usepackage{latexsym}
\usepackage{array}
\usepackage{lipsum}
\usepackage{graphicx}
\usepackage{float}
\usepackage{subcaption}
\usepackage{geometry}
\usepackage{array}
\usepackage{booktabs}

\usepackage{CJKutf8}

\usepackage{rotating} 
\usepackage{booktabs} 
\usepackage{tabularx,siunitx}
\usepackage{makecell}
\usepackage{longtable}
\usepackage{comment}

\usepackage{caption}
\usepackage{amsmath,amssymb}
\usepackage{mathtools}
\usepackage{multirow}
\usepackage{multicol}
\usepackage{adjustbox}
\usepackage{pifont}
\usepackage{tabularx}
\usepackage{enumitem}
\usepackage{array, makecell}
\newcolumntype{P}[1]{>{\centering\arraybackslash}p{#1}}
\usepackage{siunitx}

\usepackage[T1]{fontenc}

\usepackage[utf8]{inputenc}

\usepackage{microtype}

\usepackage{inconsolata}

\usepackage{inconsolata}
\usepackage{xcolor,colortbl}
\usepackage{soul}

\definecolor{headcolor}{HTML}{018161}
\definecolor{relationcolor}{HTML}{d95f02}
\definecolor{tailcolor}{HTML}{6560a3}

\title{XToM: Exploring the Multilingual Theory of Mind for Large Language Models}
\newcommand{\ust}{\ensuremath{^\spadesuit}}
\newcommand{\nvidia}{\ensuremath{^\clubsuit}}
\newcommand{\SJTU}{\ensuremath{^\S}}
\newcommand{\LMU}{\ensuremath{^\dagger}}
\newcommand{\TUD}{\ensuremath{^\diamondsuit}}
\newcommand{\MCM}{\ensuremath{^\ddagger}}
\author{Chunkit Chan\ust\thanks{\quad Equal contribution.} \ \ Yauwai Yim\ust$^{*}$ \ Hongchuan Zeng\SJTU$^{*}$ \ Zhiying Zou\SJTU \ Xinyuan Cheng\LMU \ \textbf{Zhifan Sun\TUD}\\ 
\ \ \textbf{Zheye Deng\ust}  \ \  \textbf{Kawai Chung\ust} \ \ \textbf{Yuzhuo Ao\ust} \ \ \textbf{Yixiang Fan\ust} \ \ \textbf{Cheng Jiayang\ust} \ \ \textbf{Ercong Nie\LMU\MCM}\\
\ \ \textbf{Ginny Y. Wong\nvidia}\ \ \textbf{Helmut Schmid\LMU\MCM \ \ Hinrich Schütze\LMU\MCM } \ \ \textbf{Simon See\nvidia}\ \  \textbf{Yangqiu Song\ust}\\
\ust HKUST, Hong Kong
\nvidia NVIDIA AI Technology Center (NVAITC), USA\\
\SJTU Shanghai Jiao Tong University, China
\TUD Technische Universtität Darmstadt, Germany\\
\LMU LMU Munich, Germany
\MCM Munich Center for Machine Learning, Germany\\
\texttt{\{ckchancc, yqsong\}@cse.ust.hk} \ \ \ \ 
}

\begin{document}
\maketitle
\begin{abstract}
Theory of Mind (ToM)—the ability to infer mental states in others—is pivotal for human social cognition. Existing evaluations of ToM in LLMs are largely limited to English, neglecting the linguistic diversity that shapes human cognition. This limitation raises a critical question: \textit{can LLMs exhibit Multilingual Theory of Mind—the capacity to reason about mental states across diverse linguistic contexts?} To address this gap, we present XToM, a rigorously validated multilingual benchmark that evaluates ToM across five languages and incorporates diverse, contextually rich task scenarios. Using XToM, we systematically evaluate LLMs (e.g., DeepSeek R1), revealing a pronounced dissonance: while models excel in multilingual language understanding, their ToM performance varies across languages. Our findings expose limitations in LLMs' ability to replicate human-like mentalizing across linguistic contexts\footnote{Code and data are available at \url{ https://github.com/HKUST-KnowComp/XToM}.}. 
\end{abstract}

\section{Introduction}
Theory of Mind (ToM), the capacity to infer and attribute the mental states of others, is a cornerstone of human social cognition, enabling individuals to navigate complex interpersonal interactions by understanding beliefs, intentions, and emotions~\cite{premack1978does, DBLP:conf/emnlp/MaSPC23}. Numerous scenarios rely on the ToM modeling of the mental states of others, such as 
multiagent-based simulation~\cite{DBLP:conf/ijcai/PynadathM05, DBLP:journals/corr/abs-2408-02559}, planning~\cite{DBLP:conf/ro-man/FavierSA23}, negotiation~\cite{DBLP:conf/acl/YangCN20}, and various forms of reasoning and decision-making~\cite{DBLP:journals/ai/PereiraPS16,rusch2020theory}. 
\begin{figure}[!t]
    \centering
    \includegraphics[width=0.5\textwidth]{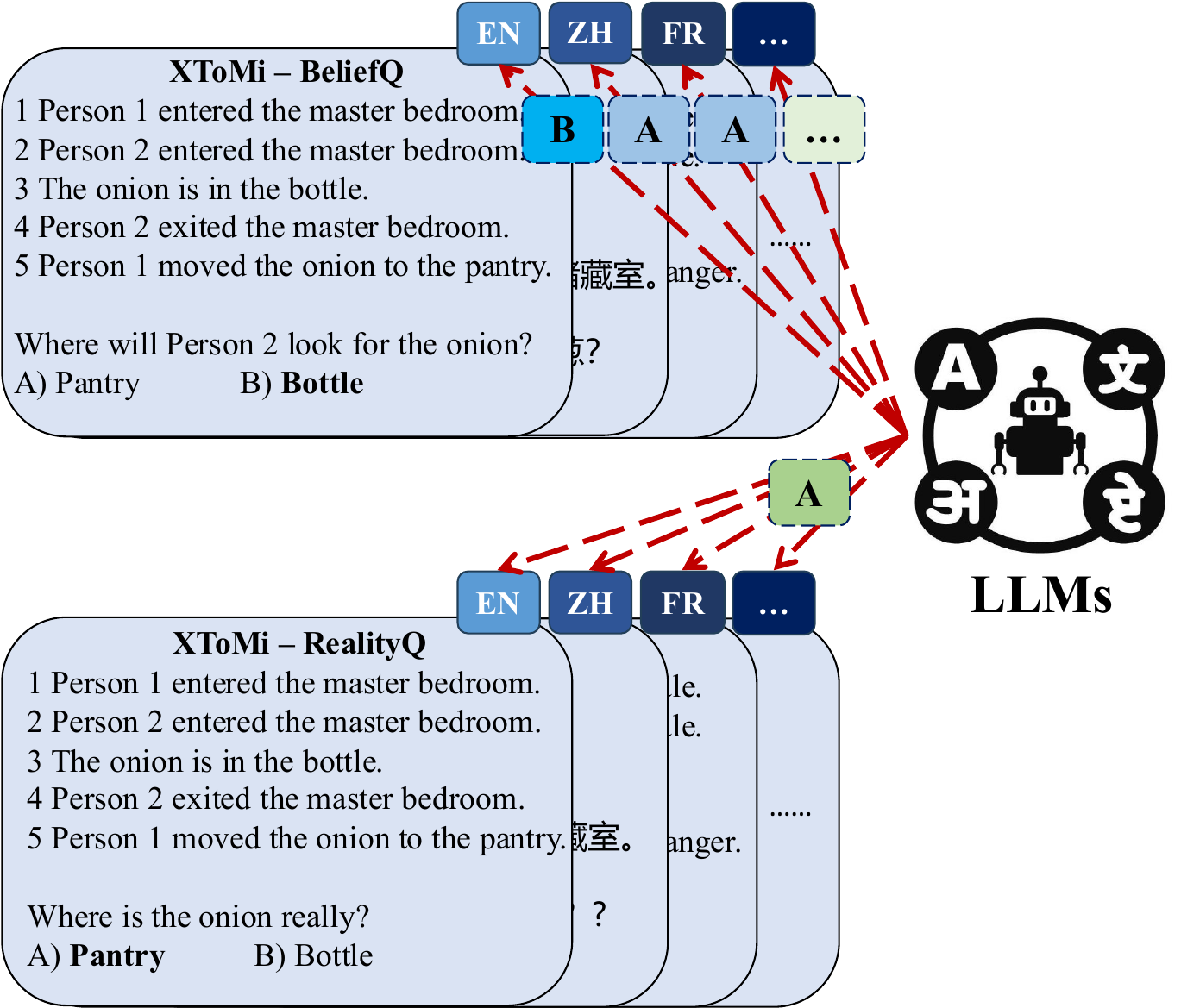}
    \vspace{-0.6cm}
    \caption{A belief question and fact question (i.e., RealityQ) example from the XToMi sub-task of XToM for assessing LLMs’ multilingual theory of mind ability. Ground truth is bold, and LLMs answer correctly for fact questions while answers vary in belief questions for different languages.
     } 
    \vspace{-0.75cm}
    \label{fig:belief_choice}
\end{figure}
Large language models (LLMs) have demonstrated remarkable proficiency in extensive natural language processing (NLP) tasks~\cite{DBLP:journals/corr/abs-2303-12712}, but their capacity for ToM reasoning remains contentious. 
Some previous research believes that LLMs already exhibit a high level of competence in addressing ToM tasks~\cite{strachan2024testing, DBLP:journals/corr/abs-2303-12712, DBLP:journals/corr/abs-2302-02083}, while other studies express doubt and develop benchmarks to illustrate that LLMs do not possess proficient ability in ToM tasks~\cite{DBLP:conf/emnlp/Sap0FC22, DBLP:journals/corr/abs-2302-08399, DBLP:conf/eacl/ShapiraLAZCGSS24}. This contentious landscape reflects a deeper issue: existing ToM research for LLMs has been predominantly confined to English-language contexts, creating a critical blind spot in our understanding of these models' cognitive capabilities.

Existing monolingual benchmarks, though valuable for initial assessment, fail to capture the linguistic diversity inherent to human cognition, potentially inflating performance estimates based on language-specific artifacts rather than genuine reasoning abilities. This limitation prevents us from understanding whether LLMs' apparent ToM capabilities represent robust, language-invariant cognitive processes or merely statistical patterns learned from English training data.
Previous research highlights a nuanced interplay between multilingualism and Theory of Mind (ToM), with bilingual individuals often demonstrating enhanced mental-state reasoning compared to monolinguals. Studies reveal that early bilingual exposure correlates with superior ToM performance, as bilinguals exhibit heightened empathy and outperform monolinguals on tasks requiring belief attribution~\cite{nguyen2014reassessing, javor2016bilingualism, schroeder2018bilinguals, buac2020predictors, DBLP:conf/icml/ZhuNB21}. Given that state-of-the-art large language models (LLMs) such as GPT-4o \cite{DBLP:journals/corr/abs-2410-21276} and DeepSeek R1 \cite{deepseekai2025deepseekr1incentivizingreasoningcapability} are pre-trained on vast multilingual corpora and have demonstrated their superior cross-lingual capabilities \cite{chirkova2024zeroshot, pires-etal-2019-multilingual, wu-dredze-2019-beto}, a critical question emerges: \textit{Do LLMs exhibit Multilingual Theory of Mind—the ability to reason about mental states consistently across diverse linguistic contexts?} 

Investigating this intersection is essential for advancing our fundamental understanding of whether LLMs' apparent "cognitive" capacities are truly language-agnostic or merely artifacts of English-centric training data—a gap underscored by critiques of their superficial statistical learning~\cite{DBLP:conf/acl/BenderK20, DBLP:conf/eacl/ShapiraLAZCGSS24}. Moreover, it has practical implications for deploying these models globally, where consistent reasoning across languages is crucial. To address this, we developed XToM, a high-quality, human-validated multilingual ToM benchmark that encompasses five languages and diverse task scenarios. This resource not only extends ToM evaluation to various languages but also incorporates contextually rich, pragmatically nuanced tasks designed to disentangle language proficiency from genuine mental state reasoning. Our comprehensive evaluation of state-of-the-art models, including DeepSeek R1~\cite{deepseekai2025deepseekr1incentivizingreasoningcapability} and Qwen3-235b-a22b~\cite{qwen3}, reveals a striking dissonance: despite LLMs excelling in multilingual language understanding, discrepancies across different languages in XToM suggest that their apparent reasoning abilities may be surface-level rather than rooted in robust, language-invariant cognition. This paper aims to bridge the gap in multilingual ToM research by introducing three key contributions:

{\begin{itemize}[leftmargin=*]
    \item \textbf{High-Quality Benchmark Creation}: We develop XToM, a high-quality multilingual ToM benchmark encompassing five languages and diverse task scenarios, human-annotated through the Multidimensional Quality Metrics (MQM) framework to ensure robustness and reliability. XToM advances ToM evaluation by integrating linguistic diversity to address critical limitations of prior benchmarks. 
    \item \textbf{Exploration Across Language}: We systematically explore LLMs' ToM capabilities across languages, exposing critical limitations in their multilingual reasoning. Our findings underscore a fundamental disconnect: LLMs' statistical mastery of language patterns does not equate to human-like social cognition.
    \item \textbf{Empirical Insights}: We undertake the necessary empirical experiments to evaluate LLMs on the XToM benchmark and conduct extensive in-depth analysis to explore the LLMs' empirical performance under various settings. 
\end{itemize}}

\section{Related Work}
\subsection{Theory of Mind Benchmarks}
Early computational efforts built upon foundational psychological experiments (i.e., the Sally-Anne test \cite{BaronCohen1985DoesTA}), including the ToM-bAbi dataset \cite{Grant2017HowCM} and its refinement into ToMi \cite{DBLP:conf/emnlp/LeBN19}, laid the groundwork for assessing false belief understanding as a core aspect of ToM. 
More recently, several new benchmarks have emerged to address limitations in previous ToM evaluations, offering more realistic and challenging scenarios.
FANToM \cite{DBLP:conf/emnlp/0002SZ0K0S23} assesses ToM capabilities in dialogue contexts, while TOMBench \cite{Chen2024ToMBenchBT} provides a comprehensive framework with 31 social cognition abilities across eight tasks. 
Furthermore, NegotiationToM \cite{DBLP:conf/emnlp/ChanJYDF0L0WS24} explores multi-dimensional mental states (i.e., desire, belief, and intention) in real-world negotiation dialogues. 
While these advancements have significantly deepened our understanding of ToM in AI systems, exploring multilingual ToM capabilities remains underexplored, presenting a critical gap that our work aims to address. More related works on the theory of mind are provided in Appendix~\ref{sec:Related_Works_for_Theory_of_Mind}. It is worth noting that concurrent work \cite{DBLP:journals/corr/abs-2411-15999} developed a multilingual theory of mind benchmark (i.e., Multi-TOM) by leveraging LLMs to translate the TOMBench \cite{Chen2024ToMBenchBT} into various languages automatically. Their work focused on how cultural contexts relate to the theory of mind, whereas our research aims to explore LLM's multilingual theory of mind ability. Moreover, a key distinction lies in the translation quality assurance process: Multi-TOM relies on the back-translation of many samples using Google Translate to preserve the core narrative. In contrast, our benchmark employs experts to manually correct and evaluate the translations based on the Multidimensional Quality Metrics (MQM) framework, which assesses quality across more diverse dimensions than Multi-TOM.

\subsection{Multilingual Capabilities of LLMs}
State-of-the-art large language models (LLMs), such as GPT-4o \cite{DBLP:journals/corr/abs-2410-21276}, LLaMA \cite{DBLP:journals/corr/abs-2407-21783}, Mistral \cite{jiang2023mistral7b}, and DeepSeek R1 \cite{deepseekai2025deepseekr1incentivizingreasoningcapability}, are pre-trained on multilingual corpora, leveraging linguistic similarities and shared representations to enhance low-resource language performance \cite{zeng-etal-2025-converging, wendler2024llamasworkenglishlatent, dumas2025separatingtonguethoughtactivation}. Despite their cross-lingual capabilities \cite{chirkova2024zeroshot, pires-etal-2019-multilingual, wu-dredze-2019-beto}, performance disparities persist, prompting efforts to quantify and mitigate them \cite{li2024languagerankermetricquantifying, kumar2024bridginggapdynamiclearning, zeng2024multilingualbrainsurgeonlarge}. Multilingual models have demonstrated strong reasoning abilities, including the capacity to reason in underrepresented languages \cite{shi2022languagemodelsmultilingualchainofthought}. Techniques like chain-of-thought prompting \cite{qin2023crosslingualpromptingimprovingzeroshot} and preference optimization \cite{she2024mapoadvancingmultilingualreasoning} have further improved reasoning performance in non-dominant languages. However, the ability of large language models to exhibit multilingual Theory of Mind reasoning across diverse linguistic contexts remains largely underexplored, and further investigation in this area is necessary.

\section{XToM}
The construction pipeline of XToM is systematically organized into four distinct phases to ensure linguistic diversity and high-quality outputs. These phases include \textsection\ref{section_Source_Dataset_Sampling} \textbf{Source Data Sampling}, \textsection\ref{section_Preprocessing_Translation} \textbf{Preprocessing and Translation}, \textsection\ref{section_Human_Annotation} \textbf{Human Annotation}, and \textsection\ref{section_Human_Evaluation} \textbf{Quality Evaluation}.

\paragraph{XToM.}
We use $\mathcal{M}$ to denote a multilingual parallel dataset for XToM and define an XToM instance $M$ as a set of story or dialogue $S$, question $Q$, and answer $A$. Each instance $M$ has multiple semantically equivalent versions in different languages. Let us denote a \textit{language} by $l \in \mathcal{L}$ where $\mathcal{L}=\{en, zh, de, fr, ja\}$ and $|\mathcal{L}|$ is the number of languages of interest. $\{en, zh, de, fr, ja\}$ represent English, Chinese, German, French, and Japanese, respectively. Then, $M^{l}$ is the instance $M$ in the language $l$.
For example, $M^{\text{en}}$ and  $M^{\text{fr}}$ denote the instance with the same meaning but in English (en) and French (fr), respectively. Therefore, the multilingual dataset $\mathcal{M}$ consists of $S\times|\mathcal{L}|\times K$ questions and answers. $K$ is the number of questions and answers in a story or dialogue with the same language. Finally, we can formally describe a multilingual dataset $\mathcal{M}$ for XToM:
\begin{align}
\begin{split}
\forall M\in \mathcal{M}, ~~&\forall (l_x,l_y)\in \mathcal{L}^2, ~~\forall i\in \mathbb{N}_{\leq K},  \\ 
    &{M}^{l_x}_i \bowtie {M}^{l_y}_i~.
\end{split}
\end{align}
We use the notation $\bowtie$ to indicate two instances in different languages (e.g., $l_x$ and $l_y$) are semantically equivalent to each other. $i$ is index of question and answer and $\mathbb{N}_{\leq K}$ is natural number smaller than $K$.

\begin{table*}[t] 
\small
\centering
\resizebox{\linewidth}{!}{ 
\begin{tabular}{c|c|c|c|c|c}
\toprule
\multirow{8}{*}{\rotatebox{90}{\parbox{3cm}{\centering\textsc{XToMi}}}} & \multirow{2}{*}{\textbf{Metric}} & \textbf{en$\rightarrow$zh} & \textbf{en$\rightarrow$de} & \textbf{en$\rightarrow$fr} & \textbf{en$\rightarrow$ja}\\
 {} & {} & \textbf{GPT-4o | DeepL | Human} & \textbf{GPT-4o | DeepL | Human} & \textbf{GPT-4o | DeepL | Human} & \textbf{GPT-4o | DeepL | Human}\\

\cmidrule{2-6}
 & LASER & 0.924 \hspace{0.1cm}|\hspace{0.1cm} 0.920 \hspace{0.1cm}|\hspace{0.1cm} \textbf{0.938} & \textbf{0.950} \hspace{0.1cm}|\hspace{0.1cm} 0.948 \hspace{0.1cm}|\hspace{0.1cm} 0.948 & 0.947 \hspace{0.1cm}|\hspace{0.1cm} \textbf{0.955} \hspace{0.1cm}|\hspace{0.1cm} \textbf{0.955} & 0.927 \hspace{0.1cm}|\hspace{0.1cm} \textbf{0.932} \hspace{0.1cm}|\hspace{0.1cm} 0.927 \\
 & $P_{\text{BERTScore}}$ & 0.870 \hspace{0.1cm}|\hspace{0.1cm} 0.868 \hspace{0.1cm}|\hspace{0.1cm} \textbf{0.877} & \textbf{0.913} \hspace{0.1cm}|\hspace{0.1cm} 0.912 \hspace{0.1cm}|\hspace{0.1cm} 0.912 & 0.899 \hspace{0.1cm}|\hspace{0.1cm} \textbf{0.901} \hspace{0.1cm}|\hspace{0.1cm} 0.900 & \textbf{0.870} \hspace{0.1cm}|\hspace{0.1cm} 0.863 \hspace{0.1cm}|\hspace{0.1cm} \textbf{0.870} \\
 & $R_{\text{BERTScore}}$ & 0.864 \hspace{0.1cm}|\hspace{0.1cm} 0.864 \hspace{0.1cm}|\hspace{0.1cm} \textbf{0.878} & \textbf{0.913} \hspace{0.1cm}|\hspace{0.1cm} 0.911 \hspace{0.1cm}|\hspace{0.1cm} 0.911 & \textbf{0.886} \hspace{0.1cm}|\hspace{0.1cm} \textbf{0.886} \hspace{0.1cm}|\hspace{0.1cm} 0.885 & \textbf{0.860} \hspace{0.1cm}|\hspace{0.1cm} 0.848 \hspace{0.1cm}|\hspace{0.1cm} \textbf{0.860} \\
 & $F_{\text{BERTScore}}$ & 0.867 \hspace{0.1cm}|\hspace{0.1cm} 0.865 \hspace{0.1cm}|\hspace{0.1cm} \textbf{0.884} & \textbf{0.913} \hspace{0.1cm}|\hspace{0.1cm} 0.911 \hspace{0.1cm}|\hspace{0.1cm} 0.911 & 0.892 \hspace{0.1cm}|\hspace{0.1cm} \textbf{0.893} \hspace{0.1cm}|\hspace{0.1cm} \textbf{0.893} & \textbf{0.865} \hspace{0.1cm}|\hspace{0.1cm} 0.855 \hspace{0.1cm}|\hspace{0.1cm} \textbf{0.865} \\
 & Human Evaluation (MQM) & 95.91 \hspace{0.1cm}|\hspace{0.1cm} \quad - \quad \hspace{0.1cm}|\hspace{0.1cm} \textbf{99.27} & 94.09 \hspace{0.1cm}|\hspace{0.1cm} \quad - \quad  \hspace{0.1cm}|\hspace{0.1cm} \textbf{99.27} & 97.09 \hspace{0.1cm}|\hspace{0.1cm}  \quad - \quad \hspace{0.1cm}|\hspace{0.1cm} \textbf{99.52} &  97.92 \hspace{0.1cm}|\hspace{0.1cm} \quad - \quad \hspace{0.1cm}|\hspace{0.1cm} \textbf{98.39} \\

\cmidrule{2-6}
 & Avg. Word Count & 70.16 & 52.03 & 61.88 & 122.16 \\
\midrule
\midrule
\multirow{7}{*}{\rotatebox{90}{\parbox{2cm}{\centering\textsc{XFANToM}}}} 
 & LASER                    & \textbf{0.958} \hspace{0.1cm}|\hspace{0.1cm} 0.950 \hspace{0.1cm}|\hspace{0.1cm} \textbf{0.958} & \textbf{0.979} \hspace{0.1cm}|\hspace{0.1cm} 0.978 \hspace{0.1cm}|\hspace{0.1cm} \textbf{0.979} & \textbf{0.974 \hspace{0.1cm}|\hspace{0.1cm} 0.974 \hspace{0.1cm}|\hspace{0.1cm} 0.974} & 0.949 \hspace{0.1cm}|\hspace{0.1cm} 0.940 \hspace{0.1cm}|\hspace{0.1cm} \textbf{0.956} \\
 & $P_{\text{BERTScore}}$ & \textbf{0.841 \hspace{0.1cm}|\hspace{0.1cm} 0.841 \hspace{0.1cm}|\hspace{0.1cm} 0.841} & \textbf{0.883} \hspace{0.1cm}|\hspace{0.1cm} 0.881 \hspace{0.1cm}|\hspace{0.1cm} \textbf{0.883} & 0.869 \hspace{0.1cm}|\hspace{0.1cm} \textbf{0.870} \hspace{0.1cm}|\hspace{0.1cm} 0.869 & \textbf{0.831} \hspace{0.1cm}|\hspace{0.1cm} 0.826 \hspace{0.1cm}|\hspace{0.1cm} \textbf{0.831} \\
 & $R_{\text{BERTScore}}$ & 0.846 \hspace{0.1cm}|\hspace{0.1cm} 0.845 \hspace{0.1cm}|\hspace{0.1cm} \textbf{0.847} & \textbf{0.883} \hspace{0.1cm}|\hspace{0.1cm} 0.881 \hspace{0.1cm}|\hspace{0.1cm} \textbf{0.883} & 0.865 \hspace{0.1cm}|\hspace{0.1cm} \textbf{0.866} \hspace{0.1cm}|\hspace{0.1cm} 0.865 & \textbf{0.834} \hspace{0.1cm}|\hspace{0.1cm} 0.820 \hspace{0.1cm}|\hspace{0.1cm} \textbf{0.834} \\
 & $F_{\text{BERTScore}}$ & \textbf{0.844} \hspace{0.1cm}|\hspace{0.1cm} 0.843 \hspace{0.1cm}|\hspace{0.1cm} \textbf{0.844} & \textbf{0.883} \hspace{0.1cm}|\hspace{0.1cm} 0.881 \hspace{0.1cm}|\hspace{0.1cm} \textbf{0.883} & 0.867 \hspace{0.1cm}|\hspace{0.1cm} \textbf{0.868} \hspace{0.1cm}|\hspace{0.1cm} 0.867 & \textbf{0.833} \hspace{0.1cm}|\hspace{0.1cm} 0.825 \hspace{0.1cm}|\hspace{0.1cm} \textbf{0.833} \\
 & Human Evaluation (MQM) & 99.21 \hspace{0.1cm}|\hspace{0.1cm} \quad - \quad \hspace{0.1cm}|\hspace{0.1cm} \textbf{99.99} & 98.87 \hspace{0.1cm}|\hspace{0.1cm} \quad - \quad \hspace{0.1cm}|\hspace{0.1cm} \textbf{99.47} & 99.47 \hspace{0.1cm}|\hspace{0.1cm} \quad - \quad \hspace{0.1cm}|\hspace{0.1cm} \textbf{99.91} & 99.90 \hspace{0.1cm}|\hspace{0.1cm} \quad - \quad \hspace{0.1cm}|\hspace{0.1cm} \textbf{99.94} \\

 \cmidrule{2-6}
 & Avg. Word Count & 1031.9 & 652.13 & 727.81 & 1522.57 \\ 

\midrule
\midrule
\multirow{7}{*}{\rotatebox{90}{\parbox{2cm}{\centering\scriptsize\textsc{XNegotiationToM}}}} 
 & LASER                    & \textbf{0.904} \hspace{0.1cm}|\hspace{0.1cm} 0.901 \hspace{0.1cm}|\hspace{0.1cm} 0.891 & \textbf{0.968} \hspace{0.1cm}|\hspace{0.1cm} 0.964 \hspace{0.1cm}|\hspace{0.1cm} \textbf{0.968} & \textbf{0.957} \hspace{0.1cm}|\hspace{0.1cm} 0.950 \hspace{0.1cm}|\hspace{0.1cm} 0.956 & 0.910 \hspace{0.1cm}|\hspace{0.1cm} 0.900 \hspace{0.1cm}|\hspace{0.1cm} \textbf{0.912} \\
 & $P_{\text{BERTScore}}$ & \textbf{0.858 \hspace{0.1cm}|\hspace{0.1cm} 0.858 \hspace{0.1cm}|\hspace{0.1cm} 0.858} & \textbf{0.904} \hspace{0.1cm}|\hspace{0.1cm} 0.901 \hspace{0.1cm}|\hspace{0.1cm} \textbf{0.904} & \textbf{0.893} \hspace{0.1cm}|\hspace{0.1cm} 0.891 \hspace{0.1cm}|\hspace{0.1cm} \textbf{0.893} & \textbf{0.844} \hspace{0.1cm}|\hspace{0.1cm} 0.831 \hspace{0.1cm}|\hspace{0.1cm} 0.831 \\
 & $R_{\text{BERTScore}}$ & 0.863 \hspace{0.1cm}|\hspace{0.1cm} 0.860 \hspace{0.1cm}|\hspace{0.1cm} \textbf{0.865} & \textbf{0.900} \hspace{0.1cm}|\hspace{0.1cm} 0.899 \hspace{0.1cm}|\hspace{0.1cm} \textbf{0.900} &  0.877 \hspace{0.1cm}|\hspace{0.1cm} \textbf{0.878} \hspace{0.1cm}|\hspace{0.1cm} 0.877 & \textbf{0.842} \hspace{0.1cm}|\hspace{0.1cm} 0.832 \hspace{0.1cm}|\hspace{0.1cm} 0.834 \\
 & $F_{\text{BERTScore}}$ & \textbf{0.861} \hspace{0.1cm}|\hspace{0.1cm} 0.860 \hspace{0.1cm}|\hspace{0.1cm} \textbf{0.861} & \textbf{0.902} \hspace{0.1cm}|\hspace{0.1cm} 0.900 \hspace{0.1cm}|\hspace{0.1cm} \textbf{0.902} & \textbf{0.885} \hspace{0.1cm}|\hspace{0.1cm} 0.884 \hspace{0.1cm}|\hspace{0.1cm} \textbf{0.885} & \textbf{0.843} \hspace{0.1cm}|\hspace{0.1cm} 0.830 \hspace{0.1cm}|\hspace{0.1cm} 0.832 \\
 & Human Evaluation (MQM) & 96.37 \hspace{0.1cm}|\hspace{0.1cm} \quad - \quad \hspace{0.1cm}|\hspace{0.1cm} \textbf{99.74} & 95.00 \hspace{0.1cm}|\hspace{0.1cm} \quad - \quad \hspace{0.1cm}|\hspace{0.1cm} \textbf{99.75} & 97.54 \hspace{0.1cm}|\hspace{0.1cm} \quad - \quad \hspace{0.1cm}|\hspace{0.1cm} \textbf{99.00} & 98.06 \hspace{0.1cm}|\hspace{0.1cm} \quad - \quad \hspace{0.1cm}|\hspace{0.1cm} \textbf{99.99} \\

 \cmidrule{2-6}
 & Avg. Word Count & 187.04 & 133.94 & 150.1 & 313.25 \\  

\bottomrule
\end{tabular}
}
\vspace{-0.3cm}
\caption{Benchmark quality evaluation for different translation methods across three partitions of XToM benchmarks (i.e., XToMi, XFANToM, and XNegotiationToM). GPT-4o, DeepL, Human indicate GPT-4o translation, DeepL translation, and Human Annotation. $P_{\text{BERTScore}}$, $R_{\text{BERTScore}}$, and $F_{\text{BERTScore}}$ refer to the BERTScore Precision, BERTScore Recall, and BERTScore F1. $en, zh, de, fr,$ and $ja$ represent English, Chinese, German, French, and Japanese, respectively. The average word count of the English version of  ToMi, FANToM, and NegotiationToM are 51.39, 635.68, and 139.12, respectively.
}  
\label{tab:dataset_quality_evaluation_metrics}
\vspace{-0.6cm}
\end{table*}

\subsection{Source Dataset Sampling} \label{section_Source_Dataset_Sampling}
To ensure the quality, diversity, and representativeness of the source data used for dataset construction, we systematically curated a balanced subset of 300 stories and dialogues from three distinct well-established benchmarks, ranging from theoretical to application of the Theory of Mind (ToM). These three well-established datasets are ToMi~\cite{DBLP:conf/emnlp/LeBN19}, FANToM~\cite{DBLP:conf/emnlp/0002SZ0K0S23}, and NegotiationToM~\cite{DBLP:conf/emnlp/ChanJYDF0L0WS24}, utilized to create three subsets of XToM (i.e., XToMi, XFANToM, and XNegotiationToM). More details of source dataset sampling are in Appendix~\ref{sec:Sampling_Source_Data}.

\paragraph{Potential Contamination}
To prevent the potential contamination issue in sampled ToM benchmarks, we follow the established protocols from prior works~\cite{DBLP:conf/iclr/GolchinS24, DBLP:conf/aaai/LiF24} to verify contamination issues. For FANToM and NegotiationToM, none of the LLMs’ generated responses matched with the data instance by using two established protocols across various languages, which indicates most instances of the XToM benchmark (i.e., two subtasks XFANToM and XNegotiationToM) are not identified as suffering from the data contamination issue, and the experimental results of the paper are reliable and valuable. An interesting finding is that some LLM-generated responses match some ToMi story patterns or even data instances. However, to ensure the representativeness of the dataset, we still have to collect the ToMi instances for reference purposes, as ToMi is a classical and widely used ToM benchmark in the ToM field. 
More details on the experimental process are reported in Appendix~\ref{Verification_Contamination}.

\subsection{Dataset Preprocessing and Translation} \label{section_Preprocessing_Translation}
\paragraph{Preprocessing.} 
The original dataset of both FANToM and ToMi contained numerous English names (e.g., Kailey and Sally). For example, FANToM sampled English names from the Top-1K names in the US SSN database, and it introduced language-specific biases into XToM, resulting in an influence on the LLM performance~\cite{DBLP:conf/emnlp/NghiemPZD24,DBLP:conf/emnlp/MukherjeeASBAC24,DBLP:conf/emnlp/SandovalZCD23}. To ensure the neutrality of our dataset, we manually replaced all English names with the generic format "Person number" to mitigate this potential bias. Moreover, label consistency in the translated instance may significantly affect the LLMs' performance during the evaluation. For instance, the intention label "Discover-Preference" from NegotiationToM may be translated to "Découvrir-Préférence" or "Découvrir la préférence" in French. To address this, we consulted native speakers of each target language to identify the most contextually appropriate and semantically accurate translation for a tailored set of labels and specific terms. Based on their feedback, we applied a mapping strategy to standardize the translation of the term "Person" and labels across all languages, ensuring uniformity in the dataset. 

\paragraph{Translation.} 
Since the translation quality is crucial for ensuring the reliability of translated datasets and experimental results in our cross-linguistic research, we employ the two commonly used translation methods (i.e., GPT-4o \cite{DBLP:journals/corr/abs-2410-21276} and DeepL\footnote{https://www.deepl.com/}) for translating the sampled data with the original language (i.e., English) to multiple target language Chinese, German, French, and Japanese. After conducting the automatic translation quality evaluation, the evaluated result shown in Table~\ref{tab:dataset_quality_evaluation_metrics} demonstrated that the overall translation ability of GPT-4o is better than DeepL in our dataset. Therefore, we chose the GPT-4o translated version for the human annotation phase. Two automatic translation quality evaluation metrics were used, which are LASER\footnote{https://github.com/facebookresearch/LASER} \cite{DBLP:conf/acl/ArtetxeS19, DBLP:journals/tacl/ArtetxeS19} and BERTScore\footnote{https://github.com/Tiiiger/bert\_score} \cite{DBLP:conf/iclr/ZhangKWWA20}.

\subsection{Human Annotation} \label{section_Human_Annotation}
We employ three qualified human annotators for each language to refine the translated data, ensuring alignment with the predefined dimensions of the Multidimensional Quality Metrics (MQM) framework\footnote{https://themqm.org/}~\cite{burchardt-2013-multidimensional, mariana2014multidimensional, DBLP:conf/amta/LommelGMWSGVBSFIHN24}. This process involves correcting errors and enhancing semantic accuracy, grammatical correctness, contextual coherence, and stylistic consistency. The MQM framework is a widely adopted standard for analytical translation quality evaluation (TQE), serving for human annotation and evaluation purposes, systematically assessing and counting translation errors across eight predefined dimensions~\cite{DBLP:journals/tacl/FreitagFGRTM21,DBLP:conf/coling/Li0MAI25}. These dimensions include Terminology, Accuracy, Linguistic Conventions, Style, Locale Conventions, Audience Appropriateness, Design and Markup, and Custom (e.g., label and name consistency, as discussed in \textsection\ref{section_Preprocessing_Translation}). To maintain high annotation quality, each annotator undergoes a preliminary training phase to understand the definitions and examples associated with each dimension. 
The qualification details of human annotators are outlined in Appendix~\ref{sec:Human_Annotator_Qualification}, and comprehensive details regarding the framework and its pre-defined dimensions are in Appendix \ref{mqm}.

\subsection{Human Evaluation}\label{section_Human_Evaluation}
To guarantee the dataset quality and accurate alignment with established dimensions, we employ three qualified human annotators for each language to assess the translation quality of XToM by using the Multidimensional Quality Metrics (MQM) framework. The details of the qualifications of human annotators can be found in Appendix~\ref{sec:Human_Annotator_Qualification}. The MQM framework computes the overall quality score (OQS) by counting the translation error across eight specified dimensions according to the formula shown in Table~\ref{tab:RQS} in Appendix~\ref{sec:Human_Annotator_Qualification}. While the standard passing threshold for the OQS in the MQM framework is typically set at 90 overall quality score, we adopted a more stringent threshold of 95 to ensure higher-quality annotated data. This elevated threshold was used to determine whether a translated instance passes or fails.

\paragraph{Discussion on Length Effect on Translation Quality Evaluation}
To explore the translation quality of our dataset, we performed an automatic quality evaluation and human evaluation on the translated instances from various source datasets, which reveals a length effect on the automatic evaluation methods. The result is shown in Table~\ref{tab:dataset_quality_evaluation_metrics}. We observed that automatic metrics like LASER and BERTScore exhibit diminished effectiveness as word length increases, resulting in the LASER score and BERTScore of the human-corrected version being similar to the GPT-4o translated version in the XFANToM portion. This aligns with findings from prior works~\cite{DBLP:journals/corr/abs-2412-17592, DBLP:conf/blackboxnlp/WangGWCSSZTY21, DBLP:journals/corr/abs-2306-05183}, which \textit{highlighted the challenges of automatic context-aware evaluation in longer texts and the fact that it leads to inconsistent evaluation scores.} Moreover, BERTScore gives different correct translations of the same sentence with lower scores when that sentence contains difficult words that have little lexical meaning or have an ambiguous meaning (i.e., function words)~\cite{DBLP:conf/wmt/HannaB21}, resulting in the BERTScore of the human corrected version lower than the GPT-4o translated version for XToMi (en$\rightarrow$de) and XNegotiationToM (en$\rightarrow$ja).
However, this issue is mitigated by using the MQM framework, which computes the overall quality score by using the \textit{Per-Word Penalty} and taking the word count into consideration. We observed that the human-annotated version received a higher overall quality score than the GPT-4o translated version, demonstrating the high quality of the human-annotated version. By using the MQM framework, we regard the pass or fail label of data instances in human evaluation phrases as the label for computing the inter-annotator agreement.
We observed high inter-annotator agreement, and the overall Fleiss’s $\kappa$ is 95.18\% \cite{fleiss1971measuring} for the XToM benchmark. The breakdown computation of $\kappa$ is shown in Table~\ref{tab:kappa} in the Appendix. 

\begin{figure*}[!t]
    \centering
    \includegraphics[width=\textwidth]{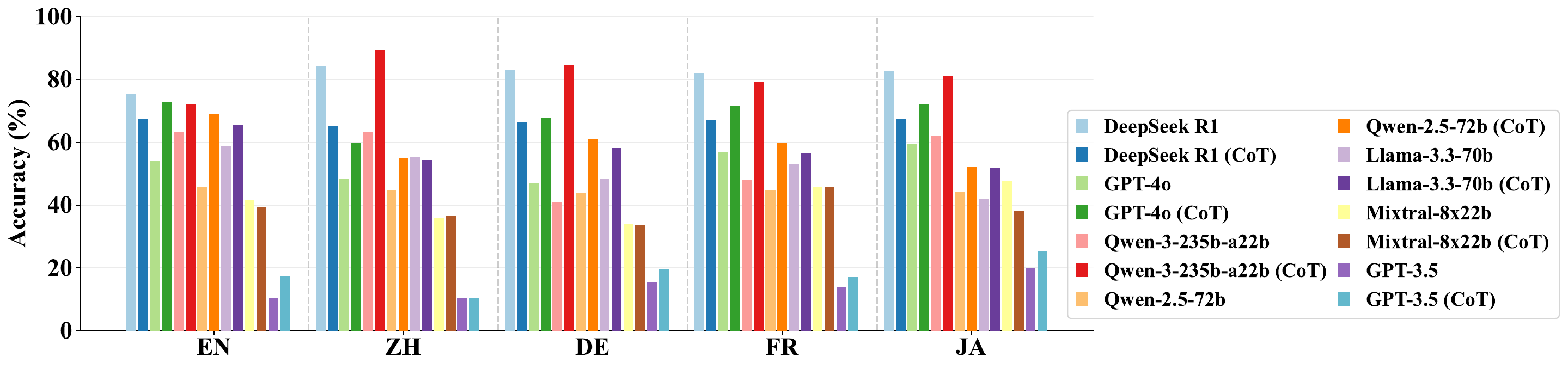}
    \vspace{-0.9cm}
    \caption{Performance comparison of different models on false belief questions across languages in XFANToM.}
    \vspace{-0.3cm}
    \label{fig:belief_choice_fantom}
\end{figure*}

\begin{figure*}[!t]
    \centering
    \includegraphics[width=\textwidth]{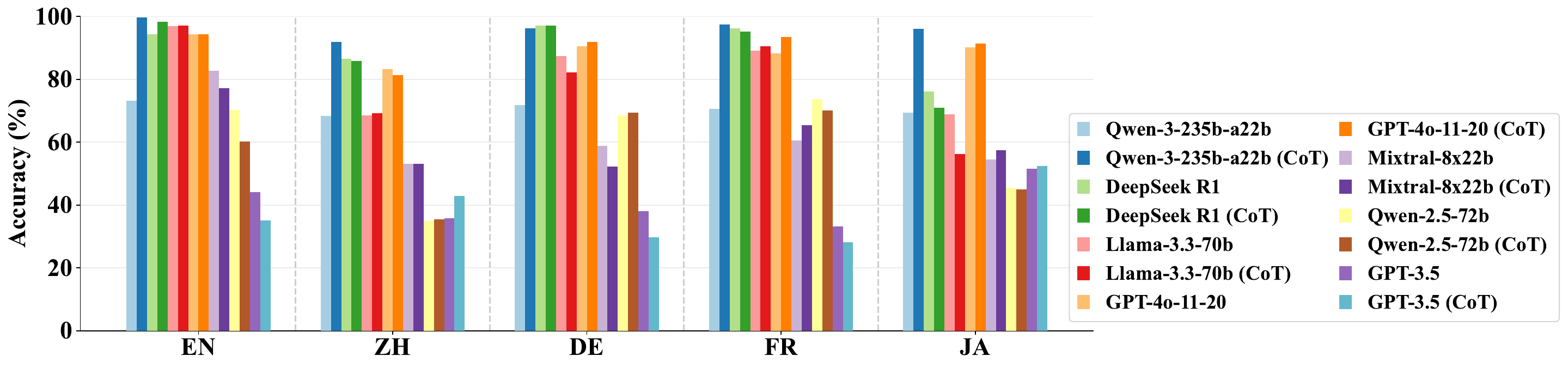}
    \vspace{-0.9cm}
    \caption{Performance comparison of different models on false belief questions across languages in XToMi.}
    \vspace{-0.3cm}
    \label{fig:belief_choice_tomi}
\end{figure*}

\begin{figure*}[!t]
    \centering
    \includegraphics[width=\textwidth]{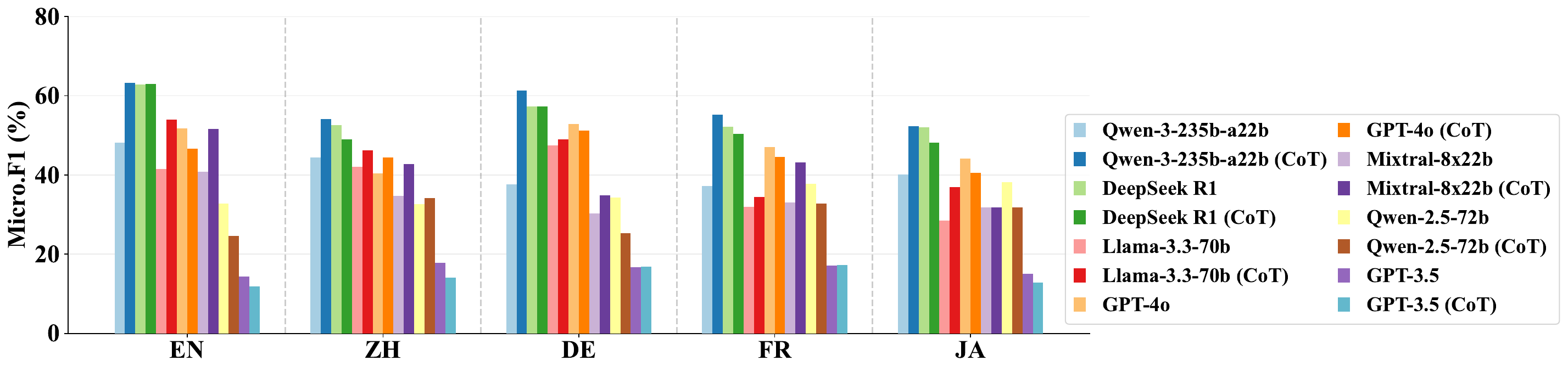}
    \vspace{-0.9cm}
    \caption{Performance comparison of different models on belief choice across languages in XNegotiationToM.}
    \vspace{-0.6cm}
    \label{fig:belief_choice_negtom}
\end{figure*}
\subsection{Dataset Statistics}
XToM contains 900 stories from ToMi, FANToM, and NegotiationToM, which range from theoretical to application and provide contextual diversity for assessing the LLM performance across different languages. There are a total of 15,115 questions, which are 3,090 questions for XFANToM, 3,235 questions for XToMi, and 8,790 questions for XNegotiationToM.
The detailed statistics are shown in Table~\ref{tab:dataset_stats} in the Appendix.

\section{Experiments}
\subsection{Experimental Settings}
In this work, we test ten recent state-of-the-art large language models: DeepSeek R1, Qwen-2.5-7b-instruct, Qwen-2.5-72b-instruct, Qwen-3-235b-a22b, llama-3.1-8b-instruct, llama-3.3-70b-instruct, mistral-7b-instruct-v0.3, mixtral-8x22b-instruct-v0.1, GPT-4o-11-20, and GPT-3.5-turbo~\cite{deepseekai2025deepseekr1incentivizingreasoningcapability, DBLP:journals/corr/abs-2410-21276, openai2022chatgpt, DBLP:journals/corr/abs-2412-15115, qwen3, DBLP:journals/corr/abs-2407-21783, DBLP:journals/corr/abs-2310-06825}. 
By following the common practices in the theory of mind field~\cite{DBLP:conf/emnlp/0002SZ0K0S23,DBLP:journals/corr/abs-2306-15448, DBLP:conf/acl/ShapiraZG23}, we test these models with two types of prompts: (1) One is zero-shot prompting, and we utilize the prompt template in the original paper. (2) Another one is the chain of thought (CoT) prompting method by following~\citet{DBLP:conf/nips/Wei0SBIXCLZ22} and using the prompt “let’s think step by step.” For the prompt template in each language, the human annotator was consulted to translate the English prompt template into another language. More configuration details can be found in Appendix~\ref{sec:hyparameter}, and the prompt template and data example refer to Appendix~\ref{sec:Prompt_Template}. 

\paragraph{Evaluation Metrics}
We adopt the identical evaluation metrics utilized in all source benchmarks, except for fact questions, because we reformulate fact questions in XFANToM as the multi-choice format for fair comparison purposes. Accuracy is used as the evaluation metric for multiple-choice belief and fact questions in XFANToM and XToMi. 
In XNegotiationToM, we report the exact match percentages across all three categories of high, medium, and low preferences for desire and belief classification. Only these three preferences that answer correctly count toward the correct answer. For intention classification, we report both micro and macro F1 scores to evaluate the model’s performance across multiple labels comprehensively.

\section{Main Results}
Figures~\ref{fig:belief_choice_fantom}, \ref{fig:belief_choice_tomi}, and \ref{fig:belief_choice_negtom} summarize the main results of the state-of-the-art large language models for belief questions in XToM, while 
Figures~\ref{fig:desire_choice_negtom} and~\ref{fig:intention_choice_negtom} report LLMs' performance on desire and intention questions in XNegotiaitonToM in Appendix~\ref{sec:Appendix_for_Experimental_Result}. 
Based on the evaluation result, we derive the following conclusion: \textbf{Language variation generally impacts model performance.} Most LLMs exhibit performance disparities across different languages, and this discrepancy may arise for two reasons: data containment issues and cultural-linguistic mismatches. First, as shown in Appendix~\ref{Verification_Contamination}, many LLMs were likely exposed to the ToMi dataset during both pre-training and post-training phases, potentially facilitating some transfer of theory of mind capabilities across tasks (as demonstrated in \textsection~\ref{Task Transfer Learning}). However, this transfer may be constrained by linguistic structural and typological differences, resulting in the model's performance varying across languages. Second, cultural context embedded in language guides social cognition, which differs significantly across languages, affecting how mental states are inferred within each linguistic framework~\cite{nguyen2014reassessing, javor2016bilingualism}. The case study in Table~\ref{tab:case_study} in the Appendix illustrates this phenomenon: when festival culture discussions from the FANToM English version were translated to other languages, models struggled to maintain performance. This degradation likely occurs because cultural references that were salient in the source language became less accessible in target languages, revealing reporting biases~\cite{DBLP:conf/cikm/GordonD13} in how models process culturally embedded mental state information across languages.
The detailed performance of more models, including smaller-scale LLMs, are provided in Tables \ref{tab:experiment_FANToM},~\ref{tab:experiment_ToMi}, and \ref{tab:experiment_NegotiationToM} in Appendix~\ref{sec:appendix_result}. Furthermore, more details of error analysis and case study can be found in Appendix~\ref{App:Error_Analysis}. 

\begin{figure}[!t]
    \centering
    \includegraphics[width=0.52\textwidth]{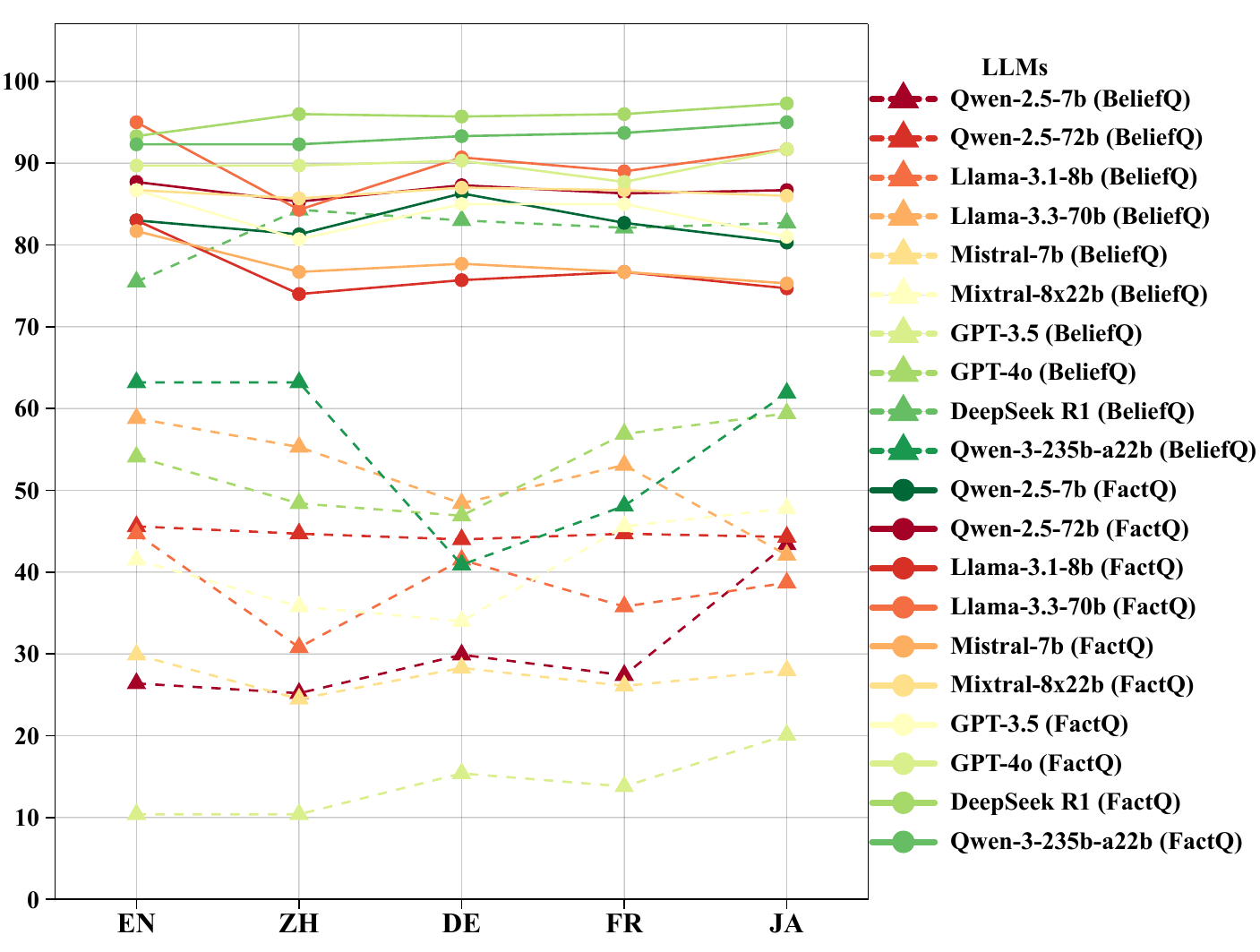}
    \vspace{-0.8cm}
    \caption{LLMs' performance comparison of belief questions and fact questions in XFANToM.}
    \vspace{-0.4cm}
    \label{fig:Fantom_belief_fact}
\end{figure}

\begin{figure}[!t]
    \centering
    \begin{subfigure}{0.5\textwidth}
        \centering
        \includegraphics[width=\textwidth]{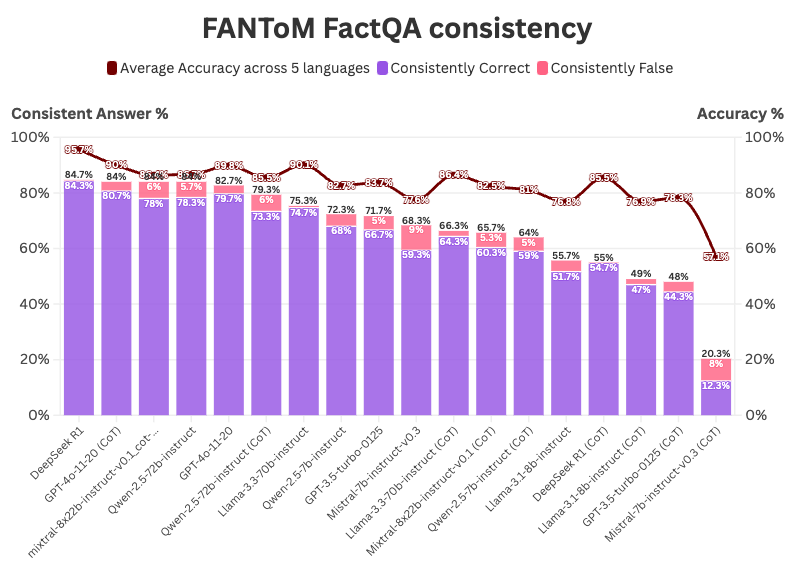}
    \end{subfigure}
    \hfill
    \begin{subfigure}{0.5\textwidth}
        \centering
        \includegraphics[width=\textwidth]{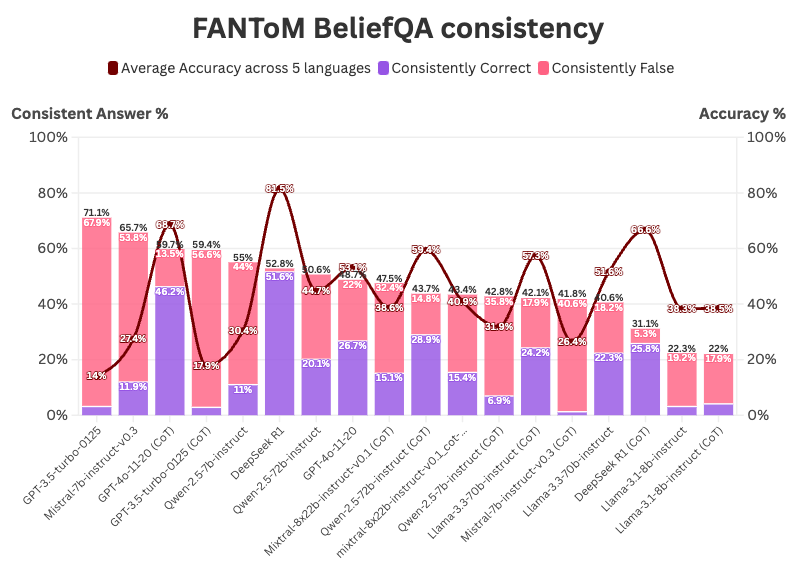}
    \end{subfigure}
    \vspace{-1cm}
    \caption{Consistency analysis across different languages.}
    \vspace{-0.6cm}
    \label{fig:Consistency_fantom}
\end{figure}

\begin{figure*}[!t]
    \centering
    \includegraphics[width=\textwidth]{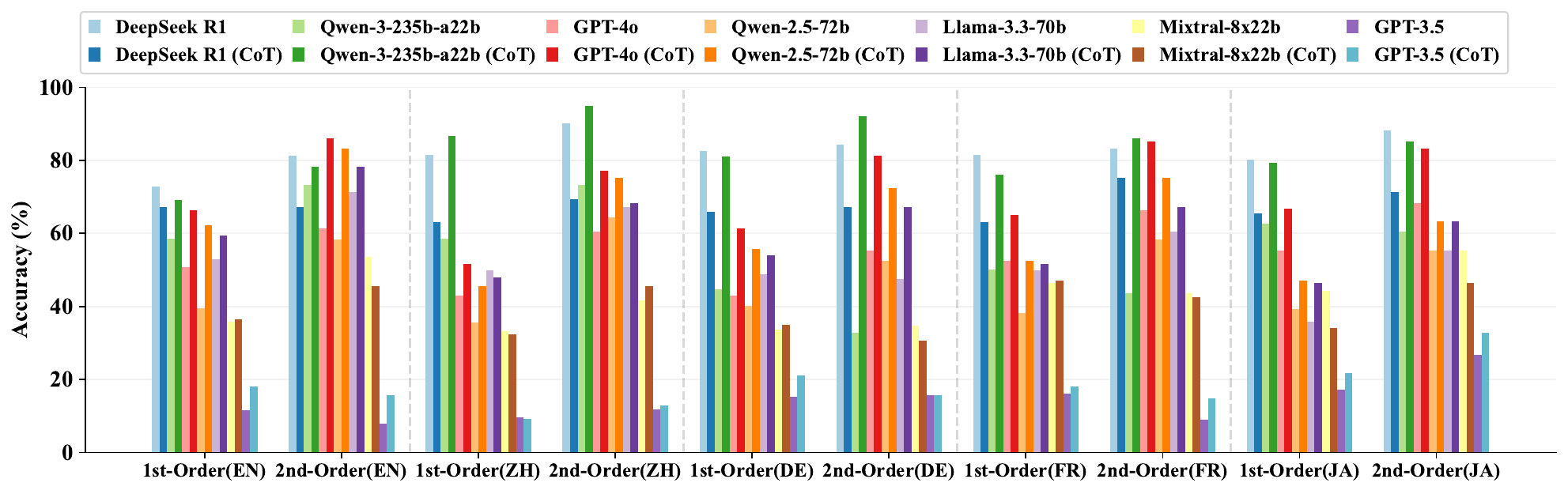}
    \vspace{-0.9cm}
    \caption{Performance comparison of different models on first- and second-order belief questions across languages in XFANToM.}
    \label{fig:1and2order_choice_fantom}
    \vspace{-0.5cm}
\end{figure*}

\begin{figure}[!t]
    \vspace{-0.25cm}
    \centering
    \includegraphics[width=0.5\textwidth]{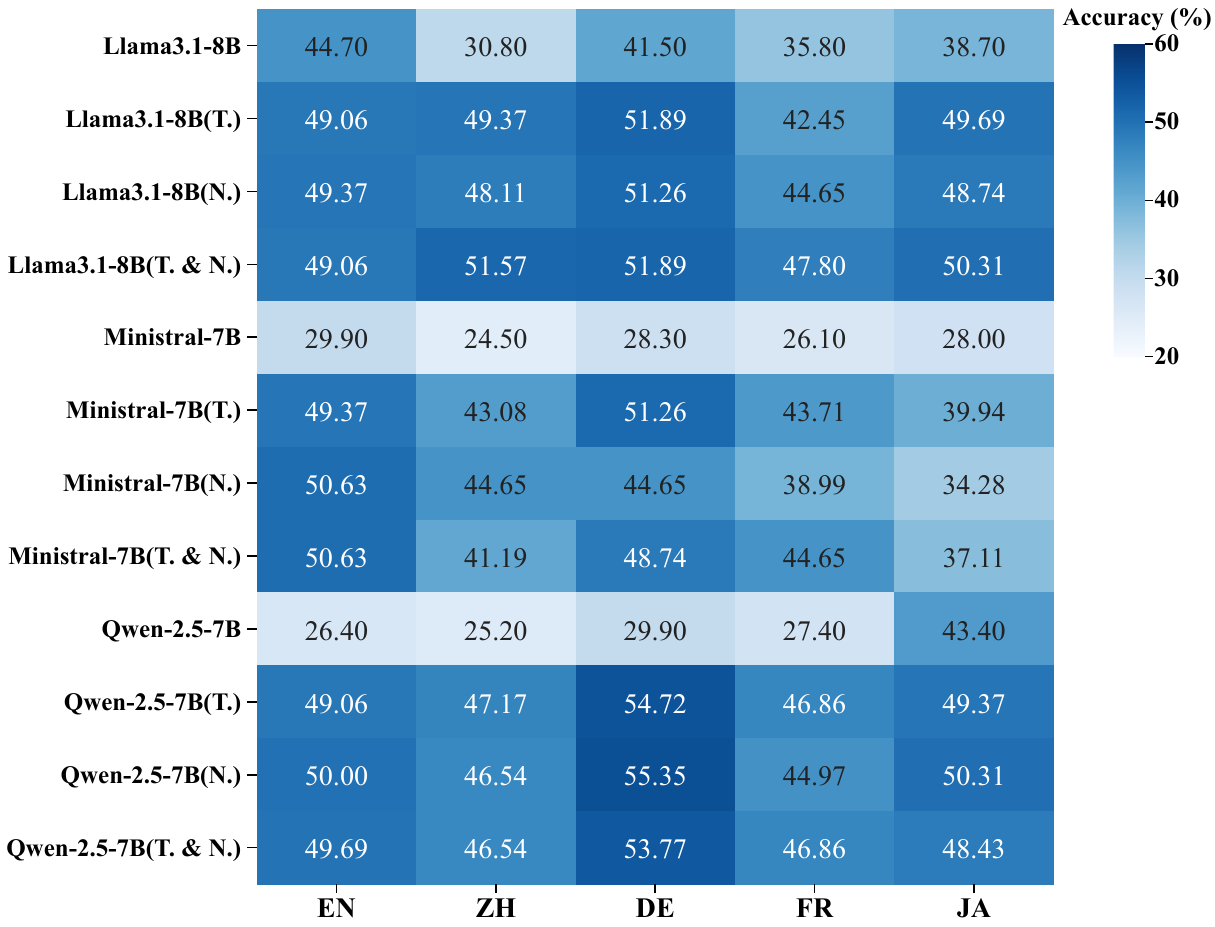}
    \vspace{-0.8cm}
    \caption{Performance comparison of task transfer on belief questions in XFANToM. T. and N. indicate the model was fine-tuned on the XToMi and XNegotiationToM tasks and evaluated on the XFANToM Task.}
    \label{fig:Transfer_Task}
    \vspace{-0.7cm}    
\end{figure}

\subsection{Most LLMs Lack Multilingual TOM Reasoning Ability}
Fact questions in XToM require LLMs to understand the provided context (i.e., story or conversation) across different languages and retrieve the answer in the provided context, while belief questions require the models not only to understand the provided context across languages but also to possess the theory of mind reasoning capability.
Figure~\ref{fig:Fantom_belief_fact} presents the performance comparison of belief and fact questions in XFANToM, and the results indicate that existing LLMs (except DeepSeek R1) lack the multilingual theory of mind capability—the ability to reason about mental states across diverse linguistic settings. While all models obtain high accuracy in the fact questions across different languages, proving their multilingual comprehension abilities, their performance is significantly lower in the multilingual theory of mind reasoning task. The substantial performance gap between the belief and fact questions suggests that LLMs’ statistical mastery of language patterns does not equate to human-like social cognition~\cite{DBLP:conf/eacl/ShapiraLAZCGSS24}.

\subsection{Multilingual Consistency}
To further analyze various LLMs' multilingual consistency, we assess whether models provide the same answers across languages. For clarity, we present FANToM results in Figure \ref{fig:Consistency_fantom} and include others in Appendix \ref{App:consistency}. Models perform well in factQA, achieving high accuracy and consistency, but struggle in BeliefQA, often giving inconsistent and incorrect responses. This suggests that the models' ToM reasoning is significantly weaker than their information retrieval ability.
Fact consistency is significantly higher than belief consistency, indicating that strong factual understanding does not guarantee robust ToM reasoning across languages. While fact accuracy and consistency are strongly correlated, belief accuracy and consistency are not, highlighting greater ToM reasoning ability variation across languages.
Scaling up the models improves fact consistency but not necessarily belief consistency, showing that scaling alone does not enhance ToM abilities.
These findings suggest models perform well on multilingual fact understanding but lack robust cross-linguistic belief reasoning, underscoring the need for specialized ToM training to improve consistency and correctness across languages.

\subsection{First Order versus Second Order ToM}
To fully assess the ToM capability of large language models to understand others' mental states in each complexity, Figure~\ref{fig:1and2order_choice_fantom} presents the LLMs' performance on first- and second-order false belief questions across various languages in XFANToM. The result illustrates that most models receive a higher accuracy in second-order belief questions compared to first-order belief questions across different languages, which is similar to the findings by~\citet{DBLP:conf/emnlp/LeBN19} and~\citet{DBLP:conf/emnlp/0002SZ0K0S23}. 

\subsection{Task Transfer Learning} \label{Task Transfer Learning}
This experiment is crucial for understanding how language models transfer their theory of mind capabilities across different tasks, revealing the effectiveness of specialized fine-tuning for belief reasoning. The results reported in Figure~\ref{fig:Transfer_Task} demonstrate several notable patterns: First, base models without fine-tuning (i.e., Llama3.1-8B, Mistral-7B, and Qwen-2.5-7B) consistently underperform compared to their fine-tuned counterparts, with accuracy improvements of around 10-50\% when models are fine-tuned on either XToMi or XNegotiationToM. Second, models fine-tuned on both datasets simultaneously (T. \& N.) generally achieve the highest performance across languages, particularly evident in Llama3.1-8B(T. \& N.), which reaches 51.57\% accuracy in Chinese (ZH) and maintains strong performance across all languages. The combined fine-tuning approach consistently produced more robust multilingual belief reasoning, suggesting that diverse theory of mind training datasets create complementary benefits that enhance cross-task generalization.

\subsection{Evaluation of LLMs’ Multilingual Transfer Capabilities}
\begin{figure}[!t]
    \centering
    \includegraphics[width=0.5\textwidth]{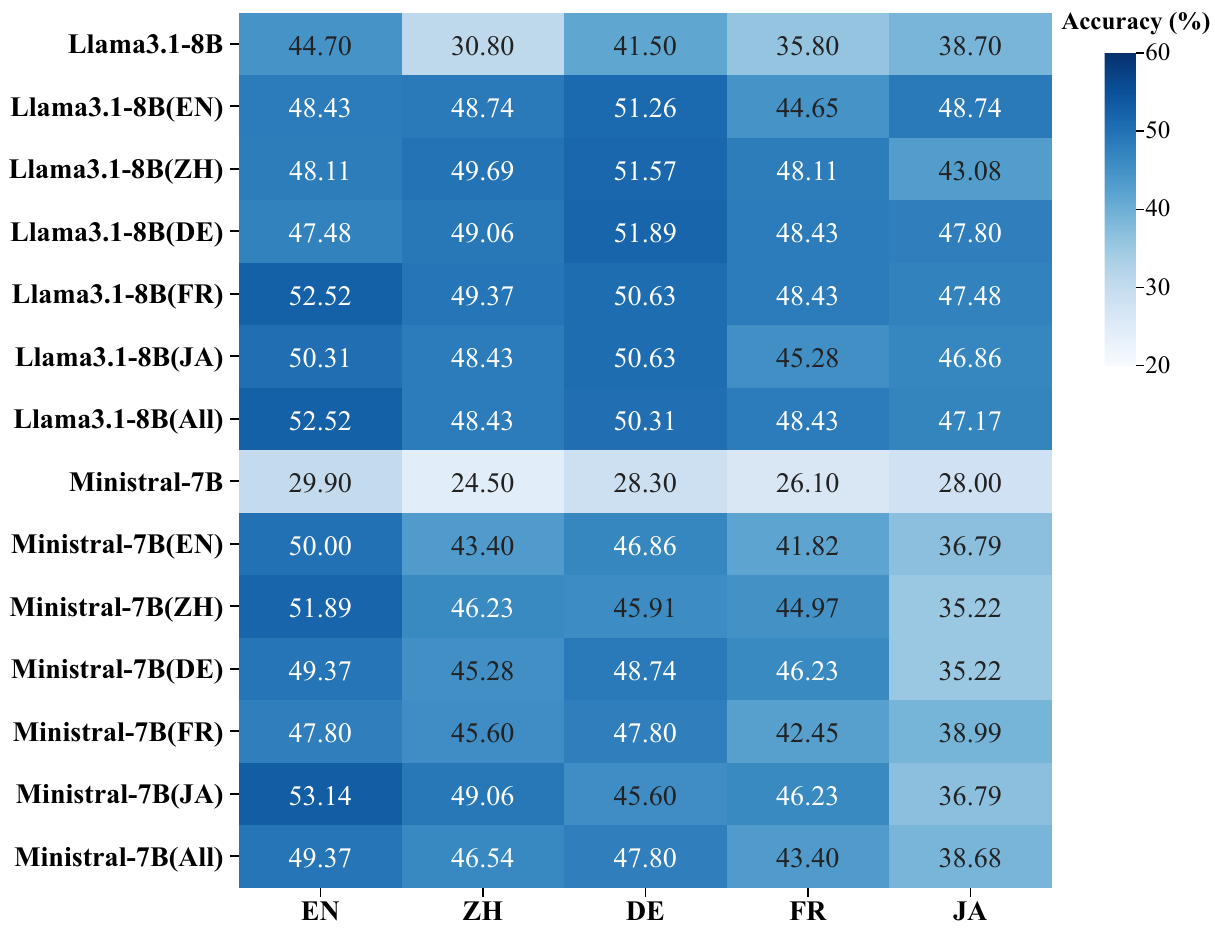}
    \vspace{-0.7cm}
    \caption{Performance comparison of language transfer on belief questions in XFANToM.}
    \vspace{-0.4cm}  
    \label{fig:Transfer_Language}
\end{figure}

By fine-tuning the same number of instances across various languages on open-source LLMs, we evaluate LLMs’ multilingual transfer capabilities. A question is raised here according to the experimental findings: How to enhance the multi-language ToM ability? Therefore, we study whether learning from other languages can help LLM understand the Theory of Mind. Exploring the potential error that occurred in this situation, we want to find whether the gap existed between the model because of pre-training data, lack of source dataset, or lack of the ToM task dataset. Moreover, we intend to see whether it can enhance the LLM performance. Figure~\ref{fig:Transfer_Language} shows the results of LLaMA3.1-8B and Mistral-7B, which are fine-tuned on five monolingual datasets separately and then evaluated across all five languages on belief questions in XFANToM. The results indicate that cross-lingual transferability can be observed in most cases, meaning that training in another monolingual can lead to performance improvements on the test set of the target language. For example, training in French achieves the best performance in English, while training in English yields the best performance in Japanese.

\paragraph{Which is Better: Monolingual Fine-tuning or Multilingual Fine-tuning?}
We analyze the effectiveness of multilingual fine-tuning on belief questions through XFANToM. The results in Figure~\ref{fig:Transfer_Language} indicate that some languages exhibit improved performance after multilingual fine-tuning. For instance, LLaMA3.1-8B outperforms monolingual training in both English and Japanese under the multilingual setting, while Mistral-7B surpasses monolingual performance in Chinese and French.

\subsection{Do Different LLMs have Similar Language Preferences?}\label{App:Language_Preferences}
\begin{figure}[!t]
    \centering
    \includegraphics[width=0.53\textwidth]{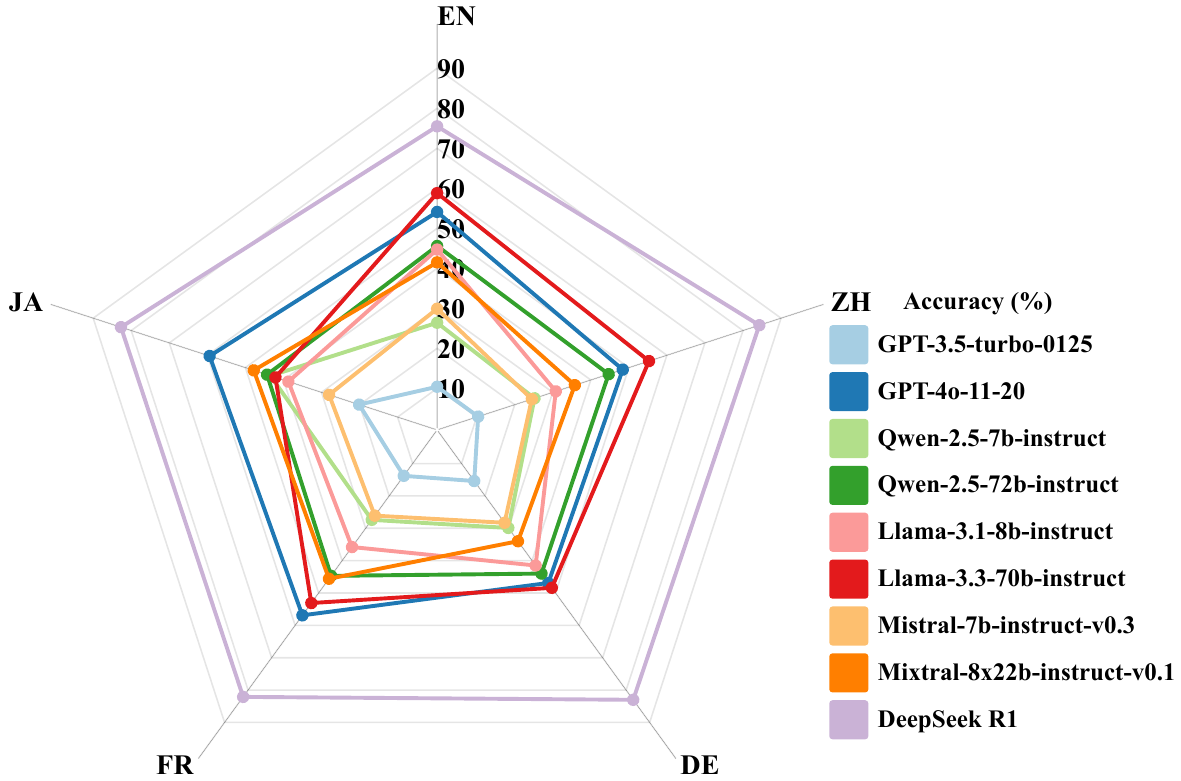}
    \vspace{-0.6cm}
    \caption{LLMs Language Preference in XFANToM.}
    \label{fig:language_preference}
    \vspace{-0.4cm}
\end{figure}

While many state-of-the-art LLMs claim multilingual proficiency, their actual performance varies significantly depending on the language, raising essential questions about their underlying linguistic biases. The experimental results in Figure~\ref{fig:language_preference} on the XFANToM reveal that many LLMs exhibit distinct language preferences, as evidenced by their varying performance across languages. For instance, GPT-3.5-turbo-0125 and GPT-4o-11-20 demonstrate a notable preference for Japanese but lag behind in Chinese, and the performance gap of GPT-4o-11-20 between Japanese and Chinese is 11\%. Interestingly, DeepSeek R1 excels across multiple languages, achieving scores of  84.3\% in Chinese and maintaining high performance in German, French, and Japanese as well, but lagging in English at 75.5\%. Notably, the Qwen models exhibit varied results, with Qwen-2.5-72b-instruct performing uniformly around 45\% across languages, while Qwen-2.5-7b-instruct shows less consistency. The Llama models present a clear preference for English, with Llama-3.3-70b-instruct scoring highest in English at 58.8\% but lagging in Japanese at 42.1\%. These findings emphasize that multilingual performance is highly model-dependent, with some LLMs favoring specific linguistic structures, potentially due to variations in the pre-training corpus, model architecture, and tokenization strategies.
\section{Conclusion}
This work introduces XToM, a high-quality multilingual Theory of Mind benchmark tailored to assess the multilingual ToM ability of current LLMs. We performed comprehensive and detailed experiments to evaluate LLMs' capability on the XToM benchmark and expose critical limitations in LLMs' multilingual reasoning. LLMs are equipped with multilingual understanding ability but fail in multilingual ToM reasoning tasks. 

\section*{Limitations}
\paragraph{Difficulties in Human Annotation}Although XToM already covers three existing benchmarks ranging from the theoretical to the application of the theory of mind, many multifaceted aspects of the theory of mind are not included in our benchmark, such as the Faux Pas test \cite{BaronCohen1999RecognitionOF} and OpenToM \cite{Xu2024OpenToMAC}. Moreover, as the goal of this study is to construct a high-quality multilingual benchmark to explore the multilingual theory of mind for LLMs, the language coverage highly relies on the human annotator, and it poses difficulties to employ sufficiently qualified annotators to cover more languages. 

\paragraph{Potential Contamination}
Most of the existing available benchmarks in the NLP field were released prior to the initiation of the LLM training process, indicating that these datasets are likely to have been utilized during the pre-training phase and post-training phase of LLMs~\cite{DBLP:conf/iclr/GolchinS24, DBLP:conf/aaai/LiF24}. Since ToMi is constructed before the initiation of the LLM training process, the ToMi benchmark may suffer from the potential contamination issue. To ensure the representativeness of XToM, we still have to collect the ToMi for reference purposes, as this dataset is a classical and widely used ToM benchmark. However, for FANToM and NegotiationToM, none of the LLMs’ generated responses matched with the sampled data across various languages, which indicates most instances of the XToM benchmark (i.e., two subtasks XFANToM and XNegotiationToM) are not identified as suffering from the data contamination issue, and the experimental results of the paper are reliable and valuable. Therefore, data contamination may not be a primary concern in XToM. 

\section*{Ethics Statement}
In this work, we conformed to recognized privacy practices and rigorously followed the data usage policy. We declare that all authors of this paper acknowledge the \emph{ACM Code of Ethics} and honor the code of conduct. This paper introduces a multilingual benchmark for assessing the multilingual theory of mind of large language models. We conduct a human evaluation to evaluate strictly the Audience Appropriateness dimension of the MQM framework, and we assign a high value of Severity Multiplier to filter out the offensive context in XToM. 
Therefore, we can foresee no immediate social consequences or ethical issues as we do not introduce social/ethical bias into the model or amplify any bias from the data. 
Moreover, the license of these datasets allows us to modify the data for evaluation and research, and this fulfills their intended use.

\paragraph{Resource Copyright} This work presents a new resource: XTOM, which is a multilingual extension of the FANToM, ToMi, and NegotiationToM, respectively. All three sources of data are publicly available for free, and we do not add any additional requirements for accessing our resources. We will highlight the sources of our data and ask users to cite the original papers when they use our extended versions for research.

\section*{Acknowledgements}
The authors of this paper were supported by the ITSP Platform Research Project (ITS/189/23FP) from ITC of Hong Kong, SAR, China, and the AoE (AoE/E-601/24-N), the RIF (R6021-20) and the GRF (16205322) from RGC of Hong Kong, SAR, China. We also thank the support from NVIDIA AI Technology Center (NVAITC).

\bibliography{anthology,custom}
\bibliographystyle{acl_natbib}

\clearpage

\appendix
\section{Appendix for XTOM}
\label{XTOM}
\subsection{Appendix for Sampling Source Data}
\label{sec:Sampling_Source_Data}
To ensure the quality, diversity, and representativeness of the source data used for dataset construction and to evaluate the multilingual Theory of Mind ability of LLMs, we systematically curated a balanced subset of stories and dialogues from three distinct and well-established benchmarks, ranging from theoretical to application of the Theory of Mind (ToM). These three well-established Theory of Mind datasets are ToMi~\cite{DBLP:conf/emnlp/LeBN19}, FANToM~\cite{DBLP:conf/emnlp/0002SZ0K0S23}, and NegotiationToM~\cite{DBLP:conf/emnlp/ChanJYDF0L0WS24}, utilized to create three subsets of XToM (i.e., XToMi, XFANToM, and XNegotiationToM). For the ToMi dataset, we randomly selected 300 stories, including first-order false belief, second-order false belief, and reality question categories, yielding a total of 647 questions. For the FANToM dataset, we similarly sampled 300 stories, prioritizing the short conversation version due to its manageable dialogue length (averaging 13.8 turns per conversation) compared to the longer version. This subset encompassed 618 questions, including fact questions (FACTQ), belief questions (BELIEFQ), and information accessibility questions (INFOACCESSQ). Finally, from the NegotiationToM dataset, we randomly selected 300 real-world negotiation dialogues, each accompanied by a fixed set of questions probing the beliefs, desires, and intentions of both agents, resulting in 1,758 questions. This systematic sampling approach not only preserves the theoretical richness of the dataset but also ensures a comprehensive evaluation of ToM reasoning capabilities in LLMs across varying levels of cognitive complexity.

\subsection{Appendix for Human Annotator Qualification}
\label{sec:Human_Annotator_Qualification}
In the construction process of XToM, all human annotators are required to manually conduct the human annotation and evaluation of the translated instances. To ensure high-quality bilingual annotations and evaluation, we implemented a rigorous annotator qualification protocol. We recruited a total of 12 annotators, with three qualified human annotators for each language in the human annotation and evaluation process. All annotators were required to be native speakers of one target language while demonstrating certified proficiency in English through standardized assessments. For instance, annotators working with the Chinese version were native Chinese speakers who possessed either TOEFL or IELTS certification, demonstrating advanced English language competency. Prior to full-scale annotation, we conducted a preliminary quality assurance phase where annotators completed a pilot set of 20 instances. This initial phase served dual purposes: it allowed us to evaluate annotator reliability and provided an opportunity to address typical errors through targeted feedback, thereby standardizing the annotation process across all participants.

\subsection{Appendix for Multidimensional Quality Metrics (MQM) Framework}
\label{mqm}
\begin{table*}[htbp!]
    \centering
    \renewcommand{\arraystretch}{1.5}
    \scalebox{0.8}{
    \begin{tabular}{|c|l|l|}
        \hline
        \textbf{Step} & \textbf{Overall Quality Score Calculation} & \textbf{Formulas} \\
        \hline
        1 & Absolute Penalty Total (APT) & 
        $
        \sum_{i,j} \text{Error Count}_{ij} \times \text{Severity Multiplier}_j \times \text{Error Type Weight}_i
        $
        \\ 
        & & Where: $i$ = index for Error Types, $j$ = index for Severity Level. \\
        \hline
        2 & Per-Word Penalty Total (PWPT) & 
        $
        \frac{\text{Absolute Penalty Total}}{\text{Evaluation Word Count}}
        $ 
        \\ 
        \hline
        3 & Overall Quality Score (OQS) & 
        $
        (1 - \text{PWPT}) \times \text{Maximum Score Value}
        $
 \\
        \hline
    \end{tabular}
    }
    \caption{Overall Quality Score Calculation}
    \label{tab:RQS}
\end{table*}

\begin{CJK*}{UTF8}{gbsn}
\begin{table*}[!] 
\small
\centering
\setlength\tabcolsep{3pt}
\scalebox{0.7}{
\begin{tabular}{ll}
\toprule
\textbf{Dataset}         & \textbf{Prompt Template} \\ 
\midrule
EN & Give a \textbf{\texttt{\textcolor{headcolor}{\{ToMi/FanToM/NegotiationToM\}}}} Test example and its answer. \\
\midrule
ZH & 给出一个\textbf{\texttt{\textcolor{headcolor}{\{ToMi/FanToM/NegotiationToM\}}}}测试示例及其答案。\\
\midrule
DE & Geben Sie ein Beispiel für den \textbf{\texttt{\textcolor{headcolor}{\{ToMi/FanToM/NegotiationToM\}}}}-Test und die dazugehörige Antwort. \\ 
\midrule
FR &  Donnez un exemple de test \textbf{\texttt{\textcolor{headcolor}{\{ToMi/FanToM/NegotiationToM\}}}} et sa réponse.\\
\midrule
JA &  \textbf{\texttt{\textcolor{headcolor}{\{ToMi/FanToM/NegotiationToM\}}}} テストの例とその解答を示します。\\ 
\midrule 
\midrule

\multirow{2}{*}{EN} & You are provided with Sentence 1 from the train split of the \textbf{\texttt{\textcolor{headcolor}{\{ToMi/FanToM/NegotiationToM\}}}} dataset. Finish Sentence 2 as appeared in the dataset. \\ 
&  Sentence 2 must exactly match the instance in the dataset.Sentence 1: \textbf{\texttt{ \textcolor{tailcolor}{<Sentence1-of-Example>} }}\\

\midrule

\multirow{2}{*}{ZH} & 你将获得来自 \textbf{\texttt{\textcolor{headcolor}{\{ToMi/FanToM/NegotiationToM\}}}} 数据集训练集的句子. 1.请按照数据集中显示的方式完成句子. 2.句子 2 必须与数据集中的实例\\
&  完全匹配.句子 1：\textbf{\texttt{ \textcolor{tailcolor}{<Sentence1-of-Example>}}}\\ 

\midrule

\multirow{2}{*}{DE} & Sie erhalten Satz 1 aus dem Train-Split des Datensatzes \textbf{\texttt{\textcolor{headcolor}{\{ToMi/FanToM/NegotiationToM\}}}}. Vervollständigen Sie Satz 2 wie im Datensatz angegeben.\\ 
& Satz 2 muss exakt mit der Instanz im Datensatz übereinstimmen.Satz 1: \textbf{\texttt{ \textcolor{tailcolor}{<Sentence1-of-Example>}}}. \\ 

\midrule

\multirow{2}{*}{FR} & Vous disposez de la phrase 1 issue de la division du train du jeu de données \textbf{\texttt{\textcolor{headcolor}{\{ToMi/FanToM/NegotiationToM\}}}}. Complétez la phrase 2 telle qu'elle apparaît\\
& dans le jeu de données. La phrase 2 doit correspondre exactement à l'instance du jeu de données.Phrase 1: \textbf{\texttt{ \textcolor{tailcolor}{<Sentence1-of-Example>}}}.\\ 

\midrule

\multirow{2}{*}{JA} &  文1は\textbf{\texttt{\textcolor{headcolor}{\{ToMi/FanToM/NegotiationToM\}}}}データセットの学習済みデータから提供されています。文2はデータセットに出現した通りに完成させてください。\\& 文2はデータセット内のインスタンスと完全に一致する必要があります。文1：\textbf{\texttt{ \textcolor{tailcolor}{<Sentence1-of-Example>}}}\\
\bottomrule
\end{tabular}
}
\caption{Prompt Template for Verification of Potential Contamination by following~\citet{DBLP:conf/aaai/LiF24} and~\citet{DBLP:conf/iclr/GolchinS24} }
\label{tab:Appendix_Potential_Contamination_pormpt_template}
\end{table*}
\end{CJK*}

The \textit{Multidimensional Quality Metrics} (MQM) framework is a structured and flexible system for assessing translation quality. It defines a set of dimensions that categorize different aspects of translation errors, allowing for precise and customizable evaluations. The core dimensions of MQM include:

\begin{itemize}
    \item \textbf{Terminology}: Ensures the correct and consistent use of terms. Errors include incorrect terminology usage, inconsistencies, and deviations from domain-specific terminology guidelines.
    
    \item \textbf{Accuracy}: Evaluates whether the translation correctly conveys the meaning of the source text. Issues include mistranslation, overtranslation, undertranslation, addition, omission, and untranslated content.
    
    \item \textbf{Linguistic Conventions}: Covers grammatical correctness and linguistic coherence, including errors in grammar, punctuation, spelling, and unintelligibility.
    
    \item \textbf{Style}: Assesses adherence to appropriate tone and register. This includes language register, awkward phrasing, unidiomatic expressions, and inconsistent style. We place particular importance on avoiding unidiomatic expressions in translations.
    
    \item \textbf{Locale Conventions}: Ensures compliance with region-specific norms, such as number format, measurement format, time format, date format, address format, telephone format, and shortcut keys.
    
    \item \textbf{Audience Appropriateness}: Examines whether the translation is suitable for its target audience, considering readability, domain-specific terminology, and adaptation to audience expectations. In our case, we particularly emphasize culture-specific references and the avoidance of offensive content.
    
    \item \textbf{Design and Markup}: Focuses on structural and formatting aspects, identifying errors such as incorrect text formatting, misplaced tags, broken links, and layout inconsistencies.
    
    \item \textbf{Custom}: Allows for the definition of additional dimensions tailored to specific translation tasks or industry requirements, ensuring adaptability for specialized evaluations. In our scenario, we ask annotators to ensure that object labels are translated consistently across the answer, question, and story within a single data instance. In NegotiationToM, we focus on maintaining consistency in the translation of water, food, firewood, and intention labels. In ToMi, we emphasize the consistency of objects mentioned in both the story and the answer, particularly the translation of containers.
\end{itemize}

\begin{figure*}
    \centering
    \includegraphics[width=\linewidth]{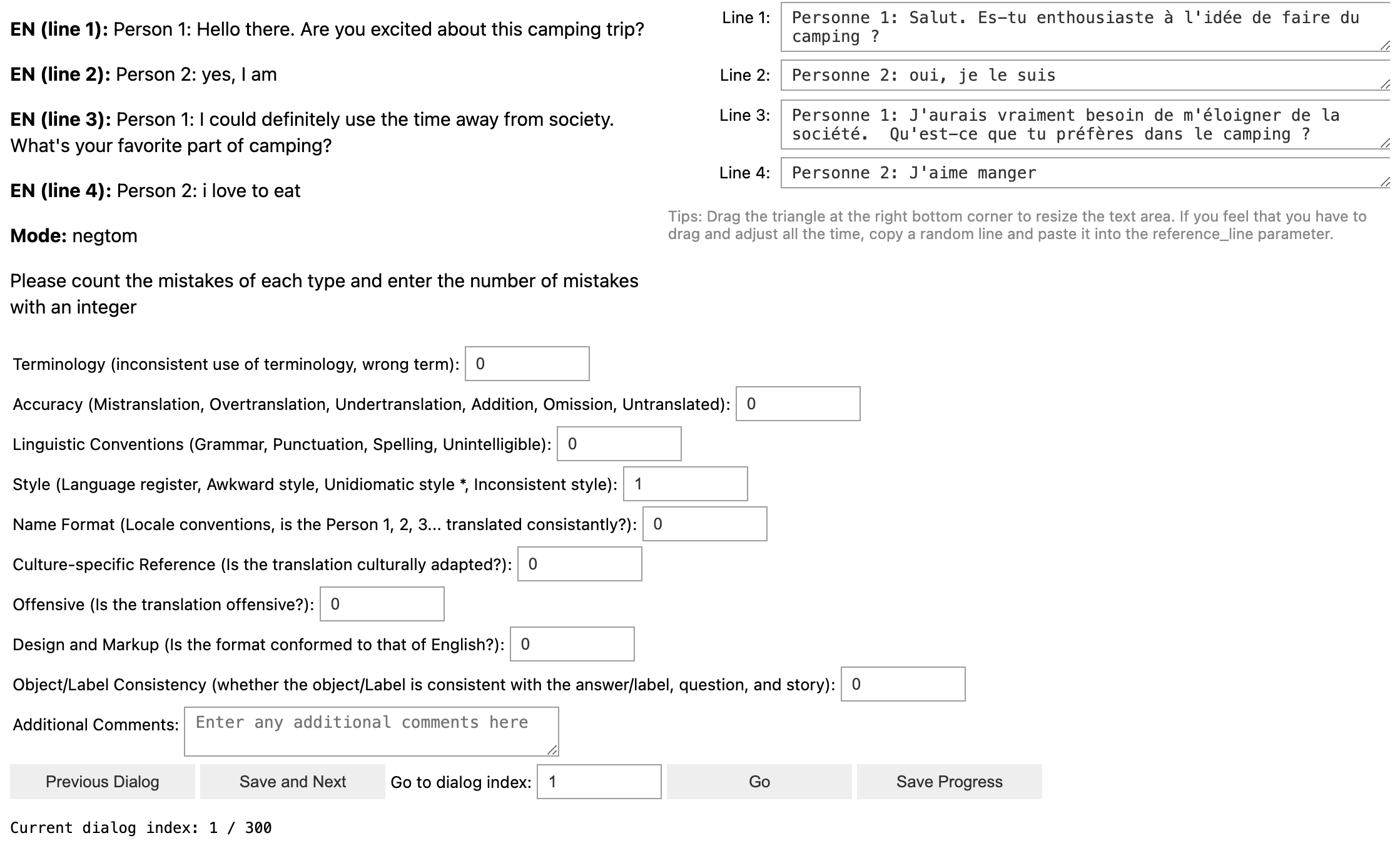}
    \caption{Interface for Human Correction and Validation}
    \label{fig:enter-label}
\end{figure*}

In practice, annotators record the number of errors in each category for each data sample, and the MQM score is calculated using the scoring method outlined in Table \ref{tab:RQS}. Maximum Score Value represents a perfect upper score on a scale, and the default value is one hundred. According to the MQM framework, we set the preset threshold score as 95 to determine whether a translated instance passes or fails.
By categorizing translation quality into these dimensions, MQM provides a systematic and adaptable approach to evaluating both human and machine translations. This structured framework enhances the reliability and consistency of translation quality assessment across diverse use cases. To streamline the annotation process, we developed a software interface (shown in Figure \ref{fig:enter-label}) to assist annotators. The code will be released alongside the dataset.

\begin{table}[!] 
\small
\centering
\setlength\tabcolsep{3pt}
\scalebox{0.9}{
\begin{tabular}{lccc}
\toprule
Dataset         & XFANToM       & XNegotiationToM      & XToMi \\ 
\midrule  
Deepseek R1(EN) & 0 &   0 &     30 \\
Deepseek R1(ZH)	& 0 & 	0 & 	33 \\ 
Deepseek R1(DE) & 0 & 	0 & 	23 \\ 
Deepseek R1(FR) & 0 & 	0 & 	25 \\ 
Deepseek R1(JA) & 0 & 	0 & 	18 \\ 
\midrule  
GPT 4o (EN) & 	0 & 	0 & 	36 \\ 
GPT 4o (ZH) & 	0 & 	0 & 	34 \\ 
GPT 4o (DE) & 	0 & 	0 & 	25 \\
GPT 4o (FR) & 	0 & 	0 & 	9 \\ 
GPT 4o (JA) & 	0 & 	0 & 	13 \\ 
\bottomrule
\end{tabular}
}

\caption{Verification of Potential Contamination in 100 Independent Trials by following~\citet{DBLP:conf/aaai/LiF24}}
\label{tab:Appendix_Potential_Contamination_1}
\end{table}

\begin{table}[!] 
\small
\centering
\setlength\tabcolsep{3pt}
\scalebox{0.9}{
\begin{tabular}{lccc}
\toprule
Dataset         & XFANToM       & XNegotiationToM      & XToMi \\ 
\midrule  
Deepseek R1(EN) & 0 &   0 &     10 \\
Deepseek R1(ZH)	& 0 & 	0 & 	4 \\ 
Deepseek R1(DE) & 0 & 	0 & 	0 \\ 
Deepseek R1(FR) & 0 & 	0 & 	0 \\ 
Deepseek R1(JA) & 0 & 	0 & 	0 \\ 
\midrule  
GPT 4o (EN) & 	0 & 	0 & 	12 \\ 
GPT 4o (ZH) & 	0 & 	0 & 	10 \\ 
GPT 4o (DE) & 	0 & 	0 & 	2 \\
GPT 4o (FR) & 	0 & 	0 & 	5 \\ 
GPT 4o (JA) & 	0 & 	0 & 	8 \\ 
\bottomrule
\end{tabular}}
\caption{Verification of Potential Contamination in 100 unique stories by following~\citet{DBLP:conf/iclr/GolchinS24}}
\label{tab:Appendix_Potential_Contamination_2}
\end{table}

\subsection{Verification of Potential Contamination} \label{Verification_Contamination}
Most of the existing available benchmarks in the NLP field were released prior to the initiation of the LLM training process, indicating that these datasets are likely to have been utilized during the pre-training phase and post-training phase (i.e., SFT~\cite{DBLP:conf/nips/Ouyang0JAWMZASR22} or RLHF~\cite{DBLP:conf/nips/ChristianoLBMLA17}) of LLMs~\cite{DBLP:conf/iclr/GolchinS24, DBLP:conf/aaai/LiF24}. Therefore, we follow the established protocols by prior works~\cite{DBLP:conf/iclr/GolchinS24, DBLP:conf/aaai/LiF24} to tailor two prompting templates in each language to assess the potential contamination issues in the sampled dataset. The prompt template is shown in Table~\ref{tab:Appendix_Potential_Contamination_pormpt_template}. For each sampled dataset, we conducted 100 independent trials or 100 unique stories per dataset using the prompting methods in~\citet{DBLP:conf/iclr/GolchinS24} and~\citet{DBLP:conf/aaai/LiF24} by utilizing state-of-the-art LLMs (DeepSeek R1 and GPT-4o). These trials aimed to detect overlaps between model-generated outputs (stories, questions, answers) and the benchmark’s ground-truth data. All responses underwent a stringent human evaluation to verify contamination issues in existing datasets. The results are reported in the following tables~\ref{tab:Appendix_Potential_Contamination_1} and~\ref{tab:Appendix_Potential_Contamination_2}. For FANToM and NegotiationToM, none of the LLMs’ generated responses matched with the ground truth by using two prompting methods across various languages, which indicates most instances of the XToM benchmark (i.e., two subtasks XFANToM and XNegotiationToM) are not identified as suffering from the data contamination issue, and the experimental results of the paper are reliable and valuable. Therefore, we believe data contamination may not be a primary concern in XToM. An interesting finding is that some LLM-generated responses match some ToMi story patterns or even data instances. However, to ensure the representativeness of the dataset, we still have to collect the ToMi for reference purposes, as ToMi is a classical ToM benchmark in the ToM field.

\begin{table}[!] 
\small
\centering
\setlength\tabcolsep{4pt}
\scalebox{0.7}{
\begin{tabular}{lcccccc}
\toprule
Dataset  & Total\#Questions    & BeliefQA    & 1st.Belief & 2nd.Belief & FactQA\\ 
\midrule  
XFANToM  & 3,090       & 1,590   & 1,085  & 505  & 1,500 \\ 
\midrule 
\midrule 
Dataset  & Total\#Questions    & Belief    & 1st.Belief & 2nd.Belief & Reality\\ 
\midrule  
XToMi    &   3,235     &   1,735     &   615     &   1,120    & 1,500 \\ 
\midrule 
\midrule 
Dataset  & Total\#Questions    & Belief  & Desire  & Intention  & -\\
\midrule 
XNegotiationToM & 8,790    & 2,930         & 2,930       & 2,930  & -\\ 
\bottomrule
\end{tabular}
}
\caption{Statistics of XTOM, which include XFANToM, XToMi, and XNegotiationToM}
\label{tab:dataset_stats}
\end{table}

\begin{table}[!] 
\small
\centering
\setlength\tabcolsep{3pt}
\begin{tabular}{lccccc}
\toprule
Dataset         & CH        & DE      & JA     & FR     & Avg.Dataset\\ 
\midrule  
XFANToM         & 96.59     & 93.63   & 95.11  & 96.55  & 96.59\\ 
\midrule  
XToMi           & 94.62     & 93.98   & 95.78  & 92.20  &  94.15\\ 
\midrule 
XNegotiationToM &  95.62    & 97.92   & 95.61  & 94.63  & 96.27\\
\midrule 
Avg.Language     & 95.61     & 97.18   & 95.78  & 95.46  & 95.18 \\
\bottomrule
\end{tabular}
\caption{Fleiss Kappa of XTOM across various tasks and languages.}
\label{tab:kappa}
\end{table}

\section{Appendix for Related Works}\label{Appendix_for_Related_Works}
\subsection{Related Works for Theory of Mind}\label{sec:Related_Works_for_Theory_of_Mind}
In recent years, benchmarks for evaluating large language models' Theory of Mind capabilities have become a crucial area of artificial intelligence research. The roots of this work can be traced back to classic psychological experiments, such as the Sally-Anne test \cite{BaronCohen1985DoesTA}, and have continually evolved to explore more complex aspects of ToM. Early significant contributions include the ToM-bAbi dataset \cite{Grant2017HowCM}, which focused on assessing false beliefs and was later improved by \citet{DBLP:conf/emnlp/LeBN19} into the more comprehensive ToMi dataset. These early works laid the foundation for subsequent research, prompting the emergence of more advanced benchmarks.

Researchers developed a series of more complex evaluation tools based on this foundation. For instance, T4D \cite{Zhou2023HowFA} specifically assesses N-ToM capabilities in AI assistants, while Hi-ToM \cite{He2023HITOMAB} explores higher-order N-ToM concepts. These works significantly expanded our understanding of AI systems' ToM capabilities. Simultaneously, researchers also sought inspiration from classic human ToM tests, such as the Smarties test \cite{Gopnik1988ChildrensUO} and the Faux Pas test \cite{BaronCohen1999RecognitionOF}. These efforts led to the creation of multiple new datasets, including ToMChallenges \cite{Ma2023ToMChallengesAP}, BigToM \cite{Gandhi2023UnderstandingSR}, and FauxPas-EAI \cite{Shapira2023HowWD}, each targeting different aspects of N-ToM assessment.

Recognizing the need for more comprehensive evaluation methods, \citet{Jones2024ComparingHA} proposed EPITOME, a comprehensive benchmark that integrates various human ToM tests. Furthermore, researchers focused on assessing N-ToM capabilities in dialogue contexts, leading to benchmarks such as G-DRAGON \cite{Zhou2022ICD}, FANToM \cite{DBLP:conf/emnlp/0002SZ0K0S23}, and SOTOPIA \cite{Zhou2023SOTOPIAIE}. These works greatly enriched our methods for evaluating AI systems' ToM capabilities.

More recently, several new benchmarks have emerged to address limitations in previous ToM evaluations, offering more realistic and challenging scenarios. TOMBench\cite{Chen2024ToMBenchBT} provides a comprehensive framework with 31 social cognition abilities across 8 tasks. At the same time, OpenToM \cite{Xu2024OpenToMAC} focuses on natural narratives featuring characters with distinct personalities and intentions, testing models on psychological and physical mental states. MMToM-QA \cite{Jin2024MMToMQAMT} introduces multimodal evaluation, combining video and text inputs to assess ToM reasoning, and NegotiationToM \cite{DBLP:conf/emnlp/ChanJYDF0L0WS24} explores multi-dimensional mental states in real-world negotiation dialogues.

\subsection{Related Works for Multilingual Capabilities of LLMs}\label{sec:Related_Works_for_Multilingual_Capabilities_of_LLMs}
Recent studies have thoroughly evaluated instruction-following large language models (LLMs)~\cite{DBLP:journals/corr/abs-2303-08774, openai2022chatgpt, DBLP:conf/emnlp/JiangCCW23, DBLP:conf/ijcnlp/ChanLCCSWS23}, demonstrating their superior zero-shot performance across numerous tasks~\cite{
DBLP:journals/corr/abs-2303-12712,
DBLP:conf/eacl/ChanCWJFLS24,
DBLP:conf/emnlp/ChengQCFWCRGZSZ23,
DBLP:conf/acl/0001FLS0XWBLJCS24,
DBLP:conf/emnlp/JiayangCZQZLS0L24,
shi2025inferencedynamicsefficientroutingllms,
DBLP:conf/coling/JiayangQC0SZ24}. However, significant challenges remain unaddressed, including complex mathematical~\cite{DBLP:journals/corr/abs-2301-13867} and theory of mind reasoning~\cite{DBLP:conf/pricai/LinCSL24}, analogical reasoning~\cite{DBLP:conf/emnlp/ChengQCFWCRGZSZ23}, discourse relation classification~\cite{DBLP:conf/acl/ChanLCLSWS23}, text-to-table generation~\cite{DBLP:journals/corr/abs-2404-14215}, complex game scenarios~\cite{DBLP:journals/corr/abs-2408-02559}, argument impact classification~\cite{DBLP:conf/ecai/ChanCLYJD0SWS24}, and associated ethical and privacy concerns~\cite{DBLP:journals/corr/abs-2310-10383,DBLP:journals/corr/abs-2212-09292,DBLP:conf/acl/0003GLFH0CYYS24,DBLP:journals/corr/abs-2302-00539,DBLP:journals/corr/abs-2405-07667}. Therefore, it is crucial to investigate whether large language models possess the theory of mind capabilities across languages.

State-of-the-art large language models (LLMs), such as GPT-4o-11-20 \cite{DBLP:journals/corr/abs-2410-21276}, LLaMA \cite{DBLP:journals/corr/abs-2407-21783}, Mistral \cite{jiang2023mistral7b}, and DeepSeek \cite{deepseekai2025deepseekr1incentivizingreasoningcapability}, are often described as multilingual. They are pre-trained on a mixture of texts in multiple languages, leveraging linguistic similarities and shared representations \cite{zeng-etal-2025-converging, wendler2024llamasworkenglishlatent, dumas2025separatingtonguethoughtactivation} to enhance performance in lower-resource languages. Recent studies have demonstrated that LLMs can develop cross-lingual capabilities, transferring knowledge and skills learned in one language to others, even those with limited training data \citep{chirkova2024zeroshot, pires-etal-2019-multilingual, wu-dredze-2019-beto}. However, a significant performance gap persists across languages, and researchers are actively exploring methods to quantify and mitigate this disparity \cite{li2024languagerankermetricquantifying, kumar2024bridginggapdynamiclearning, zeng2024multilingualbrainsurgeonlarge}.

Multilingual models have exhibited strong reasoning abilities across languages, including those that are underrepresented \cite{shi2022languagemodelsmultilingualchainofthought}. For instance, \citet{qin2023crosslingualpromptingimprovingzeroshot} proposed prompting models with chain-of-thought reasoning to solve tasks in multiple languages, then ensembling reasoning paths across languages to derive more accurate answers. Similarly, \citet{she2024mapoadvancingmultilingualreasoning} introduced preference optimization techniques to align reasoning processes in non-dominant languages with those in dominant ones, thereby enhancing multilingual reasoning capabilities. Nevertheless, while LLMs demonstrate promising surface-level cross-lingual abilities in tasks such as machine translation and embedding space alignment, they still face challenges in deeper cross-lingual knowledge transfer \cite{chua2024crosslingualcapabilitiesknowledgebarriers}.

\section{Appendix For Experimental Setting}\label{sec:appendix_For_Experimental_Setting}
\subsection{Hyperparameter} \label{sec:hyparameter}
We use the following hyperparameters to assess the large language models mentioned in this paper. 
For ChatGPT (\texttt{gpt-3.5-turbo-0125})  and GPT-4o (\texttt{gpt-4o-1120}), the default parameters\footnote{https://platform.openai.com/docs/api-reference/chat/create} are \texttt{temperature}=1 and \texttt{top\_p}=1.
For Llama-3.1-8b-instruct and Llama-3.3-70b-instruct models, we follow the default setting where \texttt{temperature}=0.7, \texttt{top\_p}=0.9.
For DeepSeek R1, the parameters are \texttt{temperature}=0.5, \texttt{top\_p}=0.7. For Qwen-3-235b-a22b, Qwen-2.5-7b-instruct and Qwen-2.5-72b-instruct, the parameters are \texttt{temperature}=0.5, \texttt{top\_p}=0.7.
For mistral-7b-instruct-v0.3 and mixtral-8x22b-instruct-v0.1, the parameters are \texttt{temperature}=0.2, \texttt{top\_p}=0.7.

\subsection{Appendix for Prompt Template} \label{sec:Prompt_Template}
In this study, there are two types of prompting methods. The first one is zero-shot prompting, and we utilize the prompt template in the original paper. Another one is the chain of thought (CoT) prompting method, which follows~\citet{DBLP:conf/nips/Wei0SBIXCLZ22} and use the prompt “let’s think step by step.” The zero-shot prompt template and data example for XNegotiationToM are presented in Tables~\ref{tab:baseline-prompt-template}, \ref{tab:baseline-prompt-template-zh}, \ref{tab:baseline-prompt-template-de}, \ref{tab:baseline-prompt-template-fr}, and \ref{tab:baseline-prompt-template-ja}. The zero-shot prompt template and data example of XToMi are illustrated in~\ref{tab:tomi-lang-all} while XFANToM is shown in Table~\ref{tab:FANToM-lang-all}. The COT prompting template is formed by appending the “let’s think step by step” to the zero-shot prompt template.

\begin{figure*}[!t]
    \centering
    \vspace{-1cm}
    \includegraphics[width=\textwidth]{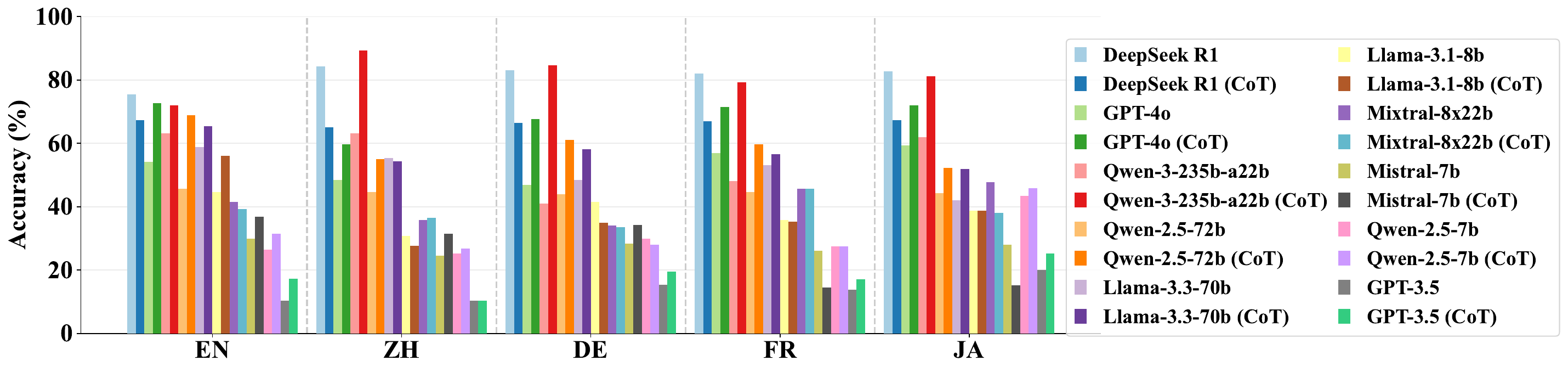}
    \vspace{-0.8cm}
    \caption{Performance comparison of different models on false belief questions across languages in XFANToM.}
    \vspace{-0.5cm}
    \label{fig:Appendix_belief_choice_fantom}
\end{figure*}

\begin{figure*}[!t]
    \centering
    \includegraphics[width=\textwidth]{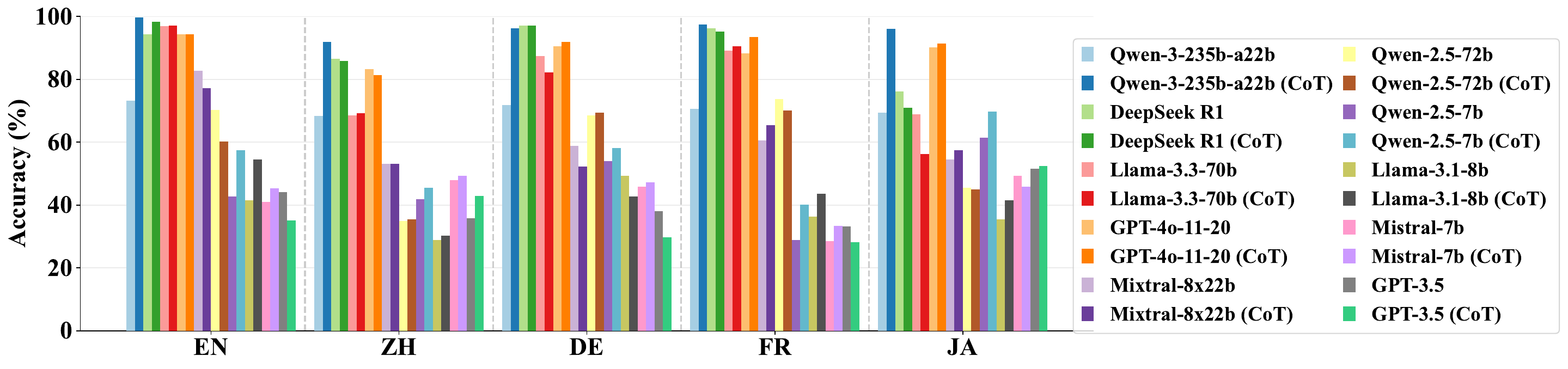}
    \vspace{-0.8cm}
    \caption{Performance comparison of different models on false belief questions across languages in XToMi.}
    \vspace{-0.5cm}
    \label{fig:Appendix_belief_choice_tomi}
\end{figure*}

\begin{figure*}[!t]
    \centering
    \includegraphics[width=\textwidth]{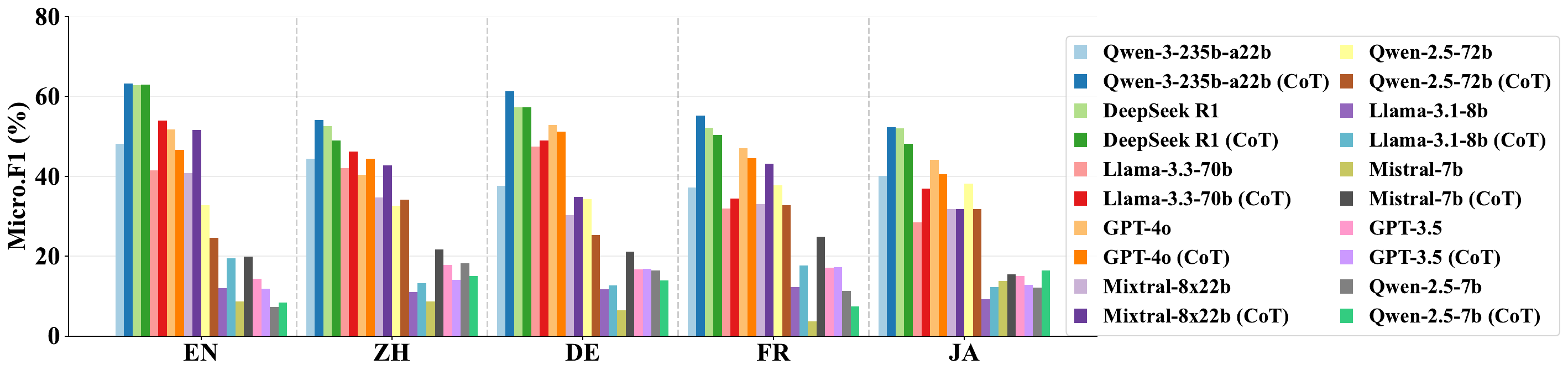}
    \vspace{-0.8cm}
    \caption{Performance comparison of different models on belief choice across languages in XNegotiationToM.}
    \vspace{-0.3cm}
    \label{fig:Appendix_belief_choice_negtom}
\end{figure*}

\begin{figure*}[!t]
    \centering
    \includegraphics[width=\textwidth]{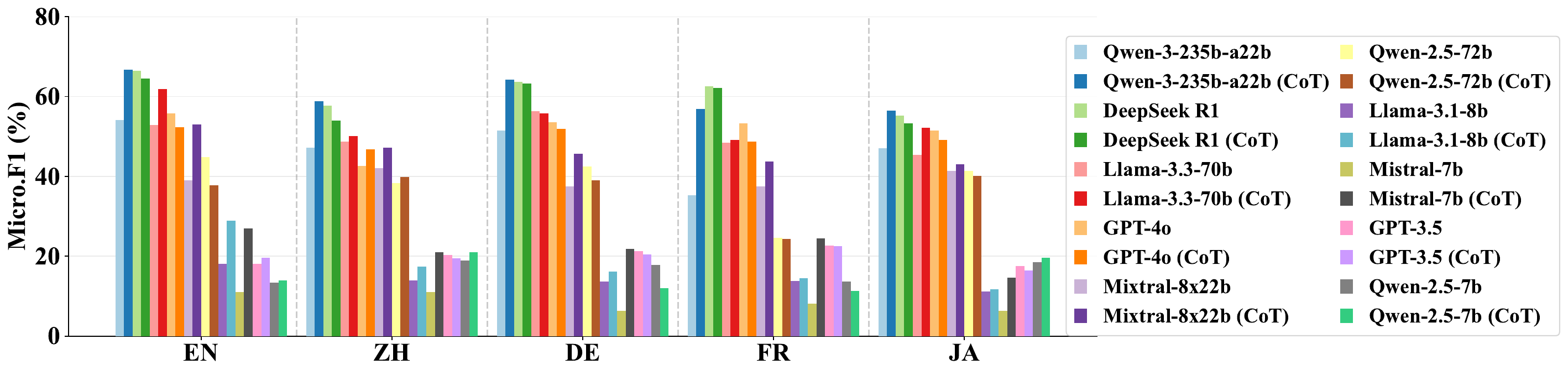}
    \vspace{-0.6cm}
    \caption{Performance comparison of different models on desire choice across languages in XNegotiationToM.}
    \label{fig:desire_choice_negtom}
\end{figure*}

\begin{figure*}[!t]
    \centering
    \includegraphics[width=\textwidth]{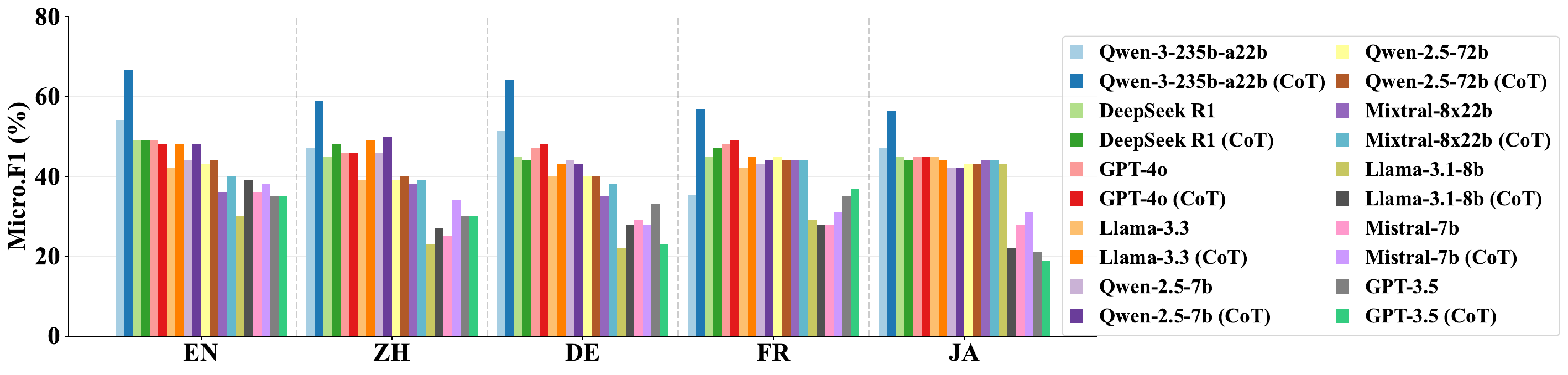}
    \vspace{-0.6cm}
    \caption{Performance comparison of different models on intention choice across languages in XNegotiationToM.}
    \label{fig:intention_choice_negtom}
\end{figure*}

\begin{figure*}[!t]
    \centering
    \includegraphics[width=\textwidth]{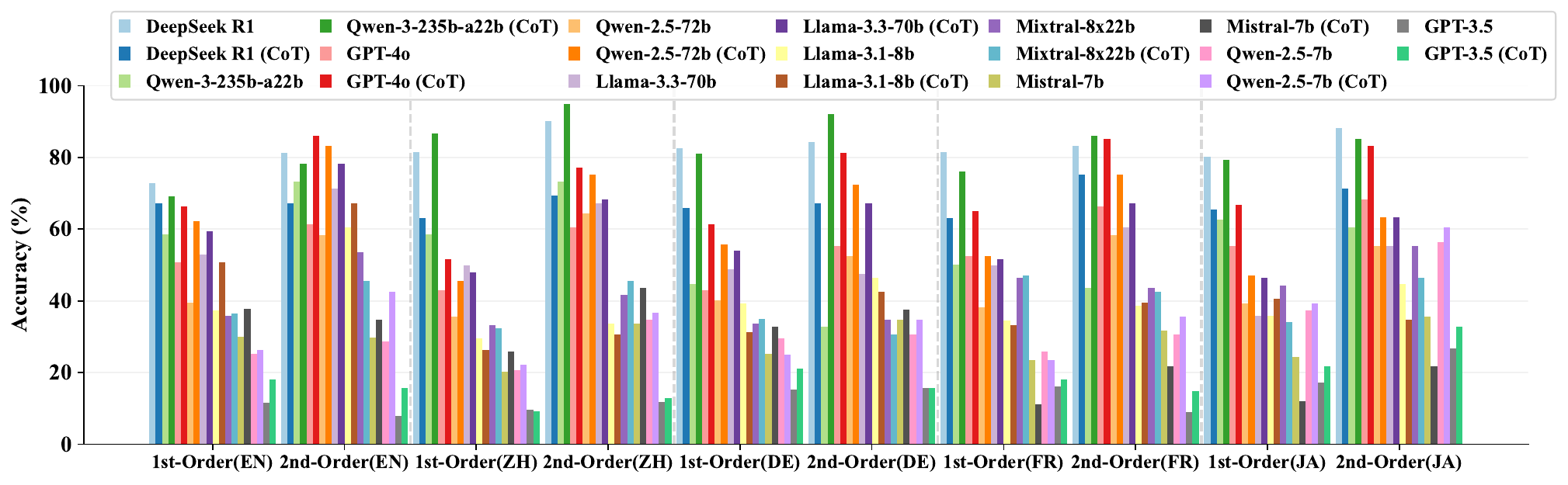}
    \vspace{-0.6cm}
    \caption{Performance comparison of different models on first order and second order belief choice across languages in XFANToM.}
    \label{fig:Appendix_1and2order_choice_FANToM}
\end{figure*}

\begin{figure*}[!t]
    \centering
    \includegraphics[width=\textwidth]{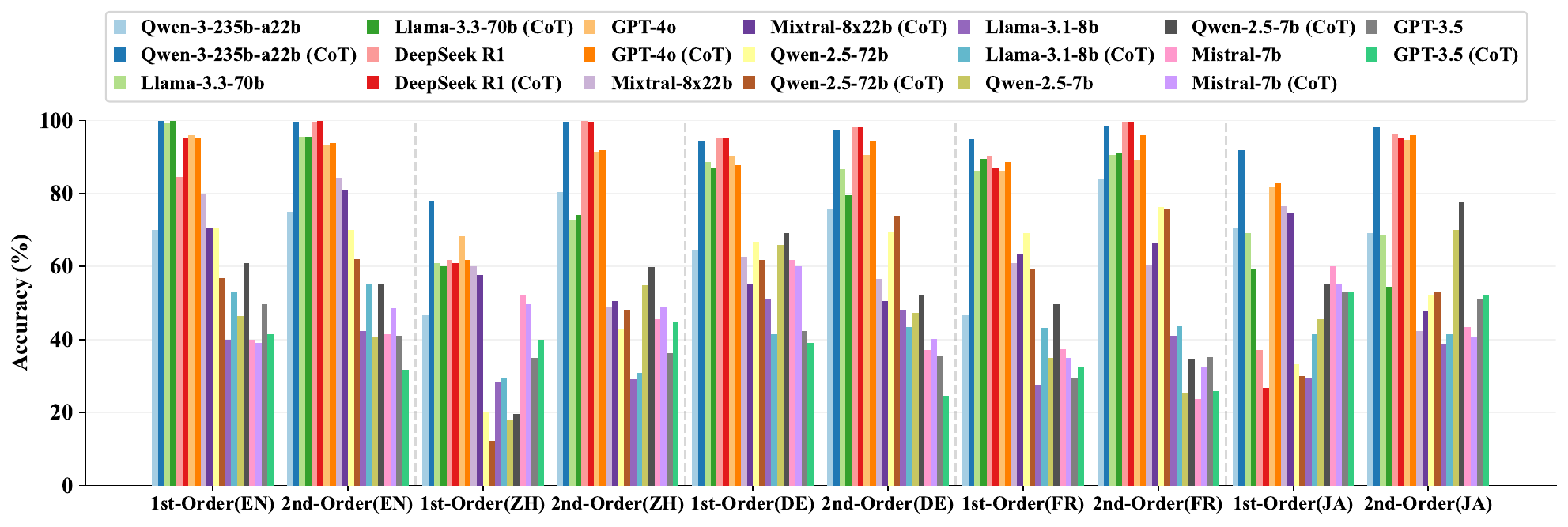}
    \vspace{-0.6cm}
    \caption{Performance comparison of different models on first order and second order belief choice across languages in XToMi.}
    \label{fig:Appendix_1and2order_choice_ToMi}
\end{figure*}

\begin{figure}[htbp]
    \vspace{-1cm}
    \centering
    \begin{subfigure}{0.45\textwidth}
        \centering
        \includegraphics[width=\textwidth]{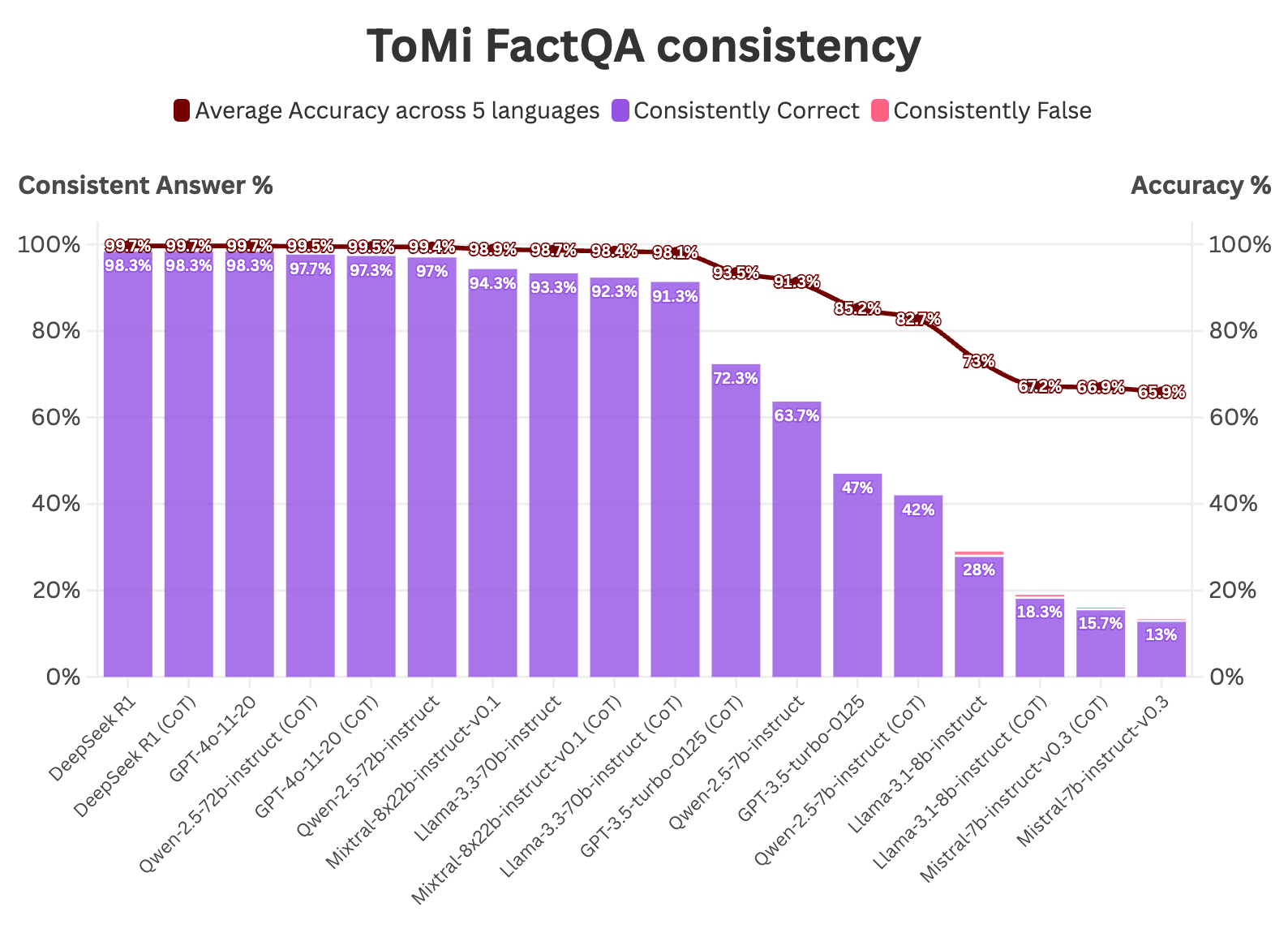}
        \caption{}
    \end{subfigure}
    \hfill
    \begin{subfigure}{0.45\textwidth}
        \centering
        \includegraphics[width=\textwidth]{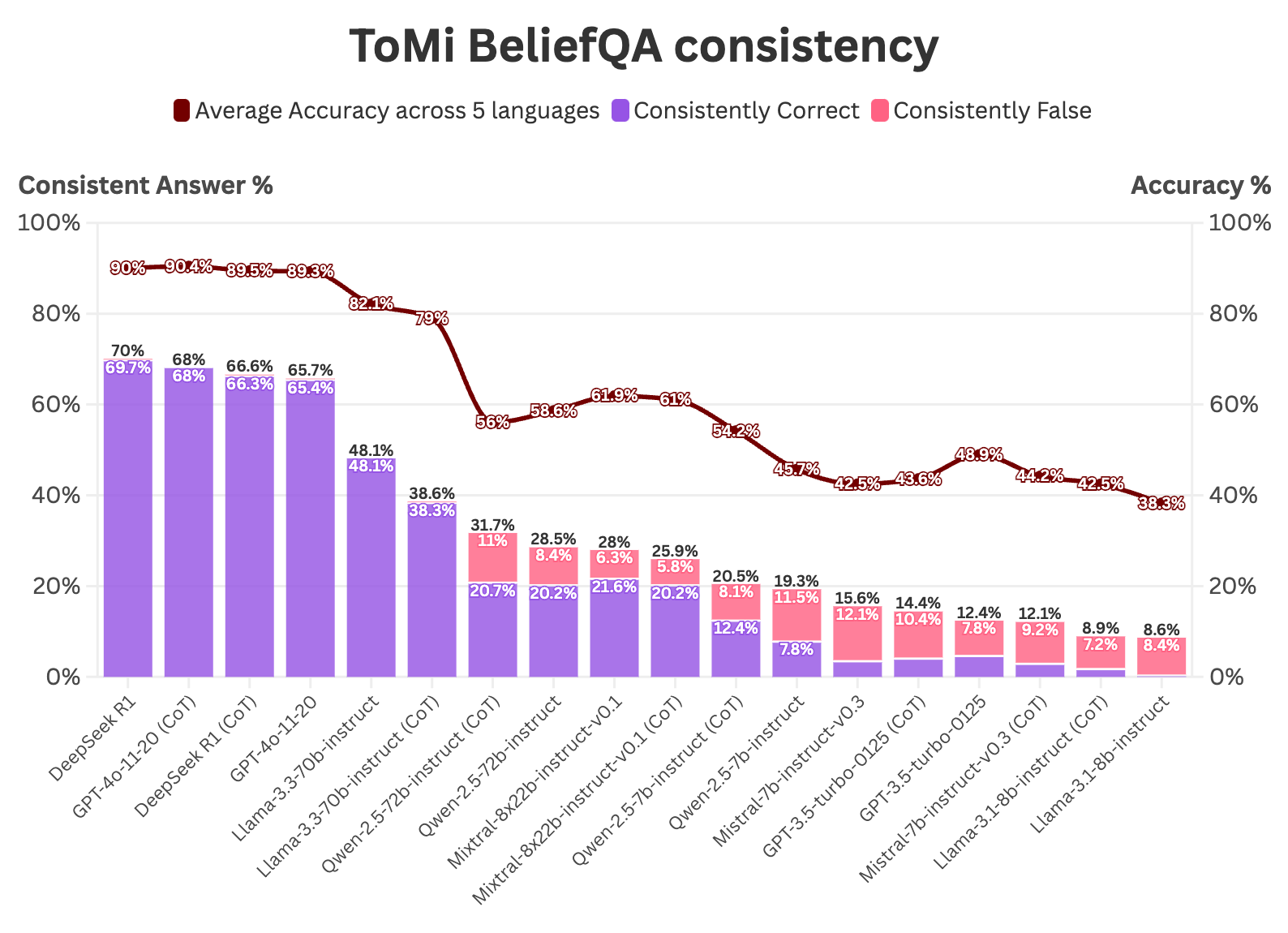}
        \caption{}
    \end{subfigure}
    \caption{ToMi consistency analysis across different languages}
    \label{fig:tomi_consistency}
    \vspace{-0.6cm}
\end{figure}


\begin{figure}[!t]
    \centering
    \includegraphics[width=0.5\textwidth]{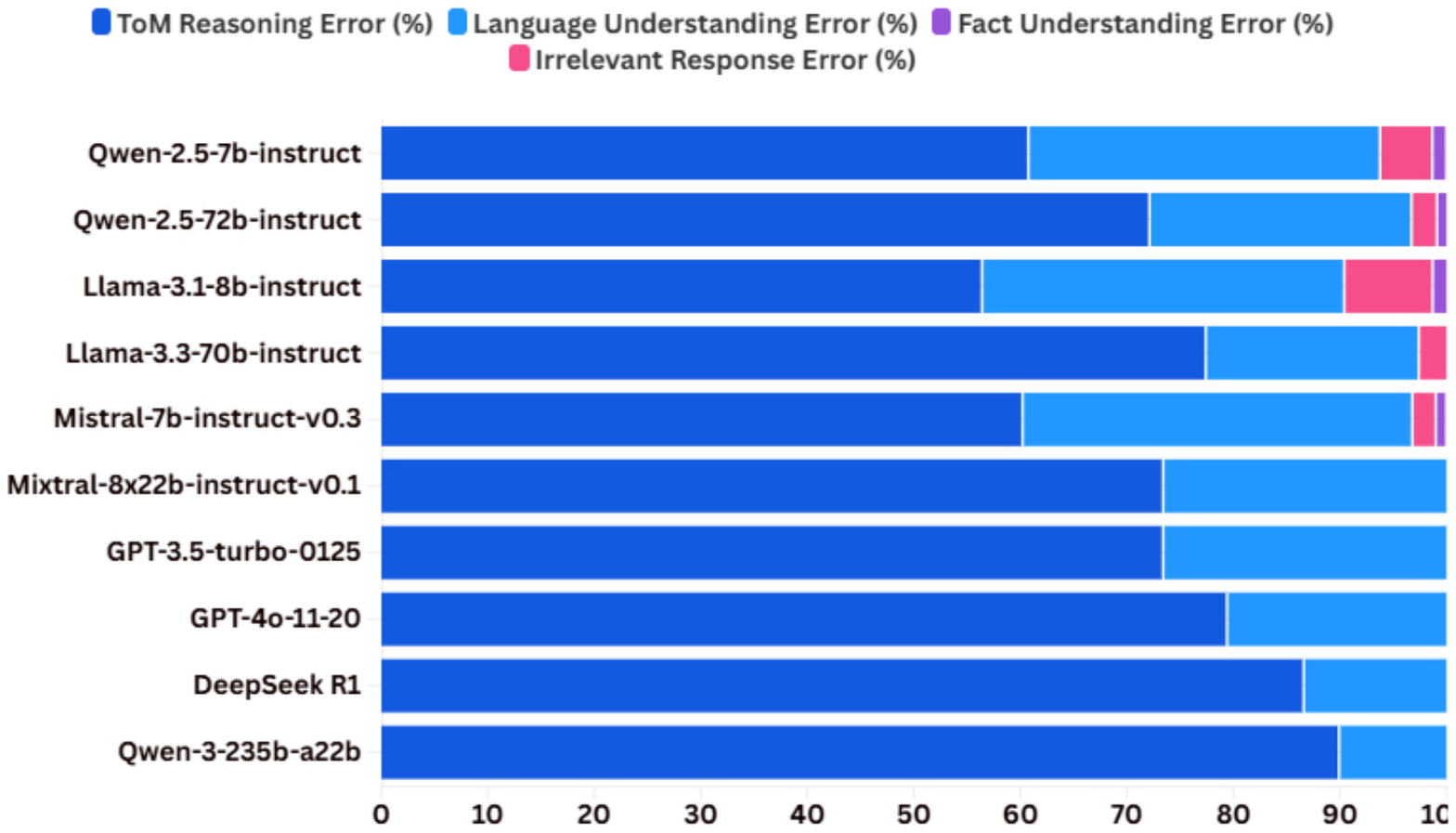}
    \vspace{-1cm}
    \caption{Error analysis of different models on belief question in XFANToM.}
    \label{fig:error_analysis}
    \vspace{-0.6cm}
\end{figure}

\begin{figure}[!t]
    \centering
    \includegraphics[width=0.5\textwidth]{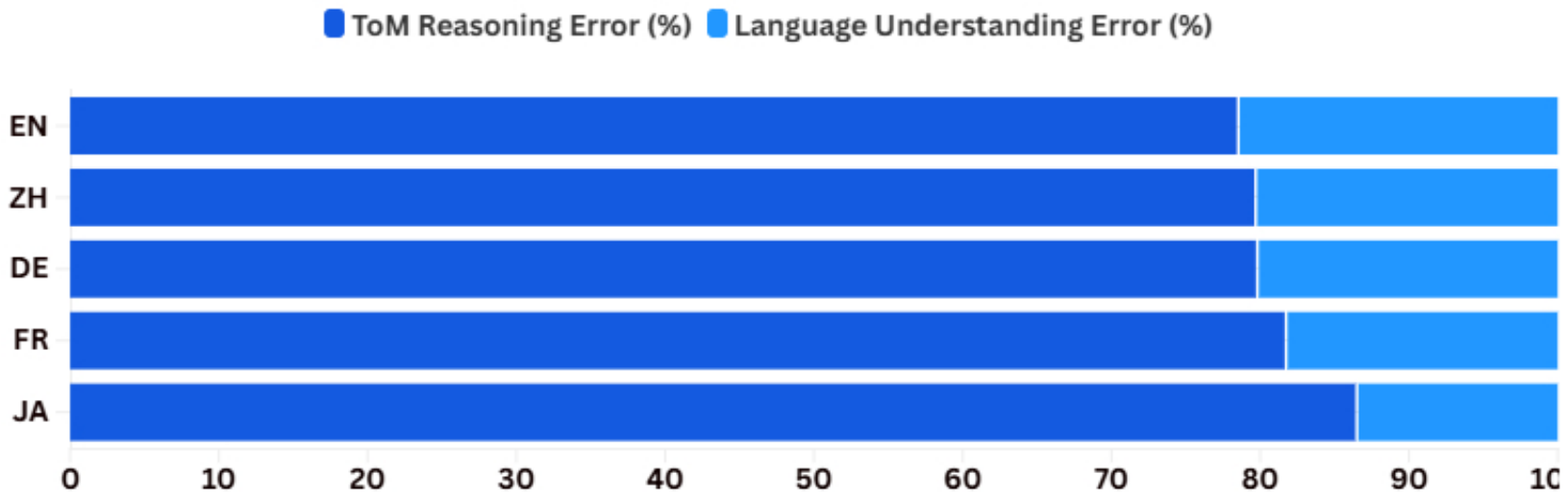}
    \vspace{-2cm}
    \caption{Error analysis of DeepSeek R1 on belief question in XFANToM across language. With DeepSeek R1 receive zero error in Fact Understanding Error and Irrelevant Response Error; these two errors are ignored here.
    }
    \label{fig:error_analysis_language}
    \vspace{-0.6cm}
\end{figure}

\subsection{Appendix for Fine-Tuning LLMs} \label{sec:Fine-Tuning_LLMs}

We employ the LLaMA-Factory~\cite{DBLP:journals/corr/abs-2403-13372} to fine-tune LLMs using Low-Rank Adaptation (LoRA)~\cite{DBLP:conf/iclr/HuSWALWWC22} to achieve parameter-efficient training. The batch size is set to 64, and the learning rate is 5e-5. The LoRA rank is 8, with $\alpha$ set to 32. The maximum length for the input is set to 4,096. We set the warm-up ratio to 0.1, trained the model with 3 epochs, and evaluated the model every 100 steps. A Cosine scheduler is also used.

\section{Appendix for Experimental Result} \label{sec:appendix_result}
\subsection{Appendix for Main Experimental Result} \label{sec:Appendix_for_Experimental_Result}
Figures~\ref{fig:Appendix_belief_choice_fantom},~\ref{fig:Appendix_belief_choice_tomi},~\ref{fig:Appendix_belief_choice_negtom},~\ref{fig:desire_choice_negtom},and~\ref{fig:intention_choice_negtom} display all LLMs' performance of the XNegotiatioNToM, XFANToM, and XToMi. 
More detailed performance (i.e., numerical value of performance) of the XFANToM, and XToMi, XNegotiatioNToM are reported in Tables~\ref{tab:experiment_FANToM},~\ref{tab:experiment_ToMi},and~\ref{tab:experiment_NegotiationToM}. Table~\ref{tab:experiment_NegotiationToM} includes the LLMs' performance on belief, desire, and intention dimensions of mental states across five languages in XNegotiatioNToM. For the desire and intention dimension of ToM, we observe a similar trend where Language variation generally impacts model performance. Moreover, Larger models consistently outperform their smaller models across all three sub-tasks. Smaller models, such as Mistral-7b-instruct-v0.3 and Llama-3.1-8b-instruct, exhibit significantly lower performance, suggesting that larger-scale training enhances ToM capabilities. Furthermore, Table~\ref{tab:experiment_FANToM} provided detailed LLMs' performance on belief questions (including first-order belief and second-order belief) and fact questions in XFANToM, while Table~\ref{tab:experiment_ToMi} displayed the belief question (including first-order belief and second-order belief) and reality question in XToMi.

\begin{table*}[t!] 
\begin{center}
\vspace{-1cm}
\scalebox{0.5}{
\begin{tabular}{cl|c|c|cc|cc|c}
\toprule
& \multirow{2}{*}{Model}  
& \multirow{2}{*}{Language}  
& \multicolumn{5}{c|}{\makecell{\textsc{Belief}\\\textsc{Questions}}}  
& \multicolumn{1}{c}{\makecell{\textsc{Fact}\\\textsc{Questions}}} \\
\cmidrule(r{0.3em}){4-8} \cmidrule(r{0.3em}){9-8}
&                  &     & Choice (\%) & First-Order (\%) & Second-Order (\%)& Acyclic (\%)& Cyclic (\%)& Choice (\%)\\
\midrule           
\multirow{80}{*}{\rotatebox[origin=c]{90}{{\footnotesize \textsc{Short Conversation}}}} 
& GPT-3.5-turbo-0125 & \multirow{18}{*}{EN} & 10.4 & 11.5 & 7.9 & 7.8 & 8.0 & 86.7 \\
& GPT-3.5-turbo-0125 (CoT) & & 17.3 & 18.0 & 15.8 & 21.6 & 10.0 & 86.0 \\
& GPT-4o-11-20 & & 54.1 & 50.7 & 61.4 & 60.8 & 62.0 & 89.7 \\
& GPT-4o-11-20 (CoT) & & 72.6 & 66.4 & \textbf{86.1} & \textbf{90.2} & 82.0 & 90.7 \\
& DeepSeek R1 & & \textbf{75.5} & \textbf{72.8} & 81.2 & 88.2 & 74.0 & \textbf{93.3} \\
& DeepSeek R1 (CoT) & & 67.3 & 67.3 & 67.3 & 66.7 & 68.0 & 84.7 \\
& Qwen-2.5-7b-instruct & & 26.4 & 25.3 & 28.7 & 31.4 & 26.0 & 83.0 \\
& Qwen-2.5-7b-instruct (CoT) & & 31.4 & 26.3 & 42.6 & 47.1 & 38.0 & 85.3 \\
& Qwen-2.5-72b-instruct & & 45.6 & 39.6 & 58.4 & 58.8 & 58.0 & 87.7 \\
& Qwen-2.5-72b-instruct (CoT) & & 68.9 & 62.2 & 83.2 & 88.2 & 78.0 & 85.7 \\
& Qwen-3-235b-a22b &  & 63.2 & 58.5 & 73.3 & 70.6 & 76.0 & 92.3 \\
& Qwen-3-235b-a22b (CoT) &  & 72.0 & 69.1 & 78.2 & 80.4 & 76.0 & 90.7 \\
& Llama-3.1-8b-instruct & & 44.7 & 37.3 & 60.4 & 62.7 & 58.0 & 83.0 \\
& Llama-3.1-8b-instruct (CoT) & & 56.0 & 50.7 & 67.3 & 68.6 & 66.0 & 85.3 \\
& Llama-3.3-70b-instruct & & 58.8 & 53.0 & 71.3 & 68.6 & 74.0 & 95.0 \\
& Llama-3.3-70b-instruct (CoT) & & 65.4 & 59.4 & 78.2 & 80.4 & 76.0 & 89.7 \\
& Mistral-7b-instruct-v0.3 & & 29.9 & 30.0 & 29.7 & 27.5 & 32.0 & 81.7 \\
& Mistral-7b-instruct-v0.3 (CoT) & & 36.8 & 37.8 & 34.7 & 35.3 & 34.0 & 76.3 \\
& Mixtral-8x22b-instruct-v0.1 & & 41.5 & 35.9 & 53.5 & 49.0 & 58.0 & 86.7 \\
& Mixtral-8x22b-instruct-v0.1 (CoT) & & 39.3 & 36.4 & 45.5 & 43.1 & 48.0 & 87.7 \\

\cmidrule{2-9}                                     
& GPT-3.5-turbo-0125 & \multirow{18}{*}{ZH} & 10.4 & 9.7 & 11.9 & 9.8 & 14.0 & 80.7 \\
& GPT-3.5-turbo-0125 (CoT) & & 10.4 & 9.2 & 12.9 & 9.8 & 16.0 & 80.3 \\
& GPT-4o-11-20 & & 48.4 & 42.9 & 60.4 & 54.9 & 66.0 & 89.7 \\
& GPT-4o-11-20 (CoT) & & 59.7 & 51.6 & 77.2 & 78.4 & 76.0 & 90.0 \\
& DeepSeek R1 & & \textbf{84.3} & 81.6 & 90.1 & 86.3 & 94.0 & \textbf{96.0} \\
& DeepSeek R1 (CoT) & & 65.1 & 63.1 & 69.3 & 70.6 & 68.0 & 85.0 \\
& Qwen-2.5-7b-instruct & & 25.2 & 20.7 & 34.70 & 31.4 & 38.0 & 81.3 \\
& Qwen-2.5-7b-instruct (CoT) & & 85.8 & 82.9 & 92.1 & 88.2 & 96.0 & 94.3 \\
& Qwen-2.5-72b-instruct & & 44.7 & 35.5 & 64.4 & 62.7 & 66.0 & 85.3 \\
& Qwen-2.5-72b-instruct (CoT) & & 55.0 & 45.6 & 75.2 & 70.6 & 80.0 & 84.0 \\
& Qwen-3-235b-a22b &  & 63.2 & 58.5 & 73.3 & 70.6 & 76.0 & 92.3 \\
& Qwen-3-235b-a22b (CoT) & & 89.3 & 86.6 & 95.0 & 94.1 & 96.0 & 96.3 \\
& Llama-3.1-8b-instruct & & 30.8 & 29.5 & 33.7 & 33.3 & 34.0 & 74.0 \\
& Llama-3.1-8b-instruct (CoT) & & 27.7 & 26.3 & 30.7 & 29.4 & 32.0 & 72.3 \\
& Llama-3.3-70b-instruct & & 55.3 & 49.8 & 67.3 & 64.7 & 70.0 & 84.3 \\
& Llama-3.3-70b-instruct (CoT) & & 54.4 & 47.9 & 68.3 & 66.7 & 70.0 & 76.7 \\
& Mistral-7b-instruct-v0.3 & & 24.5 & 20.3 & 33.7 & 31.4 & 36.0 & 76.7 \\
& Mistral-7b-instruct-v0.3 (CoT) & & 31.4 & 25.8 & 43.6 & 35.3 & 52.0 & 71.0 \\
& Mixtral-8x22b-instruct-v0.1 & & 35.8 & 33.2 & 41.6 & 31.4 & 52.0 & 85.7 \\
& Mixtral-8x22b-instruct-v0.1 (CoT) & & 36.5 & 32.3 & 45.5 & 35.3 & 56.0 & 83.7 \\

\cmidrule{2-9}                                     
& GPT-3.5-turbo-0125 & \multirow{18}{*}{DE} & 15.4 & 15.2 & 15.8 & 11.8 & 20.0 & 85.0 \\
& GPT-3.5-turbo-0125 (CoT) & & 19.5 & 21.2 & 15.8 & 11.8 & 20.0 & 86.3 \\
& GPT-4o-11-20 & & 46.9 & 42.9 & 55.4 & 60.8 & 50.0 & 90.3 \\
& GPT-4o-11-20 (CoT) & & 67.6 & 61.3 & 81.2 & 80.4 & 82.0 & 88.3 \\
& DeepSeek R1 & & \textbf{83.0} & 82.5 & 84.2 & 82.4 & 86.0 & \textbf{95.7} \\
& DeepSeek R1 (CoT) & & 66.4 & 65.9 & 67.3 & 68.6 & 66.0 & 81.3 \\
& Qwen-2.5-7b-instruct & & 29.9 & 29.5 & 30.7 & 19.6 & 42.0 & 86.3 \\
& Qwen-2.5-7b-instruct (CoT) & & 28.0 & 24.9 & 34.7 & 29.4 & 40.0 & 80.7 \\
& Qwen-2.5-72b-instruct & & 44.0 & 40.1 & 52.5 & 51.0 & 54.0 & 87.3 \\
& Qwen-2.5-72b-instruct (CoT) & & 61.0 & 55.8 & 72.3 & 68.6 & 76.0 & 86.3 \\
& Qwen-3-235b-a22b &  & 40.9 & 44.7 & 32.7 & 21.6 & 44.0 & 93.3 \\
& Qwen-3-235b-a22b (CoT) & & 84.6 & 81.1 & 92.1 & 90.2 & 94.0 & 97.3 \\
& Llama-3.1-8b-instruct & & 41.5 & 39.2 & 46.5 & 52.9 & 40.0 & 75.7 \\
& Llama-3.1-8b-instruct (CoT) & & 34.9 & 31.3 & 42.6 & 39.2 & 46.0 & 78.3 \\
& Llama-3.3-70b-instruct & & 48.4 & 48.8 & 47.5 & 49.0 & 46.0 & 90.7 \\
& Llama-3.3-70b-instruct (CoT) & & 58.2 & 53.9 & 67.3 & 68.6 & 66.0 & 90.3 \\
& Mistral-7b-instruct-v0.3 & & 28.3 & 25.3 & 34.7 & 33.3 & 36.0 & 77.7 \\
& Mistral-7b-instruct-v0.3 (CoT) & & 34.3 & 32.7 & 37.6 & 37.3 & 38.0 & 66.7 \\
& Mixtral-8x22b-instruct-v0.1 & & 34.0 & 33.6 & 34.7 & 31.4 & 38.0 & 87.0 \\
& Mixtral-8x22b-instruct-v0.1 (CoT) & & 33.6 & 35.0 & 30.7 & 27.5 & 34.0 & 86.7 \\

\cmidrule{2-9}                                     
& GPT-3.5-turbo-0125 & \multirow{18}{*}{FR} & 13.8 & 16.1 & 8.9 & 3.9 & 14.0 & 85.0 \\
& GPT-3.5-turbo-0125 (CoT) & & 17.0 & 18.0 & 14.9 & 11.8 & 18.0 & 83.0 \\
& GPT-4o-11-20 & & 56.9 & 52.5 & 66.3 & 72.5 & 60.0 & 87.7 \\
& GPT-4o-11-20 (CoT) & & 71.4 & 65.0 & 85.1 & 86.3 & 84.0 & 90.3 \\
& DeepSeek R1 & & \textbf{82.1} & 81.6 & 83.2 & 80.4 & 86.0 & \textbf{96.0} \\
& DeepSeek R1 (CoT) & & 67.0 & 63.1 & 75.2 & 78.4 & 72.0 & 93.0 \\
& Qwen-2.5-7b-instruct & & 27.4 & 25.8 & 30.7 & 23.5 & 38.0 & 82.7 \\
& Qwen-2.5-7b-instruct (CoT) & & 27.4 & 23.5 & 35.6 & 29.4 & 42.0 & 81.7 \\
& Qwen-2.5-72b-instruct & & 44.7 & 38.2 & 58.4 & 62.7 & 54.0 & 86.3 \\
& Qwen-2.5-72b-instruct (CoT) & & 59.7 & 52.5 & 75.2 & 80.4 & 70.0 & 85.0 \\
& Qwen-3-235b-a22b & & 48.1 & 50.2 & 43.6 & 37.3 & 50.0 & 93.7 \\
& Qwen-3-235b-a22b (CoT) & & 79.2 & 76.0 & 86.1 & 82.4 & 90.0 & 95.7 \\
& Llama-3.1-8b-instruct & & 35.8 & 34.6 & 38.6 & 45.1 & 32.0 & 76.7 \\
& Llama-3.1-8b-instruct (CoT) & & 35.2 & 33.2 & 39.6 & 35.3 & 44.0 & 75.0 \\
& Llama-3.3-70b-instruct & & 53.1 & 49.8 & 60.4 & 58.8 & 62.0 & 89.0 \\
& Llama-3.3-70b-instruct (CoT) & & 56.6 & 51.6 & 67.3 & 66.7 & 68.0 & 88.3 \\
& Mistral-7b-instruct-v0.3 & & 26.1 & 23.5 & 31.7 & 31.4 & 32.0 & 76.7 \\
& Mistral-7b-instruct-v0.3 (CoT) & & 14.5 & 11.1 & 21.8 & 21.6 & 22.0 & 30.7 \\
& Mixtral-8x22b-instruct-v0.1 & & 45.6 & 46.5 & 43.6 & 39.2 & 48.0 & 86.7 \\
& Mixtral-8x22b-instruct-v0.1 (CoT) & & 45.6 & 47.0 & 42.6 & 41.2 & 44.0 & 83.7 \\

\cmidrule{2-9}                                     
& GPT-3.5-turbo-0125 & \multirow{18}{*}{JA} & 20.1 & 17.1 & 26.7 & 27.5 & 26.0 & 81.0 \\
& GPT-3.5-turbo-0125 (CoT) & & 25.2 & 21.7 & 32.7 & 27.5 & 38.0 & 55.7 \\
& GPT-4o-11-20 & & 59.4 & 55.3 & 68.3 & 72.5 & 64.0 & 91.7 \\
& GPT-4o-11-20 (CoT) & & 72.0 & 66.8 & 83.2 & 86.3 & 80.0 & 90.7 \\
& DeepSeek R1 & & \textbf{82.7} & 80.2 & 88.1 & 90.2 & 86.0 & \textbf{97.3} \\
& DeepSeek R1 (CoT) & & 67.3 & 65.4 & 71.3 & 70.6 & 72.0 & 83.3 \\
& Qwen-2.5-7b-instruct & & 43.4 & 37.3 & 56.4 & 51.0 & 62.0 & 80.3 \\
& Qwen-2.5-7b-instruct (CoT) & & 45.9 & 39.2 & 60.4 & 56.9 & 64.0 & 76.3 \\
& Qwen-2.5-72b-instruct & & 44.3 & 39.2 & 55.4 & 47.1 & 64.0 & 86.7 \\
& Qwen-2.5-72b-instruct (CoT) & & 52.2 & 47.0 & 63.4 & 58.8 & 68.0 & 86.3 \\
& Qwen-3-235b-a22b & & 61.9 & 62.7 & 60.4 & 54.9 & 66.0 & 95.0 \\
& Qwen-3-235b-a22b (CoT) & & 81.1 & 79.3 & 85.1 & 84.3 & 86.0 & 94.3 \\
& Llama-3.1-8b-instruct & & 38.7 & 35.9 & 44.6 & 47.1 & 42.0 & 74.7 \\
& Llama-3.1-8b-instruct (CoT) & & 38.7 & 40.6 & 34.7 & 37.3 & 32.0 & 73.7 \\
& Llama-3.3-70b-instruct & & 42.1 & 35.9 & 55.4 & 56.9 & 54.0 & 91.7 \\
& Llama-3.3-70b-instruct (CoT) & & 51.9 & 46.5 & 63.4 & 66.7 & 60.0 & 87.0 \\
& Mistral-7b-instruct-v0.3 & & 28.0 & 24.4 & 35.6 & 35.3 & 36.0 & 75.3 \\
& Mistral-7b-instruct-v0.3 (CoT) & & 15.1 & 12.0 & 21.8 & 21.6 & 22.0 & 41.0 \\
& Mixtral-8x22b-instruct-v0.1 & & 47.8 & 44.2 & 55.4 & 47.1 & 64.0 & 86.0 \\
& Mixtral-8x22b-instruct-v0.1 (CoT) & & 38.1 & 34.1 & 46.5 & 47.1 & 46.0 & 70.7 \\
\bottomrule
\end{tabular}
}
\caption{Results of models for XFANToM (short conversation).
}
\vspace{-10pt}
\label{tab:experiment_FANToM}
\end{center}
\end{table*}

\begin{table*}[!t]
\small
\vspace{-1cm}
\centering
\setlength\tabcolsep{4pt}
\scalebox{0.68}{
\begin{tabular}{l|c|c|c|c|c}
\toprule
\multicolumn{1}{c|}{\multirow{2}{*}{\textbf{Model}}} & \multicolumn{1}{c|}{\multirow{2}{*}{\textbf{Language}}} & \multicolumn{1}{c|}{\textbf{1st Belief}} & \multicolumn{1}{c|}{\textbf{2nd Belief}} & \multicolumn{1}{c|}{\textbf{Average}} & \multicolumn{1}{c}{\textbf{Reality}}\\

                          &  & Accuracy(\%)    & Accuracy(\%)      & Accuracy(\%)     & Accuracy(\%)  \\

\midrule
GPT-3.5-turbo-0125 & \multirow{18}{*}{\centering EN} & 49.59 & 41.07 & 44.09 & 92.00 \\
GPT-3.5-turbo-0125 (CoT) & & 41.46 & 31.70 & 35.16 & 85.33 \\
GPT-4o-11-20 & & 96.02 & 93.30 & 94.24 & \textbf{100.00} \\
GPT-4o-11-20 (CoT) & & 95.12 & 93.75 & 94.24 & \textbf{100.00} \\
DeepSeek R1 & & 84.55 & 99.55 & 94.24 & 99.67 \\
DeepSeek R1 (CoT) & & 95.12 & \textbf{100.00} & \textbf{98.27} & \textbf{100.00} \\
Qwen-2.5-7b-instrcut & & 46.34 & 40.63 & 42.65 & 99.33 \\
Qwen-2.5-7b-instrcut (CoT) & & 60.98 & 55.36 & 57.35 & 96.67 \\
Qwen-2.5-72b-instrcut & & 70.73 & 70.09 & 70.32 & \textbf{100.00} \\
Qwen-2.5-72b-instrcut (CoT) & & 56.91 & 62.05 & 60.23 & \textbf{100.00} \\
Qwen-3-235b-a22b & & 70.05 & 75.00 & 73.20 & \textbf{100.00} \\
Qwen-3-235b-a22b  (CoT) &  & 100.00 & 99.55 & 99.71 & \textbf{100.00} \\
Llama-3.1-8b-instruct & & 39.84 & 42.41 & 41.50 & 77.00 \\
Llama-3.1-8b-instruct (CoT) & & 52.85 & 55.36 & 54.47 & 67.00 \\
Llama-3.3-70b-instruct & & 99.19 & 95.54 & 96.83 & \textbf{100.00} \\
Llama-3.3-70b-instruct (CoT) & & \textbf{100.00} & 95.54 & 97.12 & \textbf{100.00} \\
Mistral-7b-instruct-v0.3 & & 39.84 & 41.52 & 40.92 & 93.00 \\
Mistral-7b-instruct-v0.3 (CoT) & & 39.02 & 48.66 & 45.24 & 93.00 \\
Mixtral-8x22b-instruct-v0.1 & & 79.67 & 84.38 & 82.71 & \textbf{100.00} \\
Mixtral-8x22b-instruct-v0.1 (CoT) & & 70.73 & 80.80 & 77.23 & \textbf{100.00} \\

\midrule
GPT-3.5-turbo-0125 & \multirow{18}{*}{\centering ZH} & 34.96 & 36.16 & 35.73 & 85.67 \\
GPT-3.5-turbo-0125 (CoT) & & 39.84 & 44.64 & 42.94 & 78.67 \\
GPT-4o-11-20 & & \textbf{68.29} & 91.52 & 83.29 & \textbf{100.00} \\
GPT-4o-11-20 (CoT) & & 61.79 & 91.96 & 81.27 & \textbf{100.00} \\
DeepSeek R1 & & 61.79 & \textbf{100.00} & \textbf{86.46} & \textbf{100.00} \\
DeepSeek R1 (CoT) & & 60.98 & 99.55 & 85.88 & \textbf{100.00} \\
Qwen-2.5-7b-instruct & & 17.89 & 54.91 & 41.79 & 97.00 \\
Qwen-2.5-7b-instruct (CoT) & & 19.51 & 59.82 & 45.53 & 87.00 \\
Qwen-2.5-72b-instruct & & 20.33 & 42.86 & 34.87 & \textbf{100.00} \\
Qwen-2.5-72b-instruct (CoT) & & 12.20 & 48.21 & 35.45 & \textbf{100.00} \\
Qwen-3-235b-a22b &  & 46.72 & 80.36 & 68.30 & \textbf{100.00} \\
Qwen-3-235b-a22b (CoT) & & 78.02 & 99.55 & 91.93 & \textbf{100.00} \\
Llama-3.1-8b-instruct & & 28.46 & 29.02 & 28.82 & 83.00 \\
Llama-3.1-8b-instruct (CoT) & & 29.27 & 30.80 & 30.26 & 73.67 \\
Llama-3.3-70b-instruct & & 60.98 & 72.77 & 68.59 & 99.67 \\
Llama-3.3-70b-instruct (CoT) & & 60.16 & 74.11 & 69.16 & \textbf{100.00} \\
Mistral-7b-instruct-v0.3 & & 52.03 & 45.54 & 47.84 & 57.33 \\
Mistral-7b-instruct-v0.3 (CoT) & & 49.59 & 49.11 & 49.28 & 64.00 \\
Mixtral-8x22b-instruct-v0.1 & & 60.16 & 49.11 & 53.03 & \textbf{100.00} \\
Mixtral-8x22b-instruct-v0.1 (CoT) & & 57.72 & 50.45 & 53.03 & \textbf{100.00} \\

\midrule
GPT-3.5-turbo-0125 & \multirow{18}{*}{\centering DE} & 42.28 & 35.71 & 38.04 & 86.33 \\
GPT-3.5-turbo-0125 (CoT) & & 39.02 & 24.55 & 29.68 & 83.00 \\
GPT-4o-11-20 & & 90.24 & 90.63 & 90.49 & \textbf{98.67} \\
GPT-4o-11-20 (CoT) & & 87.80 & 94.20 & 91.93 & 97.67 \\
DeepSeek R1 & & \textbf{95.12} & \textbf{98.21} & \textbf{97.12} & \textbf{98.67} \\
DeepSeek R1 (CoT) & & \textbf{95.12} & \textbf{98.21} & \textbf{97.12} & 98.33 \\
Qwen-2.5-7b-instruct & & 65.85 & 47.32 & 53.89 & 87.00 \\
Qwen-2.5-7b-instruct (CoT) & & 69.11 & 52.23 & 58.21 & 71.00 \\
Qwen-2.5-72b-instruct & & 66.67 & 69.64 & 68.59 & 98.00 \\
Qwen-2.5-72b-instruct (CoT) & & 61.79 & 73.66 & 69.45 & 98.00 \\
Qwen-3-235b-a22b & & 64.30 & 75.89 & 71.76 & 97.00 \\
Qwen-3-235b-a22b (CoT) & & 94.34 & 97.32 & 96.25 & 98.00 \\
Llama-3.1-8b-instruct & & 51.22 & 48.21 & 49.28 & 71.67 \\
Llama-3.1-8b-instruct (CoT) & & 41.46 & 43.30 & 42.65 & 68.00 \\
Llama-3.3-70b-instruct & & 88.62 & 86.61 & 87.32 & 96.33 \\
Llama-3.3-70b-instruct (CoT) & & 87.00 & 79.46 & 82.13 & 93.33 \\
Mistral-7b-instruct-v0.3 & & 61.79 & 37.05 & 45.82 & 57.33 \\
Mistral-7b-instruct-v0.3 (CoT) & & 60.16 & 40.18 & 47.26 & 58.67 \\
Mixtral-8x22b-instruct-v0.1 & & 62.60 & 56.70 & 58.79 & 97.67 \\
Mixtral-8x22b-instruct-v0.1 (CoT) & & 55.28 & 50.45 & 52.16 & 96.67 \\

\midrule
GPT-3.5-turbo-0125 & \multirow{18}{*}{\centering FR} & 29.27 & 35.27 & 33.14 & 96.00 \\
GPT-3.5-turbo-0125 (CoT) & & 32.52 & 25.89 & 28.24 & 93.00 \\
GPT-4o-11-20 & & 86.18 & 89.29 & 88.18 & 99.67 \\
GPT-4o-11-20 (CoT) & & 88.62 & 95.98 & 93.37 & 99.67 \\
DeepSeek R1 & & \textbf{90.24} & \textbf{99.55} & \textbf{96.25} & \textbf{100.00} \\
DeepSeek R1 (CoT) & & 86.99 & \textbf{99.55} & 95.10 & \textbf{100.00} \\
Qwen-2.5-7b-instruct & & 34.96 & 25.45 & 28.82 & 99.00 \\
Qwen-2.5-7b-instruct (CoT) & & 49.59 & 34.82 & 40.06 & 97.00 \\
Qwen-2.5-72b-instruct & & 69.11 & 76.34 & 73.78 & 99.33 \\
Qwen-2.5-72b-instruct (CoT) & & 59.35 & 75.89 & 70.03 & 99.67 \\
Qwen-3-235b-a22b & & 46.54 & 83.93 & 70.61 & 98.33 \\
Qwen-3-235b-a22b (CoT) & & 95.01 & 98.66 & 97.41 & 98.67 \\
Llama-3.1-8b-instruct & & 27.64 & 41.07 & 36.31 & 68.33 \\
Llama-3.1-8b-instruct (CoT) & & 43.09 & 43.75 & 43.52 & 70.67 \\
Llama-3.3-70b-instruct & & 86.18 & 90.63 & 89.05 & 99.33 \\
Llama-3.3-70b-instruct (CoT) & & 89.43 & 91.07 & 90.49 & 99.33 \\
Mistral-7b-instruct-v0.3 & & 37.40 & 23.66 & 28.53 & 89.67 \\
Mistral-7b-instruct-v0.3 (CoT) & & 34.96 & 32.59 & 33.43 & 84.00 \\
Mixtral-8x22b-instruct-v0.1 & & 60.98 & 60.27 & 60.52 & 99.33 \\
Mixtral-8x22b-instruct-v0.1 (CoT) & & 63.41 & 66.52 & 65.42 & 99.67 \\

\midrule
GPT-3.5-turbo-0125 & \multirow{18}{*}{\centering JA} & 52.85 & 50.89 & 51.59 & 65.00 \\
GPT-3.5-turbo-0125 (CoT) & & 52.85 & 52.23 & 52.45 & 66.00 \\
GPT-4o-11-20 & & 81.71 & 94.64 & 90.20 & \textbf{100.00} \\
GPT-4o-11-20 (CoT) & & 82.93 & 95.98 & 91.35 & \textbf{100.00} \\
DeepSeek R1 & & 37.2 & 96.40 & 76.08 & \textbf{100.00} \\
DeepSeek R1 (CoT) & & 26.83 & 95.09 & 70.89 & \textbf{100.00} \\
Qwen-2.5-7b-instruct & & 45.53 & 70.09 & 61.38 & 74.00 \\
Qwen-2.5-7b-instruct (CoT) & & 55.28 & 77.68 & 69.74 & 61.67 \\
Qwen-2.5-72b-instruct & & 33.33 & 52.23 & 45.53 & 99.67 \\
Qwen-2.5-72b-instruct (CoT) & & 30.08 & 53.13 & 44.96 & \textbf{100.00} \\
Qwen-3-235b-a22b & & 70.52 & 69.20 & 69.45 & 99.00 \\
Qwen-3-235b-a22b (CoT) & & \textbf{91.84} & \textbf{98.21} & \textbf{95.97} & 98.67 \\
Llama-3.1-8b-instruct & & 29.27 & 38.84 & 35.45 & 65.00 \\
Llama-3.1-8b-instruct (CoT) & & 41.46 & 41.52 & 41.50 & 56.67 \\
Llama-3.3-70b-instruct & & 69.11 & 68.75 & 68.88 & 98.00 \\
Llama-3.3-70b-instruct (CoT) & & 59.35 & 54.46 & 56.20 & 98.00 \\
Mistral-7b-instruct-v0.3 & & 60.16 & 43.30 & 49.28 & 32.00 \\
Mistral-7b-instruct-v0.3 (CoT) & & 55.28 & 40.63 & 45.82 & 35.00 \\
Mixtral-8x22b-instruct-v0.1 & & 76.42 & 42.41 & 54.47 & 97.33 \\
Mixtral-8x22b-instruct-v0.1 (CoT) & & 74.80 & 47.77 & 57.35 & 95.67 \\
\bottomrule
\end{tabular}
}
\caption{
Results of models for XToMi.
}
\label{tab:experiment_ToMi}
\end{table*}

\begin{table*}[!t]
\small
\vspace{-1cm}
\centering
\setlength\tabcolsep{4pt}
\scalebox{0.68}{
\begin{tabular}{l|c|c|c|c|c}
\toprule
\multicolumn{1}{c|}{\multirow{2}{*}{\textbf{Model}}} & \multicolumn{1}{c|}{\multirow{2}{*}{\textbf{Language}}} & \multicolumn{1}{c|}{\textbf{Belief}} & \multicolumn{1}{c|}{\textbf{Desire}} & \multicolumn{2}{c}{\textbf{Intention}} \\

                          &  & Exact.Match.(\%)    & Exact.Match.(\%)      & Micro.F1(\%)     & Macro.F1(\%)  \\

\midrule
GPT-3.5-turbo-0125 & \multirow{18}{*}{\centering EN} & 14.33 & 18.17 & 35.00 & 27.00 \\
GPT-3.5-turbo-0125 (CoT) & & 11.83 & 19.67 & 35.00 & 24.00 \\
GPT-4o-11-20 & & 51.70 & 55.80 & \textbf{49.00} & \textbf{42.00} \\
GPT-4o-11-20 (CoT) & & 46.70 & 52.30 & \underline{48.00} & \underline{41.00} \\
DeepSeek R1 & & \underline{62.83} & \textbf{66.50} & \textbf{49.00} & \underline{41.00} \\
DeepSeek R1 (CoT) & & \textbf{63.00} & \underline{64.50} & \textbf{49.00} & \underline{41.00} \\
Qwen-2.5-7b-instrcut & & 7.34 & 13.33 & 44.00 & 35.00 \\
Qwen-2.5-7b-instrcut(CoT) & & 8.34 & 14.00 & \underline{48.00} & 38.00 \\
Qwen-2.5-72b-instrcut & & 32.84 & 44.84 & 43.00 & 35.00 \\
Qwen-2.5-72b-instrcut(CoT) & & 24.67 & 37.84 & 44.00 & 37.00 \\
Qwen-3-235b-a22b & & 48.17 & 54.17 & 44.00 & 37.00\\
Qwen-3-235b-a22b (CoT) & & 63.33 & 66.67 & 50.00 & 43.00 \\
Llama-3.1-8b-instruct & & 12.00 & 18.17 & 30.00 & 24.00 \\
Llama-3.1-8b-instruct (CoT) & & 19.50 & 28.84 & 39.00 & 29.00 \\
Llama-3.3-70b-instruct & & 41.50 & 52.84 & 42.00 & 35.00 \\
Llama-3.3-70b-instruct (CoT) & & 54.00 & 61.84 & \underline{48.00} & 40.00 \\
Mistral-7b-instruct-v0.3 & & 8.67 & 11.00 & 36.00 & 28.00 \\
Mistral-7b-instruct-v0.3 (CoT) & & 19.84 & 27.00 & 38.00 & 28.00 \\
Mixtral-8x22b-instruct-v0.1 & & 40.84 & 39.00 & 36.00 & 30.00 \\
Mixtral-8x22b-instruct-v0.1 (CoT) & & 51.67 & 53.00 & 40.00 & 33.00 \\
\midrule
GPT-3.5-turbo-0125              & \multirow{18}{*}{\centering ZH} & 17.76   & 20.34 & 30.00 & 22.00 \\
GPT-3.5-turbo-0125 (CoT) & & 14.14 & 19.48 & 30.00 & 22.00  \\
GPT-4o-11-20 & & 40.35 & 42.59 & 46.00 & 39.00 \\
GPT-4o-11-20 (CoT) & & 44.48 & 46.73 & 46.00 & 40.00 \\
DeepSeek R1 & & \textbf{52.59} & \textbf{57.76} & 45.00 & \textbf{43.00} \\
DeepSeek R1 (CoT) & & \underline{48.97} & \underline{53.97} & \underline{48.00} & 40.00 \\
Qwen-2.5-7b-instrcut & & 18.28 & 18.97 & 46.00 & 34.00 \\
Qwen-2.5-7b-instrcut(CoT) & & 15.00 & 21.03 & 50.00 & 36.00 \\
Qwen-2.5-72b-instrcut & & 32.59 & 38.28 & 39.00 & 34.00 \\
Qwen-2.5-72b-instrcut(CoT) & & 34.14 & 39.83 & 40.00 & 34.00 \\
Qwen-3-235b-a22b & & 44.48 & 47.24 & 45.00 & 38.00 \\
Qwen-3-235b-a22b (CoT) & & 54.14 & 58.79 & 45.00 & 38.00 \\
Llama-3.1-8b-instruct  & & 11.03 & 13.97 & 23.00 & 18.00  \\
Llama-3.1-8b-instruct (CoT) & & 13.28 & 17.42 & 27.00 & 21.00 \\
Llama-3.3-70b-instruct & & 42.00 & 48.67 & 39.00 & 35.00 \\
Llama-3.3-70b-instruct (CoT) & & 46.21 & 50.17 & \textbf{49.00} & \underline{41.00} \\
Mistral-7b-instruct-v0.3 & & 8.67 & 11.00 & 25.00 & 18.00 \\
Mistral-7b-instruct-v0.3 (CoT) & & 21.72 & 21.03 & 34.00 & 25.00 \\
Mixtral-8x22b-instruct-v0.1 & & 34.67 & 42.00 & 38.00 & 28.00 \\
Mixtral-8x22b-instruct-v0.1 (CoT) & & 42.76 & 47.25 & 39.00 & 32.00 \\

\midrule

GPT-3.5-turbo-0125 & \multirow{18}{*}{\centering DE} & 16.67 & 21.33 & 33.00 & 25.00 \\
GPT-3.5-turbo-0125 (CoT) & & 16.83 & 20.50 & 23.00 & 15.00 \\
GPT-4o-11-20 & & \underline{52.84} & 53.50 & \underline{47.00} & \underline{39.00} \\
GPT-4o-11-20 (CoT) & & 51.17 & 51.84 & \textbf{48.00} & \textbf{41.00} \\
DeepSeek R1 & & \textbf{57.33} & \textbf{63.67} & 45.00 & 37.00 \\
DeepSeek R1 (CoT) & & \textbf{57.33} & \underline{63.33} & 44.00 & 36.00 \\
Qwen-2.5-7b-instrcut & & 16.50 & 17.84 & 44.00 & 31.00 \\
Qwen-2.5-7b-instrcut(CoT) & & 14.00 & 12.00 & 43.00 & 29.00 \\
Qwen-2.5-72b-instrcut & & 34.33 & 42.50 & 40.00 & 34.00 \\
Qwen-2.5-72b-instrcut(CoT) & & 25.33 & 39.00 & 40.00 & 33.00 \\
Qwen-3-235b-a22b & & 37.67 & 51.50 &44.00 & 37.00 \\
Qwen-3-235b-a22b (CoT) &  & 61.33 & 64.17 & 41.00 & 34.00 \\
Llama-3.1-8b-instruct & & 11.67 & 13.67 & 22.00 & 15.00 \\
Llama-3.1-8b-instruct (CoT) & & 12.67 & 16.17 & 28.00 & 19.00 \\
Llama-3.3-70b-instruct & & 47.50 & 56.34 & 40.00 & 33.00 \\
Llama-3.3-70b-instruct (CoT) & & 49.00 & 55.84 & 43.00 & 36.00 \\
Mistral-7b-instruct-v0.3 & & 6.50 & 6.34 & 29.00 & 23.00 \\
Mistral-7b-instruct-v0.3 (CoT) & & 21.17 & 21.83 & 28.00 & 22.00 \\
Mixtral-8x22b-instruct-v0.1 & & 30.34 & 37.50 & 35.00 & 28.00 \\
Mixtral-8x22b-instruct-v0.1 (CoT) & & 34.84 & 45.67 & 38.00 & 30.00 \\

\midrule
GPT-3.5-turbo-0125 & \multirow{18}{*}{\centering FR} & 17.17 & 22.67 & 35.00 & 26.00 \\
GPT-3.5-turbo-0125 (CoT) & & 17.33 & 22.50 & 37.00 & 29.00 \\
GPT-4o-11-20 & & 47.00 & 53.33 & \underline{48.00} & 40.00 \\
GPT-4o-11-20 (CoT) & & 44.50 & 48.67 & \textbf{49.00} & \underline{41.00} \\
DeepSeek R1 & & \textbf{52.17} & \textbf{62.50} & 45.00 & \textbf{44.00} \\
DeepSeek R1 (CoT) & & \underline{50.34} & \underline{62.17} & 47.00 & 40.00 \\
Qwen-2.5-7b-instrcut & & 11.34 & 13.67 & 43.00 & 32.00 \\
Qwen-2.5-7b-instrcut (CoT) & & 7.50 & 11.34 & 44.00 & 32.00 \\
Qwen-2.5-72b-instrcut & & 37.84 & 24.67 & 45.00 & 37.00 \\
Qwen-2.5-72b-instrcut (CoT) & & 32.83 & 24.33 & 44.00 & 37.00 \\
Qwen-3-235b-a22b & & 37.17 & 35.33 & 38.00 & 34.00 \\
Qwen-3-235b-a22b (CoT) & & 55.17 & 56.83 & 38.00 & 34.00 \\
Llama-3.1-8b-instruct & & 12.33 & 13.83 & 29.00 & 23.00 \\
Llama-3.1-8b-instruct (CoT) & & 17.67 & 14.50 & 28.00 & 21.00 \\
Llama-3.3-70b-instruct & & 32.00 & 48.50 & 42.00 & 36.00 \\
Llama-3.3-70b-instruct (CoT) & & 34.50 & 49.17 & 45.00 & 37.00 \\
Mistral-7b-instruct-v0.3 & & 3.67 & 8.17 & 28.00 & 23.00 \\
Mistral-7b-instruct-v0.3 (CoT) & & 24.83 & 24.50 & 31.00 & 24.00 \\
Mixtral-8x22b-instruct-v0.1 & & 33.00 & 37.50 & 44.00 & 34.00 \\
Mixtral-8x22b-instruct-v0.1 (CoT) & & 43.17 & 43.67 & 44.00 & 36.00 \\

\midrule
GPT-3.5-turbo-0125 & \multirow{18}{*}{\centering JA} & 15.00 & 17.50 & 21.00 & 17.00 \\
GPT-3.5-turbo-0125 (CoT) & & 12.83 & 16.50 & 19.00 & 14.00 \\
GPT-4o-11-20 & & 44.17 & 51.50 & \textbf{45.00} & \underline{37.00} \\
GPT-4o-11-20 (CoT) & & 40.50 & 49.16 & \textbf{45.00} & \textbf{39.00} \\
DeepSeek R1 & & \textbf{52.00} & \textbf{55.17} & \textbf{45.00} & 36.00 \\
DeepSeek R1 (CoT) & & \underline{48.17} & \underline{53.34} & \underline{44.00} & 35.00 \\
Qwen-2.5-7b-instrcut & & 12.17 & 18.50 & 42.00 & 34.00 \\
Qwen-2.5-7b-instrcut (CoT) & & 16.50 & 19.67 & 42.00 & 34.00 \\
Qwen-2.5-72b-instrcut & & 38.17 & 41.33 & 43.00 & 36.00 \\
Qwen-2.5-72b-instrcut (CoT) & & 31.84 & 40.17 & 43.00 & 35.00 \\
Qwen-3-235b-a22b & & 40.17 & 47.00 & 40.00 & 33.00 \\
Qwen-3-235b-a22b (CoT) & & 52.33 & 56.50 & 40.00 & 34.00 \\
Llama-3.1-8b-instruct & & 9.17 & 11.17 & 43.00 & 32.00 \\
Llama-3.1-8b-instruct (CoT) & & 12.34 & 11.67 & 22.00 & 15.00 \\
Llama-3.3-70b-instruct & & 28.50 & 45.33 & \textbf{45.00} & \underline{37.00} \\
Llama-3.3-70b-instruct (CoT) & & 37.00 & 52.17 & \underline{44.00} & \underline{37.00} \\
Mistral-7b-instruct-v0.3 & & 13.83 & 6.33 & 28.00 & 23.00 \\
Mistral-7b-instruct-v0.3 (CoT) & & 15.50 & 14.67 & 31.00 & 24.00 \\
Mixtral-8x22b-instruct-v0.1 & & 31.83 & 41.33 & 44.00 & 34.00 \\
Mixtral-8x22b-instruct-v0.1 (CoT) & & 31.83 & 43.00 & 44.00 & 36.00 \\

\midrule
\end{tabular}
}
\caption{
Results of models for XNegotiationToM.
}
\label{tab:experiment_NegotiationToM}
\end{table*}

\begin{table*}[!t]
\small
\centering
\setlength\tabcolsep{4pt}
\scalebox{1.0}{
\begin{tabular}{l|c|c|c|c|c}
\toprule
\multicolumn{1}{c|}{\multirow{2}{*}{\textbf{Model}}} & \multicolumn{1}{c|}{\multirow{2}{*}{\textbf{Language}}} & \multicolumn{1}{c|}{\textbf{1st Belief}} & \multicolumn{1}{c|}{\textbf{2nd Belief}} & \multicolumn{1}{c|}{\textbf{Average}} & \multicolumn{1}{c}{\textbf{Reality}}\\

                          &  & Accuracy(\%)    & Accuracy(\%)      & Accuracy(\%)     & Accuracy(\%)  \\

\midrule
GPT-4o-11-20 & \multirow{6}{*}{\centering ZH} & 69.92 & 93.75 & 85.30 & 100.00 \\
GPT-4o-11-20 (CoT) & & 65.85 & 95.09 & 84.73 & 100.00 \\
Llama-3.3-70b-instruct & & 65.85 & 83.48 & 77.23 & 99.67 \\
Llama-3.3-70b-instruct (CoT) & & 68.29 & 84.82 & 78.96 & 100.00 \\
Mixtral-8x22b-instruct-v0.1 & & 56.91 & 56.25 & 56.48 & 100.00 \\
Mixtral-8x22b-instruct-v0.1 (CoT) & & 51.22 & 51.79 & 51.59 & 100.00 \\

\midrule
GPT-4o-11-20 & \multirow{6}{*}{\centering DE} & 85.37 & 89.29 & 87.90 & 98.67 \\
GPT-4o-11-20 (CoT) & & 73.98 & 89.73 & 84.15 & 98.33 \\
Llama-3.3-70b-instruct & & 90.24 & 87.50 & 88.47 & 96.33 \\
Llama-3.3-70b-instruct (CoT) & & 88.62 & 87.50 & 87.90 & 95.67 \\
Mixtral-8x22b-instruct-v0.1 & & 67.48 & 45.98 & 53.60 & 96.33 \\
Mixtral-8x22b-instruct-v0.1 (CoT) & & 65.85 & 46.88 & 53.60 & 96.00 \\

\midrule
GPT-4o-11-20 & \multirow{6}{*}{\centering FR} & 79.67 & 94.20 & 89.05 & 99.67 \\
GPT-4o-11-20 (CoT) & & 86.18 & 98.21 & 93.95 & 100.00 \\
Llama-3.3-70b-instruct & & 88.62 & 95.98 & 93.37 & 100.00 \\
Llama-3.3-70b-instruct (CoT) & & 91.06 & 94.64 & 93.37 & 99.33 \\
Mixtral-8x22b-instruct-v0.1 & & 65.04 & 56.70 & 59.65 & 100.00 \\
Mixtral-8x22b-instruct-v0.1 (CoT) & & 57.72 & 56.25 & 56.77 & 100.00 \\

\midrule
GPT-4o-11-20 & \multirow{6}{*}{\centering JA} & 73.17 & 93.75 & 86.46 & 100.00 \\
GPT-4o-11-20 (CoT) & & 78.05 & 95.98 & 89.63 & 100.00 \\
Llama-3.3-70b-instruct & & 70.73 & 87.05 & 81.27 & 98.67 \\
Llama-3.3-70b-instruct (CoT) & & 67.48 & 75.00 & 72.33 & 99.33 \\
Mixtral-8x22b-instruct-v0.1 & & 61.79 & 51.34 & 55.04 & 91.00 \\
Mixtral-8x22b-instruct-v0.1 (CoT) & & 60.16 & 49.11 & 53.03 & 92.00 \\

\bottomrule
\end{tabular}
}
\caption{
Results of models for other languages in EN template (XToMi benchmark).
}
\label{tab:experiment}
\end{table*}

\subsection{Appendix for Consistency Analysis}
\label{App:consistency}
The consistency analysis of ToMi is shown in Figure \ref{fig:tomi_consistency}, revealing a similar trend to FANToM. As ToMi is a relatively easier task, models achieve higher overall consistency. For fact questions, models demonstrate high consistency, with nearly zero consistently false answers. However, for belief questions, consistency drops significantly, further emphasizing the models’ weaker ToM reasoning compared to fact retrieval.

\subsection{Error Analysis}\label{App:Error_Analysis}
The error analysis of different models in XFANToM is presented in Tables~\ref{fig:error_analysis} and~\ref{fig:error_analysis_language}. We identified several distinct error types and conducted a statistical analysis of their distributions across the models. These error categories are defined as follows:(1) \textit{Theory of Mind Reasoning Error}: The model provides a correct prediction for the fact-based question but an incorrect response for the Theory of Mind (ToM) question, indicating a specific deficit in ToM reasoning. (2) \textit{Language Understanding Error}: The model produces incorrect predictions for both the fact-based question and the ToM question, suggesting broader challenges in language comprehension. (3) \textit{Fact Understanding Error}: The model gives an incorrect prediction for the fact-based question but a correct response for the ToM question, pointing to issues in factual understanding rather than ToM reasoning. (4)\textit{Irrelevant Response Error}: The model generates responses that are unrelated to the posed questions (e.g., repeating the question without providing an answer), reflecting a failure to produce meaningful output. Table~\ref{fig:error_analysis} summarizes the errors made by each model across these categories, while Table~\ref{fig:error_analysis_language} details the distribution of errors made by DeepSeek R1 across different languages. The results, as shown in Table~\ref{fig:error_analysis_language}, reveal that all models exhibit a significant proportion of ToM reasoning errors. This finding suggests that large language models (LLMs) struggle with Theory of Mind reasoning abilities, reflecting inherent limitations in social cognition rather than deficiencies in language understanding. Consistent with prior research~\cite{DBLP:conf/eacl/ShapiraLAZCGSS24}, LLMs’ proficiency in statistical language patterns does not translate to human-like social reasoning capabilities. Moreover, we also provide a case study of XFANToM in Table~\ref{tab:case_study}, investigating the cultural context embedded in language influencing the model's performance.

\begin{table*}[htbp]
\scriptsize 
\renewcommand{\arraystretch}{1.0}
\centering
\begin{tabularx}{\textwidth}{>{\centering\arraybackslash}p{2.5cm} X}
\toprule
\textbf{Dimension} & \textbf{Example} \\
\midrule

\textbf{Belief} &
\textbf{Background:} Here is a negotiation conversation for a camping trip.
There are two agents who own some basic supplies and negotiate 
with each other to split the additional food packages, water bottles, 
and firewood to make their camping trip even better. Each of these 
items will be of either High, Medium or Low priority for these 
two agents. Each of the additional items only has an available 
quantity of 3. Please answer the following three questions using 
``A'', ``B'', ``C'', ``D'' without any explanation.\par
\textbf{Dialogue History:}\par
\textbf{\{Context\}}\par
\textbf{Question1:} Based on the dialogue, what is the high preference for items Agent 1 thinks Agent 2 is?\par
A.Not given\par
B.Water\par
C.Food\par
D.Firewood\par
\textbf{Question2:} Based on the dialogue, what is the medium preference for items Agent 1 thinks Agent 2 is?\par
A.Not given\par
B.Water\par
C.Food\par
D.Firewood\par
\textbf{Question3:} Based on the dialogue, what is the low preference for items Agent 1 thinks Agent 2 is?\par
A.Not given\par
B.Water\par
C.Food\par
D.Firewood\par
\textbf{Answer:}
\\
\midrule

\textbf{Desire} &
\textbf{Background:} Here is a negotiation conversation for a camping trip.
There are two agents who own some basic supplies and negotiate 
with each other to split the additional food packages, water bottles, 
and firewood to make their camping trip even better. Each of these items 
will be of either High, Medium or Low priority for these two agents. 
Each of the additional items only has an available quantity of 3. 
Please answer the following three questions using ``A'', ``B'', ``C'', 
``D'' without any explanation.\par
\textbf{Dialogue History:}\par
\textbf{\{Context\}}\par
\textbf{Question1:} What is agent 1's high preference for items based on the dialogue history?\par
A.Not given\par
B.Water\par
C.Food\par
D.Firewood\par
\textbf{Question2:} What is agent 1's medium preference for items based on the dialogue history?\par
A.Not given\par
B.Water\par
C.Food\par
D.Firewood\par
\textbf{Question3:} What is agent 1's low preference for items based on the dialogue history?\par
A.Not given\par
B.Water\par
C.Food\par
D.Firewood\par
\textbf{Answer:}
\\
\midrule

\textbf{Intention} &
\textbf{Background:} Here is a negotiation conversation for a camping trip.
There are two agents who own some basic supplies and negotiate 
with each other to split the additional food packages, water bottles, 
and firewood to make their camping trip even better. Each of these items 
will be of either High, Medium or Low priority for these two agents. 
Each of the additional items only has an available quantity of 3. \par
\textbf{Dialogue History:} \par
\textbf{\{Context\}} \par
\textbf{Question:} What are the plausible intentions of \{agent\} expressed 
in '\{utterance\}'? Based on the dialogue history, select one or more strategies 
(i.e., 'A', 'B', 'C', \ldots, 'I') from the following choices and their definition. 
Please select 'A', 'B', 'C', \ldots, 'I' without any explanation.\par

A. Build-Rapport: Participants discussing topics apart from the negotiation, 
in an attempt to build a rapport with the partner.\par

B. Show-Empathy: An utterance depicts empathy when there is evidence of positive 
acknowledgments or empathetic behavior towards a personal context of the partner.\par

C. Promote-Coordination: Used when a participant promotes coordination among the two partners.\par

D. Callout-Fairness: A callout to fairness for personal benefit, either when acknowledging 
a fair deal or when the opponent offers a deal that benefits them.\par

E. Undermine-Requirements: Refers to the scenario where a participant undermines 
the requirements of their opponent.\par

F. Discover-Preference: An attempt to discover the preference order of the opponent.\par

G. Describe-Need: Refers to arguments for creating a personal need for an item in the negotiation.\par

H. No-Need: When a participant points out that they do not need an item based on personal context.\par

I. No-Intention: If no strategy is evident, the utterance is labeled as No-Intention.\par

\textbf{Answer:}
\\
\bottomrule
\end{tabularx}

\centering 
\caption{Baseline prompt template. (XNegotiationToM - EN)}
\label{tab:baseline-prompt-template}
\end{table*}

\begin{CJK*}{UTF8}{gbsn}

\begin{table*}[htbp]
\scriptsize
\renewcommand{\arraystretch}{1.0}
\centering
\begin{tabularx}{\textwidth}{>{\centering\arraybackslash}p{2.5cm} X}
\toprule
\textbf{Dimension} & \textbf{Example (ZH)} \\
\midrule

\textbf{Belief} &
\textbf{Background:} 背景：以下是一次关于露营旅行的谈判对话。两位参与者拥有一些基本物资，并相互协商如何分配额外的食物、水和火柴，以使他们的露营旅行更加愉快。这些物品对每位参与者的重要性优先级可以是高、中或低。每种额外物品的最大可用数量为3。请仅使用"A"、"B"、"C"、"D"作答，不需要解释。\par
\textbf{Dialogue History:}\par
\textbf{\{Context\}}\par
\textbf{问题1：}根据对话，人物1认为人物2高优先级的物品是什么？\par
A.未提供\par
B.水\par
C.食物\par
D.火柴\par
\textbf{问题2：}根据对话，人物1认为人物2中优先级的物品是什么？\par
A.未提供\par
B.水\par
C.食物\par
D.火柴\par
\textbf{问题3：}根据对话，人物1认为人物2低优先级的物品是什么？\par
A.未提供\par
B.水\par
C.食物\par
D.火柴\par
\textbf{答案：}
\\
\midrule

\textbf{Desire} &
\textbf{Background:} 背景：以下是一次关于露营旅行的谈判对话。两位参与者拥有一些基本物资，并相互协商如何分配额外的食物、水和火柴，以使他们的露营旅行更加愉快。这些物品对每位参与者的重要性优先级可以是高、中或低。每种额外物品的最大可用数量为3。请仅使用"A"、"B"、"C"、"D"作答，不需要解释。\par
\textbf{Dialogue History:}\par
\textbf{\{Context\}}\par
\textbf{问题1：}根据对话历史，人物1高优先级的物品是什么？\par
A.未提供\par
B.水\par
C.食物\par
D.火柴\par
\textbf{问题2：}根据对话历史，人物1中优先级的物品是什么？\par
A.未提供\par
B.水\par
C.食物\par
D.火柴\par
\textbf{问题3：}根据对话历史，人物1低优先级的物品是什么？\par
A.未提供\par
B.水\par
C.食物\par
D.火柴\par
\textbf{答案：}
\\
\midrule

\textbf{Intention} &
\textbf{Background:} 背景：以下是一次关于露营旅行的谈判对话。两位参与者拥有一些基本物资，并相互协商如何分配额外的食物、水和火柴，以使他们的露营旅行更加愉快。这些物品对每位参与者的重要性优先级可以是高、中或低。每种额外物品的最大可用数量为3。\par
\textbf{Dialogue History:}\par
\textbf{\{Context\}}\par
\textbf{问题：}\{agent\}在"\{utterance\}"中表达的可能意图是什么？基于对话历史，从以下选项（即"A"、"B"、"C"、...、"I"）及其定义中选择一个或多个策略。请仅选择"A"、"B"、"C"、...、"I"，无需解释。\par

A. 建立融洽关系：参与者讨论与谈判无关的主题，试图与对方建立融洽关系。\par

B. 表达同情：当对方提到个人背景时，表现出积极的认可或同情行为的语句。\par

C. 促进协调：当参与者促进双方之间的协调时使用。\par

D. 呼吁公平：为了个人利益而呼吁公平，包括承认公平交易或对方提出有利于自己的交易时。\par

E. 破坏要求：指参与者破坏对方需求的情境。\par

F. 发现偏好：试图发现对方偏好顺序的行为。\par

G. 描述需求：为某一物品的个人需求提供论据。\par

H. 没有需求：根据个人背景指出他们不需要某一物品。\par

I. 没有意图：如果没有明显的策略，则将语句标记为"没有意图"。\par

\textbf{答案：}
\\
\bottomrule
\end{tabularx}

\centering
\caption{Baseline prompt template (XNegotiationToM - ZH)}
\label{tab:baseline-prompt-template-zh}
\end{table*}

\end{CJK*}

\begin{table*}[htbp]
\scriptsize
\renewcommand{\arraystretch}{1.0}
\centering
\begin{tabularx}{\textwidth}{>{\centering\arraybackslash}p{2.5cm} X}
\toprule
\textbf{Dimension} & \textbf{Example (DE)} \\
\midrule

\textbf{Belief} &
\textbf{Background:} Hintergrund: Hier ist ein Verhandlungsgespräch für einen Campingausflug. Es gibt zwei Agenten, die einige grundlegende Vorräte besitzen und miteinander verhandeln, um die zusätzlichen Lebensmittelpakete, Wasserflaschen und Brennholz aufzuteilen, um ihren Campingausflug noch besser zu machen. Jeder dieser Gegenstände hat für diese beiden Agenten entweder eine hohe, mittlere oder niedrige Priorität. Für jeden der zusätzlichen Gegenstände ist nur eine Menge von 3 verfügbar. Bitte beantworten Sie die folgenden drei Fragen mit "A", "B", "C", "D" ohne Begründung.\par
\textbf{Dialogverlauf:}\par
\textbf{\{Context\}}\par
\textbf{Frage1:} Nach dem Dialog, was ist die hohe Präferenz für Gegenstände von Person 2 laut Person 1?\par
A. Nicht angegeben\par
B. Wasser\par
C. Essen\par
D. Brennholz\par
\textbf{Frage2:} Nach dem Dialog, was ist die mittlere Präferenz für Gegenstände von Person 2 laut Person 1?\par
A. Nicht angegeben\par
B. Wasser\par
C. Essen\par
D. Brennholz\par
\textbf{Frage3:} Nach dem Dialog, was ist die niedrige Präferenz für Gegenstände von Person 2 laut Person 1?\par
A. Nicht angegeben\par
B. Wasser\par
C. Essen\par
D. Brennholz\par
\textbf{Antwort:}
\\
\midrule

\textbf{Desire} &
\textbf{Background:} Hintergrund: Hier ist ein Verhandlungsgespräch für einen Campingausflug. Es gibt zwei Agenten, die einige grundlegende Vorräte besitzen und miteinander verhandeln, um die zusätzlichen Lebensmittelpakete, Wasserflaschen und Brennholz aufzuteilen, um ihren Campingausflug noch besser zu machen. Jeder dieser Gegenstände hat für diese beiden Agenten entweder eine hohe, mittlere oder niedrige Priorität. Für jeden der zusätzlichen Gegenstände ist nur eine Menge von 3 verfügbar. Bitte beantworten Sie die folgenden drei Fragen mit "A", "B", "C", "D" ohne Begründung.\par
\textbf{Dialogverlauf:}\par
\textbf{\{Context\}}\par
\textbf{Frage1:} Was ist die hohe Präferenz für Gegenstände von Person 1 nach dem Dialogverlauf?\par
A. Nicht angegeben\par
B. Wasser\par
C. Essen\par
D. Brennholz\par
\textbf{Frage2:} Was ist die mittlere Präferenz für Gegenstände von Person 1 nach dem Dialogverlauf?\par
A. Nicht angegeben\par
B. Wasser\par
C. Essen\par
D. Brennholz\par
\textbf{Frage3:} Was ist die niedrige Präferenz für Gegenstände von Person 1 nach dem Dialogverlauf?\par
A. Nicht angegeben\par
B. Wasser\par
C. Essen\par
D. Brennholz\par
\textbf{Antwort:}
\\
\midrule

\textbf{Intention} &
\textbf{Background:} Hintergrund: Hier ist ein Verhandlungsgespräch für einen Campingausflug. Es gibt zwei Agenten, die einige grundlegende Vorräte besitzen und miteinander verhandeln, um die zusätzlichen Lebensmittelpakete, Wasserflaschen und Brennholz aufzuteilen, um ihren Campingausflug noch besser zu machen. Jeder dieser Gegenstände hat für diese beiden Agenten entweder eine hohe, mittlere oder niedrige Priorität. Für jeden der zusätzlichen Gegenstände ist nur eine Menge von 3 verfügbar.\par
\textbf{Dialogverlauf:}\par
\textbf{\{Context\}}\par
\textbf{Frage:} Was sind die plausiblen Absichten von \{agent\}, die in "\{utterance\}" ausgedrückt werden? Wählen Sie nach dem Dialogverlauf eine oder mehrere Strategien (d. h. "A", "B", "C", …, "I") aus den folgenden Optionen und deren Definition aus. Wählen Sie "A", "B", "C", …, "I" ohne Begründung aus.\par

A. Beziehung aufbauen: Teilnehmer diskutieren Themen abseits der Verhandlung, um ein Vertrauensverhältnis zum Partner aufzubauen.\par

B. Empathie zeigen: Eine Äußerung zeigt Empathie, wenn es Anzeichen für positive Anerkennungen oder empathisches Verhalten gegenüber einem persönlichen Kontext des Partners gibt.\par

C. Koordination fördern: Wird verwendet, wenn ein Teilnehmer die Koordination zwischen den beiden Partnern fördert.\par

D. Fairness einfordern: Ein Aufruf zur Fairness für persönlichen Vorteil, entweder wenn ein fairer Deal anerkannt wird oder wenn der Gegner einen Deal anbietet, der ihm Vorteile bringt.\par

E. Anforderungen untergraben: Bezieht sich auf das Szenario, in dem ein Teilnehmer die Anforderungen seines Gegners untergräbt.\par

F. Präferenz herausfinden: Ein Versuch, die Präferenzreihenfolge des Gegners herauszufinden.\par

G. Bedarf beschreiben: Bezieht sich auf Argumente für die Schaffung eines persönlichen Bedarfs für einen Gegenstand in der Verhandlung.\par

H. Kein Bedarf: Wenn ein Teilnehmer darauf hinweist, dass er einen Gegenstand aufgrund des persönlichen Kontexts nicht benötigt.\par

I. Keine Absicht: Wenn keine Strategie erkennbar ist, wird die Äußerung als Keine Absicht gekennzeichnet.\par

\textbf{Antwort:}
\\
\bottomrule
\end{tabularx}
\centering
\caption{Baseline prompt template (XNegotiationToM - DE)}
\label{tab:baseline-prompt-template-de}
\end{table*}

\begin{table*}[htbp]
\scriptsize
\renewcommand{\arraystretch}{1.0}
\centering
\begin{tabularx}{\textwidth}{>{\centering\arraybackslash}p{2.5cm} X}
\toprule
\textbf{Dimension} & \textbf{Example (FR)} \\
\midrule

\textbf{Belief} &
\textbf{Background:} Contexte : Voici une conversation de négociation pour un voyage de camping. Il y a deux agents qui possèdent quelques fournitures de base et négocient entre eux pour répartir les paquets de nourriture supplémentaires, les bouteilles d'eau et les bois de chauffage afin d'améliorer leur voyage de camping. Chacun de ces éléments sera d'une priorité Haute, Moyenne ou Faible pour ces deux agents. Chacun des articles supplémentaires n'a qu'une quantité disponible de 3. Veuillez répondre aux trois questions suivantes en utilisant "A", "B", "C", "D" sans aucune explication.\par
\textbf{Historique de la conversation:}\par
\textbf{\{Context\}}\par
\textbf{Question 1:} D'après le dialogue, quels sont les articles que la Personne 1 considère comme étant de haute priorité pour la Personne 2?\par
A. Pas donné\par
B. Eau\par
C. Nourriture\par
D. Bois de chauffage\par
\textbf{Question 2:} D'après le dialogue, quels sont les articles que la Personne 1 considère comme étant de priorité moyenne pour la Personne 2?\par
A. Pas donné\par
B. Eau\par
C. Nourriture\par
D. Bois de chauffage\par
\textbf{Question 3:} D'après le dialogue, quels sont les articles que la Personne 1 considère comme étant de faible priorité pour la Personne 2?\par
A. Pas donné\par
B. Eau\par
C. Nourriture\par
D. Bois de chauffage\par
\textbf{Réponse:}
\\
\midrule

\textbf{Desire} &
\textbf{Background:} Contexte : Voici une conversation de négociation pour un voyage de camping. Il y a deux agents qui possèdent quelques fournitures de base et négocient entre eux pour répartir les paquets de nourriture supplémentaires, les bouteilles d'eau et les bois de chauffage afin d'améliorer leur voyage de camping. Chacun de ces éléments sera d'une priorité Haute, Moyenne ou Faible pour ces deux agents. Chacun des articles supplémentaires n'a qu'une quantité disponible de 3. Veuillez répondre aux trois questions suivantes en utilisant "A", "B", "C", "D" sans aucune explication.\par
\textbf{Historique de la conversation:}\par
\textbf{\{Context\}}\par
\textbf{Question 1:} Quels sont les articles de haute priorité pour la Personne 1?\par
A. Pas donné\par
B. Eau\par
C. Nourriture\par
D. Bois de chauffage\par
\textbf{Question 2:} Quels sont les articles de priorité moyenne pour la Personne 1?\par
A. Pas donné\par
B. Eau\par
C. Nourriture\par
D. Bois de chauffage\par
\textbf{Question 3:} Quels sont les articles de priorité faible pour la Personne 1?\par
A. Pas donné\par
B. Eau\par
C. Nourriture\par
D. Bois de chauffage\par
\textbf{Réponse:}
\\
\midrule

\textbf{Intention} &
\textbf{Background:} Contexte : Voici une conversation de négociation pour un voyage de camping. Il y a deux agents qui possèdent quelques fournitures de base et négocient entre eux pour répartir les paquets de nourriture supplémentaires, les bouteilles d'eau et les bois de chauffage afin d'améliorer leur voyage de camping. Chacun de ces éléments sera d'une priorité Haute, Moyenne ou Faible pour ces deux agents. Chacun des articles supplémentaires n'a qu'une quantité disponible de 3.\par
\textbf{Historique de la conversation:}\par
\textbf{\{Context\}}\par
\textbf{Question:} Quelles sont les intentions plausibles de \{agent\} exprimées dans "\{utterance\}"? Sur la base de l'historique de la conversation, sélectionnez une ou plusieurs stratégies (c.-à-d., "A", "B", "C", ..., "I") parmi les choix suivants et leur définition. Veuillez sélectionner "A", "B", "C", ..., "I" sans aucune explication.\par

A. Établir des relations: Les participants discutent de sujets autres que la négociation, dans le but de créer une relation avec le partenaire.\par

B. Faire preuve d'empathie: Une énonciation montre de l'empathie lorsqu'il y a des preuves de reconnaissance positive ou de comportement empathique envers un contexte personnel du partenaire.\par

C. Promouvoir la coordination: Utilisé lorsqu'un participant favorise la coordination entre les deux partenaires.\par

D. Revendiquer l'équité: Un appel à l'équité pour un avantage personnel, soit en reconnaissant un accord équitable, soit lorsque l'adversaire propose un accord qui lui profite.\par

E. Saper les exigences: Désigne le cas où un participant sape les exigences de son adversaire.\par

F. Découvrir la préférence: Une tentative de découvrir l'ordre de préférence de l'adversaire.\par

G. Décrire le besoin: Arguments visant à créer un besoin personnel pour un article dans la négociation.\par

H. Aucun besoin: Lorsqu'un participant indique qu'il n'a pas besoin d'un article selon son contexte personnel.\par

I. Aucune intention: Si aucune stratégie n'est évidente, l'énoncé est étiqueté comme Aucune-Intention.\par

\textbf{Réponse:}
\\
\bottomrule
\end{tabularx}
\centering
\caption{Baseline prompt template (XNegotiationToM - FR)}
\label{tab:baseline-prompt-template-fr}
\end{table*}

\begin{CJK*}{UTF8}{min} 

\begin{table*}[htbp]
\scriptsize
\renewcommand{\arraystretch}{1.0}
\centering
\begin{tabularx}{\textwidth}{>{\centering\arraybackslash}p{2.5cm} X}
\toprule
\textbf{Dimension} & \textbf{Example (JA)} \\
\midrule

\textbf{Belief} &
\textbf{Background:} 背景：以下はキャンプ旅行に関する交渉の会話です。二人の登場人物が基本的な用品を所有しており、追加の食べ物、水、薪を分配して、キャンプ旅行をより良くするために交渉しています。これらの各アイテムは、二人にとって「高」、「中」、「低」のいずれかの優先度を持ちます。追加アイテムは各々最大3個までしか利用できません。\par

以下の3つの質問に"A"、"B"、"C"、"D"を使って説明なしで答えてください。\par

\textbf{対話履歴:}\par
\textbf{\{Context\}}\par
\textbf{質問1:} 対話に基づき、人物1が人物2について考える「高」優先度のアイテムは何ですか？\par
A. 未提供\par
B. 水\par
C. 食べ物\par
D. 薪\par
\textbf{質問2:} 対話に基づき、人物1が人物2について考える「中」優先度のアイテムは何ですか？\par
A. 未提供\par
B. 水\par
C. 食べ物\par
D. 薪\par
\textbf{質問3:} 対話に基づき、人物1が人物2について考える「低」優先度のアイテムは何ですか？\par
A. 未提供\par
B. 水\par
C. 食べ物\par
D. 薪\par
\textbf{回答:}
\\
\midrule

\textbf{Desire} &
\textbf{Background:} 背景：以下はキャンプ旅行に関する交渉の会話です。二人の登場人物が基本的な用品を所有しており、追加の食べ物、水、薪を分配して、キャンプ旅行をより良くするために交渉しています。これらの各アイテムは、二人にとって「高」、「中」、「低」のいずれかの優先度を持ちます。追加アイテムは各々最大3個までしか利用できません。\par

以下の3つの質問に"A"、"B"、"C"、"D"を使って説明なしで答えてください。\par

\textbf{対話履歴:}\par
\textbf{\{Context\}}\par
\textbf{質問1:} 対話履歴に基づいて、人物1の「高」優先度のアイテムは何ですか？\par
A. 未提供\par
B. 水\par
C. 食べ物\par
D. 薪\par
\textbf{質問2:} 対話履歴に基づいて、人物1の「中」優先度のアイテムは何ですか？\par
A. 未提供\par
B. 水\par
C. 食べ物\par
D. 薪\par
\textbf{質問3:} 対話履歴に基づいて、人物1の「低」優先度のアイテムは何ですか？\par
A. 未提供\par
B. 水\par
C. 食べ物\par
D. 薪\par
\textbf{回答:}
\\
\midrule

\textbf{Intention} &
\textbf{Background:} 背景：以下はキャンプ旅行に関する交渉の会話です。二人の登場人物が基本的な用品を所有しており、追加の食べ物、水、薪を分配して、キャンプ旅行をより良くするために交渉しています。これらの各アイテムは、二人にとって「高」、「中」、「低」のいずれかの優先度を持ちます。追加アイテムは各々最大3個までしか利用できません。\par

\textbf{対話履歴:}\par
\textbf{\{Context\}}\par
\textbf{質問:} \{agent\}が発した「\{utterance\}」において、考えられる意図は何ですか？対話履歴に基づいて、以下の選択肢から1つ以上の戦略（"A"、"B"、"C"、...、"I"）を選択してください。説明なしで"A"、"B"、"C"、...、"I"を選択してください。\par

A. 信頼関係を築く：交渉とは別の話題を議論し、相手との信頼関係を築こうとする発言。\par

B. 共感を示す：相手の個人的な文脈に対して、肯定的な応答や共感的な行動が見られる発言。\par

C. 協調を促進する：両者間の協調を促進しようとする発言。\par

D. 公平性を求める：自分に有利な条件を認める、または相手の提案が自分に利益をもたらすことを指摘する発言。\par

E. 要件を損なう：相手の要件を軽視または否定する発言。\par

F. 好みを見つける：相手の優先順位を探ろうとする発言。\par

G. 需要を説明する：自分のアイテム需要を論じる発言。\par

H. 要求なし：個人的な文脈に基づき、アイテムが必要ないことを指摘する発言。\par

I. 意図なし：特定の戦略が明らかでない場合、このラベルが適用される。\par

\textbf{回答:}
\\
\bottomrule
\end{tabularx}
\centering
\caption{Baseline prompt template (XNegotiationToM - JA)}
\label{tab:baseline-prompt-template-ja}
\end{table*}

\end{CJK*}

\begin{table*}[htbp]
\scriptsize
\renewcommand{\arraystretch}{1.0}
\centering
\begin{tabularx}{\textwidth}{>{\centering\arraybackslash}p{2cm} X}
\toprule
\textbf{Language} & \textbf{Template} \\
\midrule

\textbf{EN} & 
\{story\}\par
\{question\}\par
Choose from the following:\par
(a) \{containers\_0\}\par
(b) \{containers\_1\}\par
Keep your answer concise. Answer with a single word.
\\
\midrule

\textbf{ZH} &
\begin{CJK*}{UTF8}{gbsn}
\{story\}\par
\{question\}\par
请从以下选项中选择：\par
(a) \{containers\_0\}\par
(b) \{containers\_1\}\par
请保持答案简洁。使用单个词回答。
\end{CJK*}
\\
\midrule

\textbf{DE} &
\{story\}\par
\{question\}\par
Wählen Sie aus den folgenden Möglichkeiten:\par
(a) \{containers\_0\}\par
(b) \{containers\_1\}\par
Halten Sie Ihre Antwort kurz. Antworten Sie mit einem einzigen Wort.
\\
\midrule

\textbf{FR} &
\{story\}\par
\{question\}\par
Choisissez parmi les options suivantes :\par
(a) \{containers\_0\}\par
(b) \{containers\_1\}\par
Gardez votre réponse concise. Répondez avec un seul mot.
\\
\midrule

\textbf{JA} &
\begin{CJK*}{UTF8}{min}
\{story\}\par
\{question\}\par
以下から選択してください：\par
(a) \{containers\_0\}\par
(b) \{containers\_1\}\par
簡潔に答えてください。一語で回答してください。
\end{CJK*}
\\

\bottomrule
\end{tabularx}
\caption{Multi-language prompt templates (XToMi)}
\label{tab:tomi-lang-all}
\end{table*}

\begin{table*}[htbp]
\scriptsize
\renewcommand{\arraystretch}{1.0}
\centering
\begin{tabularx}{\textwidth}{>{\centering\arraybackslash}p{2cm} X}
\toprule
\textbf{Language} & \textbf{Template} \\
\midrule

\textbf{EN} & 
\{Context\}\par
Question: \par
\{Question\} \par
(a) \{Answer\_0\}\par
(b) \{Answer\_1\}\par
Choose an answer from above:
\\
\midrule

\textbf{ZH} &
\begin{CJK*}{UTF8}{gbsn}
\{Context\}\par
问题： \par
\{Question\} \par
(a) \{Answer\_0\}\par
(b) \{Answer\_1\}\par
从上面选择一个答案：
\end{CJK*}
\\
\midrule

\textbf{DE} &
\{Context\}\par
Frage: \par
\{Question\} \par
(a) \{Answer\_0\}\par
(b) \{Answer\_1\}\par
Wähle eine Antwort von oben: 
\\
\midrule

\textbf{FR} &
\{Context\}\par
Question : \par
\{Question\} \par
(a) \{Answer\_0\}\par
(b) \{Answer\_1\}\par
Choisissez une réponse ci-dessus :
\\
\midrule

\textbf{JA} &
\begin{CJK*}{UTF8}{min}
\{Context\}\par
質問: \par
\{Question\} \par
(a) \{Answer\_0\}\par
(b) \{Answer\_1\}\par
上記の中から答えを選んでください: 
\end{CJK*}
\\

\bottomrule
\end{tabularx}
\caption{Multi-language prompt templates (XFANToM)}
\label{tab:FANToM-lang-all}
\end{table*}

\begin{table*}[!] 
\small
\centering
\scalebox{0.86}{
\begin{tabular}{l}
\toprule
\textbf{FANToM Conversation:}\\
...\textbf{\texttt{ \textcolor{headcolor}{<The-Beginning-Sentences-of-Example>}}}...\\
\\
Person 1: Definitely the vibe. The charm of Paris during Christmas is unmatched. The beautiful decorations, the bustling markets, and the \\ 
lovely smell of fresh baked goods - it all was so different and vibrant. Have you ever had a chance to explore Paris yourself?\\
Person 2: No, not yet, but I would definitely love to. I've heard the Eiffel Tower looks phenomenal during Christmas. You know, the holiday\\
which I found interesting was \textbf{Thanksgiving}. It was during one instance when I was in college.\\
Person 1: That's interesting. What was so memorable about it?\\
Person 2: We didn't return home during the break because we were preparing for our finals. My friends and I decided to cook a \\ 
\textbf{Thanksgiving meal} together; none of us had any cooking experience prior to that. The \textbf{turkey}, a total disaster, but the cornbread and pie \\
were amazing. That was the occasion when I realized I was quite a decent baker!\\
Person 1: That sounds like a fun experience, especially the part where you discovered your hidden talent. Isn't it amazing how holidays often \\
lead us to such happy surprises?
Person 2: Absolutely. Holidays indeed have a charm of their own. They bring us closer, build strong bonds\\
and generate memories that last a lifetime.\\
Person 1: Couldn't agree more, Person 2. Let's hope we create some fantastic memories here today as well!\\
Person 2: Certainly, Person 1. I’m looking forward to it.\\
Person 3: Hello Person 1, Person 2, I hope I'm not interrupting anything. This place is really coming alive with preparations\\
for the iday.\\
Person 1: Hello Person 3, not at all! We were just sharing our cherished holiday experiences. Person 2 here discovered he could bake \\
uring a college \textbf{Thanksgiving}!\\
Person 3: Baking, huh? That sounds a lot like my New Year's experience. I had to whip up a dessert at the last minute.\\
Person 2: Really? What was it? I love hearing about impromptu cooking ordeals!\\
Person 3: It was a chocolate cheesecake. I'm personally more of a main course guy, but that was the only thing we were missing. \\
Turned out well, surprisingly!\\
Person 1: Haha, both of you men of hidden talents! You know, I've had my own fair share of kitchen successes and failures during Easter.\\
Quite the adventure it is.\\
Person 3: Then I guess we can all agree that holidays are for discovering new facets about ourselves, right?\\
Person 2: Couldn't have put it better, Person 3. To holidays and their endless surprises!
Person 4: Hello everyone, what an exciting\\
conversation you're all having!\\
Person 2: Hi Person 4! Good to see you. We've been sharing our holiday stories and about the discoveries we made about\\
ourselves during those times.\\
Person 4: Absolutely love the idea. We all have something to learn from our holidays.\\
\\
...\textbf{\texttt{ \textcolor{relationcolor}{<The-Middle-Sentences-of-Example>}}}...\\
\\
Person 5: Hello everyone, I see we have a lively discussion happening over here!\\
Person 1: Hi Person 5, good to see you. We were just reminiscing about our holiday experiences and how they've led to surprising discoveries. \\
Now, if we're talking about holidays, we surely can't ignore the fun and sometimes stress of holiday gifting and shopping, right?\\
Person 5: Oh, absolutely. I usually take the role of Santa in my family, deciding and buying gifts for everyone. It's a challenge, but I love \\
the joy it brings.\\
Person 2: Funny you mentioned Santa, Person 5. One Christmas, I decided to handcraft all my gifts. It showed me I have a decent knack\\
for arts and crafts.\\
Person 4: Crafting your own gifts - that's lovely, Person 2. I remember one year, I ended up forgetting about buying gifts until the last minute.\\
I ended up learning that I'm pretty good at making quick decisions!\\
Person 3: Seems like we've all learned something new about ourselves from holiday shopping. For me, it's probably the fact that I'm surprisingly \\
frugal. I tend to find great gifts without breaking the bank!\\
\\
...\textbf{\texttt{ \textcolor{tailcolor}{<The-Rest-Sentences-of-Example>}}}...\\
\midrule

\textbf{Belief Question:} What does Person 5 believe about how these holiday experiences led to surprising discoveries about their cooking and baking abilities?\\
\textbf{A :} "Person 5 is unaware of how these holiday experiences led to surprising discoveries about their cooking and baking abilities. This is due to the fact\\
that he was not involved in the conversation when this topic was discussed."\\
\textbf{B :} "Person 5 believes that these holiday experiences led Person 2, Person 3, and Person 4 to surprising discoveries about their cooking and baking\\
abilities. Person 2 discovered his baking skills when he made an amazing cornbread and pie during a Thanksgiving in college. Person 3 found out about\\
his knack for dessert-making when he made a chocolate cheesecake for New Year's unexpectedly. Person 4 found her passion for cooking during a\\ 
family spring holiday where she enjoyed preparing the main meal."\\
            
Ground Truth: A \\

\midrule

\textbf{English Version Answer:} A\\
\textbf{Chinese Version Answer:} B\\
\textbf{German Version Answer:} A\\
\textbf{French Version Answer:} A\\
\textbf{Japanese Version Answer:} B\\

\bottomrule

\end{tabular}}
\caption{A case study on the cultural context embedded in language in XFANToM}
\label{tab:case_study}
\end{table*}

\end{document}